\documentclass[cspaper,compsoc]{IEEEtran} 

\ifCLASSOPTIONcompsoc
  \usepackage[nocompress]{cite}
\else
  \usepackage{cite}
\fi
\ifCLASSINFOpdf
\else
\fi

\usepackage{ifpdf}

\usepackage[pdftex]{graphicx}
\usepackage{epsfig} 
\usepackage{amsthm}
\usepackage{amsmath}
\usepackage{amsfonts}
\usepackage{amssymb}
\usepackage{bm}
\usepackage{cases}
\usepackage{multirow}

\usepackage{multicol}
\usepackage{xcolor}
\usepackage{float,subfig}
\usepackage{caption}
\usepackage{cite}
\usepackage{url}
\usepackage{here}
\usepackage{stfloats} 
\usepackage{braket}
\usepackage[linesnumbered,ruled,vlined]{algorithm2e}
\usepackage{algorithmic}

\usepackage{multicol}
\usepackage{enumerate}
\usepackage[export]{adjustbox}
\usepackage{hhline}
\usepackage{colortbl}

\usepackage{hyperref}
\hypersetup{
    colorlinks=true,
    linkcolor=blue, 
    urlcolor=magenta,
    }
\makeatletter
\let\saved@hyper@linkurl\hyper@linkurl
\let\saved@hyper@link@\hyper@link@
\AtBeginDocument{%

  \NoHyper 

  \let\hyper@linkurl\saved@hyper@linkurl 
  \let\hyper@link@\saved@hyper@link@ 
}
\makeatother

\usepackage{array}
\newcolumntype{C}[1]{>{\hfil}m{#1}<{\hfil}}

\usepackage{lscape} 

\newcommand{\argmin}{\mathop{\rm \text{arg~min}}\limits}

\begin{document}
%
\title{Privacy-preserving Continual Federated Clustering via Adaptive Resonance Theory}
%
%
%
%

\author{
	Naoki~Masuyama,
	Yusuke~Nojima,
	Yuichiro~Toda,
	Chu Kiong Loo,
	Hisao~Ishibuchi,
	and~Naoyuki Kubota
	\thanks{N. Masuyama, and Y. Nojima are with the Graduate School of Informatics, Department of Core Informatics, Osaka Metropolitan University, 1-1 Gakuen-cho Naka-ku, Sakai-Shi, Osaka 599-8531, Japan, e-mails: masuyama@omu.ac.jp, nojima@omu.ac.jp.}
	\thanks{Y. Toda is with the Faculty of Environmental, Life, Natural Science and Technology, Okayama University, 3-1-1 Tsushima-Naka, Kita-ku, Okayama-shi, Okayama 700-8530, Japan, e-mail: ytoda@okayama-u.ac.jp}
	\thanks{C. K. Loo is with the Department of Artificial Intelligence, Faculty of Computer Science and Information Technology, Universiti Malaya, 50603 Kuala Lumpur, Malaysia, e-mail: ckloo.um@um.edu.my}
	\thanks{H. Ishibuchi is with the Guangdong Provincial Key Laboratory of Brain-inspired Intelligent Computation, Department of Computer Science and Engineering, Southern University of Science and Technology, Shenzhen 518055, China, e-mail: hisao@sustech.edu.cn.}
	\thanks{N. Kubota is with the Graduate School of Systems Design, Department of Mechanical Systems Engineering, Tokyo Metropolitan University, 6-6 Asahigaoka, Hino-Shi, Tokyo 191-0065, Japan, e-mail: kubota@tmu.ac.jp.}
	\thanks{Corresponding author: Naoki Masuyama (e-mail: masuyama@omu.ac.jp).}
	\thanks{Manuscript received April 19, 2005; revised August 26, 2015.}
}

%
%

\markboth{Journal of \LaTeX\ Class Files,~Vol.~14, No.~8, August~2015}%
{Shell \MakeLowercase{\textit{et al.}}: Bare Demo of IEEEtran.cls for Computer Society Journals}
%



\IEEEtitleabstractindextext{%
\begin{abstract}
	With the increasing importance of data privacy protection, various privacy-preserving machine learning methods have been proposed. In the clustering domain, various algorithms with a federated learning framework (i.e., federated clustering) have been actively studied and showed high clustering performance while preserving data privacy. However, most of the base clusterers (i.e., clustering algorithms) used in existing federated clustering algorithms need to specify the number of clusters in advance. These algorithms, therefore, are unable to deal with data whose distributions are unknown or continually changing. To tackle this problem, this paper proposes a privacy-preserving continual federated clustering algorithm. In the proposed algorithm, an adaptive resonance theory-based clustering algorithm capable of continual learning is used as a base clusterer. Therefore, the proposed algorithm inherits the ability of continual learning. Experimental results with synthetic and real-world datasets show that the proposed algorithm has superior clustering performance to state-of-the-art federated clustering algorithms while realizing data privacy protection and continual learning ability. The source code is available at \url{https://github.com/Masuyama-lab/FCAC}.
\end{abstract}

}

\maketitle

\IEEEdisplaynontitleabstractindextext

%
\IEEEpeerreviewmaketitle

\IEEEraisesectionheading{\section{Introduction}\label{sec:introduction}}

%
%
%
\IEEEPARstart{I}{N} a society with advanced information technology, privacy protection techniques in data utilization are becoming increasingly important. Various methods have been proposed to protect data privacy. Secure computation and secret sharing are widely used as cryptographic methods \cite{sun20}. These methods realize the sharing of information while preserving anonymity and confidentiality by using encryption keys. As mathematical-based methods, anonymization improves data privacy by converting or removing personally identifiable information \cite{majeed22}. Differential privacy adds noise to training data or outputs of a trained model. This is to make it difficult to determine whether a particular individual exists in the training data \cite{dwork06b,dwork06}. Specifically, the approach that adds noise to the training data is called local differential privacy, while the approach that adds noise to the output of the trained model is called global differential privacy. In the machine learning domain, federated learning is known as a privacy-preserving distributed learning method \cite{mcmahan17}. Federated learning performs learning from distributed data across multiple clients without aggregating the data on a single server. The parameters of a trained model in each client are then consolidated on a model in a server. In addition, federated learning can further improve data privacy by applying differential privacy \cite{zhao20,li23}.

In recent years, the importance of data privacy has also increased in the clustering domain \cite{biswas19,majeed22,ikotun22}. Among privacy-preserving clustering algorithms, federated clustering (i.e., an clustering algorithm applies a federated learning framework) has attracted much attention because of its usefulness in practical applications \cite{dennis21}. As a base clusterer (i.e., a clustering algorithm) of federated clustering, in general, centroid-based clustering algorithms such as $k$-means, fuzzy $c$-means, and Gaussian Mixture Model (GMM) are often used \cite{dennis21,pedrycz21,stallmann22,pandhare21}. Although centroid-based clustering algorithms are simple and highly applicable, these algorithms need to specify the number of centroids in advance. This drawback makes it difficult to apply these algorithms to data whose distributions are unknown and/or continually changing.

Among clustering algorithms, Adaptive Resonance Theory (ART)-based approaches are capable of continual learning without catastrophic forgetting by adaptively generating centroids (nodes) depending on the distributions of given data \cite{masuyama19b}. In particular, ART-based clustering algorithms with Correntropy-Induced Metric (CIM) \cite{liu07} as a similarity measure show faster and more stable self-organizing ability than other clustering algorithms \cite{carpenter91b,vigdor07,wang19,masuyama22b,masuyama23}. In general, ART-based clustering algorithms generate a number of representative points (nodes) from given data. Thus, data privacy is more or less considered since the given data itself is not retained. However, the ability to protect data privacy is insufficient because no data privacy protection techniques are explicitly applied to those algorithms.

This paper proposes a new privacy-preserving federated clustering algorithm, called Federated Clustering via ART-based Clustering (FCAC), which applies local differential privacy to an ART-based clustering algorithm in a federated learning framework. FACA performs clustering for distributed training data across multiple clients without aggregating the data on a single server while preserving data privacy. In FCAC, CIM-based ART with Edge (CAE) \cite{masuyama23} after a minor modification, which is called CAE for Federated Clustering (CAE$_{\text{FC}}$), is used as a base clusterer for a server. Although the original CAE is a state-of-the-art parameter-free ART-based topological clustering algorithm, CAE often generates a large number of nodes. Therefore, we introduce a minor modification for reducing the number of generated nodes. A base clusterer for each client of FCAC is CAE$_{\text{FC}}$ without topology (i.e., edges), called CA+, which is first introduced in this paper. Since a trained model of a server is a clustering result, the server is required to continually generate well-separated clusters. In contrast, each client is required to generate a number of nodes that can continually and appropriately approximate the distributions of the training data in the client. This is because the generated nodes in each client are utilized as training data for the server. Therefore, we introduce CA+ as the base clusterer for each client, which only generates nodes from given data in each client. Thanks to CAE$_{\text{FC}}$ and CA+, FCAC can adaptively, efficiently, and continually generate topological networks from the given data in each client. Note that continual learning is generally categorized into three scenarios: domain incremental learning, task incremental learning, and class incremental learning \cite{van19, wiewel19}. Since CAE$_{\text{FC}}$ and CA+ are capable of class incremental learning, FCAC inherits the ability of class incremental learning.

The contributions of this paper are summarized as follows:
\begin{itemize}
	\item[(I)] FCAC is proposed as a new privacy-preserving federated clustering algorithm capable of continual learning. FCAC explicitly considers the protection of data privacy by applying local differential privacy a federated learning framework. To the best of our knowledge, FCAC is the first ART-based privacy-preserving federated clustering algorithm.
	
	\item[(I\hspace{-1.2pt}I)] A new clustering algorithm called CA+, which is a variant of CAE$_{\text{FC}}$, is introduced. The self-organizing ability of CA+ satisfies the demand of a client in FCAC, i.e., generated nodes by CA+ can continually and appropriately approximate the distributions of the training data in each client.
	
	\item[(I\hspace{-1.2pt}I\hspace{-1.2pt}I)] Empirical studies show that FCAC has superior clustering performance to state-of-the-art algorithms while protecting data privacy and maintaining continual learning ability.
\end{itemize}

The paper is organized as follows. Section \ref{sec:literature} presents a literature review for growing self-organizing clustering and privacy-preserving clustering algorithms. Section \ref{sec:preliminary} presents the preliminary knowledge for CAE. Section \ref{sec:proposedAlgorithm} explains the learning procedure of the proposed FCAC algorithm in detail. Section \ref{sec:experiment} presents extensive simulation experiments to evaluate its clustering performance by using synthetic and real-world datasets. Section \ref{sec:conclusion} concludes this paper.

Table \ref{tab:notations} summarizes the main notations used in FCAC and related functions/algorithms.

\begin{table}[htbp]
	\centering
	\caption{Summary of notations}
	\renewcommand{\arraystretch}{1.2}
	\label{tab:notations}
	\scalebox{0.98}{
	\begin{tabular}{ll}
		\hline\hline
		Notation               & Description  \\
		\hline
		$\mathbf{x}$           & $d$-dimensional data point  \\
		$\mathcal{X}_{c} $     & set of data points for a client $c$ ($\mathbf{x} \in \mathcal{X}_{c}$) \\
		$\mathbf{y}_{k}$       & $k$-th node \\
		$\mathcal{Y}$          & set of nodes ($\mathbf{y}_{k} \in \mathcal{Y}$) \\
		$|\mathcal{Y}|$        & number of nodes in $\mathcal{Y}$  \\
		$\mathbf{y}_{s_{1}}$   & 1st winner node \\
		$\mathbf{y}_{s_{2}}$   & 2nd winner node  \\
		$\hat{C}(\cdot)$       & correntropy  \\
		$\mathrm{CIM}(\cdot)$  & correntropy-induced metric  \\
		$\sigma$               & bandwidth of a kernel function in CIM  \\
		$\mathcal{S}$          & set of bandwidths of a kernel function $\sigma \in \mathcal{S}$ \\
		$ \mathcal{N}_{k} $    & set of neighbor nodes of node $ \mathbf{y}_{k} $   \\
		$\lambda$              & number of active nodes  \\
		$\mathcal{A}$          & set of active nodes \\
		$D$          		   & diversity of a set of active nodes \\
		$V_{s_{1}}$            & CIM value between $\mathbf{x}$ and $\mathbf{y}_{s_{1}}$ \\
		$V_{s_{2}}$            & CIM value between $\mathbf{x}$ and $\mathbf{y}_{s_{2}}$ \\
		$\mathbf{R}$           & matrix of pairwise similarities \\
		$V_{\text{threshold}}$ & similarity threshold (a vigilance parameter)   \\
		$ M_{k} $              & winning count of $ \mathbf{y}_{k} $ \\
		$ \mathcal{M} $        & set of winning counts $ M_{k} $ ($ M_{k} \in \mathcal{M}$)  \\
		$ a(\mathbf{y}_{k},\mathbf{y}_{l}) $   & age of an edge between nodes $\mathbf{y}_{k}$ and $\mathbf{y}_{l} \in \mathcal{Y} \setminus \mathbf{y}_{k}$ \\
		$ \mathcal{E} $   	   & set of ages of edges ($a(\mathbf{y}_{k},\mathbf{y}_{l}) \in \mathcal{E}$) \\
		$\alpha_{\text{del}}$  & set of ages of deleted edges  \\
		$a_{\text{max}}$       & edge deletion threshold  \\
		
		CA+                    & base clusterer for a client \\
		CAE$_{\text{FC}}$      & base clusterer for a server \\
		$\epsilon$             & privacy budget  \\
		$\Delta f$       	   & local sensitivity \\
		$L(\cdot)$             & inverse cumulative density function \\
		$\mathcal{X}_{c}' $    & set of data points for a client $c$ with differential privacy \\
		$ C $       	       & number of clients for federatd clustering \\
		$ \mathbf{Y}^{\geq75\text{th}} $ & nodes above the 75th percentile of elements in $\mathcal{M}$ \\
		$ \mathbf{Y}^{<75\text{th}} $ & nodes below the 75th percentile of elements in $\mathcal{M}$ \\
		$ \mathbf{Y}^{\text{CA+}} $  &  data points for a server (i.e., nodes generated by CA+) \\
		$ \mathcal{Y}^{\text{CAE$_{\text{FC}}$}} $  &  set of nodes generated by CAE$_{\text{FC}}$ \\
		\hline\hline
	\end{tabular}
	}
	\vspace{-2mm}
\end{table}

\section{Literature Review}
\label{sec:literature}

\subsection{Growing Self-organizing Clustering Algorithms}
\label{sec:literature_GC}
As classical clustering algorithms, GMM \cite{mclachlan19} and $ k $-means \cite{lloyd82} are fundamental and widely used approaches in many fields. Although these algorithms have shown their adaptability and applicability, one well-known drawback exists, namely the number of clusters/partitions has to be pre-specified. To solve this problem, growing self-organizing clustering algorithms such as Growing Neural Gas (GNG) \cite{fritzke95} and Adjusted Self-Organizing Incremental Neural Network (ASOINN) \cite{shen08} have been proposed. GNG and ASOINN adaptively generate topological networks (i.e., nodes and edges) for representing the distributions of given data. SOINN+ \cite{wiwatcharakoses20} is an ASOINN-based algorithm that can handle arbitrary data distributions in noisy data streams without any pre-defined parameters. However, since these algorithms permanently insert new nodes and edges for learning new information, there is a possibility of forgetting previously learned information (i.e., catastrophic forgetting). As a GNG-based algorithm, Grow When Required (GWR) \cite{marsland02} successfully avoids catastrophic forgetting by adding a node only when the state of the current network does not sufficiently match a new instance. One common problem of GWR and SOINN+ is that as the number of nodes in the topological network increases, the cost of calculating a threshold for each node increases, and therefore the learning efficiency decreases.

One promising approach for avoiding catastrophic forgetting is an ART-based algorithm \cite{carpenter91b, vigdor07, wang19, da20}. In particular, algorithms that use CIM as a similarity measure have shown superior clustering performance to other clustering algorithms \cite{masuyama18, masuyama19a, masuyama19b, masuyamaFTCA}. A well-known drawback of ART-based algorithms is the specification of significantly data-dependent parameters such as a similarity threshold (i.e., a vigilance parameter). Several studies have proposed to avoid and/or suppress the effect of the above-mentioned drawback by using multiple vigilance values \cite{da19}, by specifying the vigilance parameter indirectly \cite{da17,masuyama22a}, and by adjusting some data-dependent parameters during the learning process \cite{meng15}. A state-of-the-art parameter-free algorithm is CAE \cite{masuyama23}. In CAE, a similarity threshold is calculated based on pairwise similarities by using well-diversified generated nodes. The diversified nodes are selected by a Determinantal Point Processes (DPP)-based criterion \cite{kulesza12,holder20} incorporating CIM.

\subsection{Privacy-preserving Clustering Algorithms}
\label{sec:literature_PC}
The protection of data privacy is generally realized by cryptograph-, mathematical-, and machine learning-based methods. Secure computation and secret sharing are widely used as cryptograph-based methods \cite{bunn07,sun20,zhang22}. Secure computation is also known as secure multi-party computation, which allows multiple parties to compute a function over their data while keeping those data private. Secret sharing divides sensitive data into multiple parts to preserve data privacy. In general, secure computation is often time-consuming \cite{anju22}, while secret sharing needs to process the creation, distribution, and combination of secret shares \cite{shivhare22}.

As mathematical-based methods, anonymization provides a simple and efficient data privacy protection mechanism \cite{majeed22}. One problem with anonymization is that the original data can be estimated by combining it with specific information. Differential privacy provides mathematically-defined strict privacy protection \cite{dwork06}. Differential privacy protects sensitive information in training data by adding noise to the training data or the output of a trained model. In the clustering domain, local differential privacy is often applied thanks to its simple mechanism and mathematical guarantees for privacy protection \cite{stemmer18,tang19,fioretto21,yang22}.

One recent successful machine learning-based method is federated learning \cite{mcmahan17,li23}. In the clustering domain, an algorithm with a federated learning framework is called federated clustering \cite{ghosh20,liu20,dennis21,pedrycz21}. $k$-FED \cite{dennis21} is a communication-efficient federated clustering algorithm that requires one-shot communication from clients to a server. In $k$-FED, each client performs $k$-means on local data and sends clustering results (i.e., centroids) to the server. The server performs $k$-means on the centroids sent from all clients. Federated Fuzzy $c$-Means (FedFCM) \cite{stallmann22} is an algorithm similar to $k$-FED that utilizes fuzzy $c$-means \cite{bezdek84}. Machine Unlearning Federated Clustering (MUFC) \cite{pan23} introduced a novel sparse compressed multi-set aggregation scheme that satisfies a new privacy criterion. In MUFC, a server only receives the information on the centroids and the number of data points related to each centroid from each client. Thus, the server has no information on each data point.

Most of base clusterers used in existing federated clustering (and also cryptograph- and mathematical-based methods) are centroid-based algorithms such as GMM \cite{pandhare21}, $k$-means \cite{yuan19,shewale19,dennis21,wang22,pan23}, fuzzy $c$-means \cite{pedrycz21,stallmann22}, and spectral clustering \cite{wang20b}. Although centroid-based clustering algorithms are simple and highly applicable, these algorithms require the number of clusters to be specified in advance. In a society with rapidly growing and diverse data, it is difficult to know the true number of clusters of the data a priori. Moreover, there is a possibility that the distributions of data and the number of clusters are frequently changed dynamically. The above-mentioned difficulty in each existing method emphasizes the significance of developing federated clustering algorithms capable of continual learning.

\section{Preliminary Knowledge}
\label{sec:preliminary}
This section provides preliminary knowledge related to FCAC. First, local differential privacy is explained. Next, a similarity measure and a kernel density estimator used in CAE are explained. Last, the learning procedure of CAE is explained in detail.

\subsection{Local Differential Privacy}
\label{sec:localDP}
Local differential privacy includes two approaches: randomized response and adding noise to the data itself \cite{dwork06b,dwork06,dwork14}. In general, the former is applied to discrete values, while the latter is applied to continuous values. Since FCAC is a clustering algorithm, the latter approach is applied, i.e., adding noise to the data itself.

The definition of local differential privacy is as follows. An algorithm $\Psi$ satisfies $\epsilon$-differential privacy ($\epsilon > 0$) if and only if for any data points $\mathbf{x}$ and $\mathbf{x}'$ ($\mathbf{x},\mathbf{x}' \in \mathbb{R}^{d}$), the following relation holds:
\begin{equation}
	\mathrm{Pr}\left[ \Psi(\mathbf{x}) = \mathbf{y} \right] \leq e^{\epsilon} \hspace{1pt} \mathrm{Pr}\left[ \Psi(\mathbf{x}') = \mathbf{y} \right], \ \forall \mathbf{y} \in \mathrm{Range}(\Psi),
	\label{eq:defDP}
\end{equation}
where $\mathrm{Pr}[\cdot]$ is a probability, $\mathrm{Range}(\Psi)$ is every possible output of the algorithm $\Psi$, and $\epsilon$ is a privacy budget. In general, $\epsilon = 0$ means perfect privacy, while $\epsilon = \infty$ means no privacy guarantee.

In this paper, local $\epsilon$-differential privacy is realized by using a Laplace mechanism \cite{dwork06b}. More specifically, the inverse cumulative density function of Laplace distribution \cite{devroye86} is used. Suppose that a set of data points $\mathcal{X}_{c} = \{\mathbf{x}_{1},\mathbf{x}_{2},\ldots,\mathbf{x}_{N}\}$ is given, where $\mathbf{x}_{i} = (x_{i,1},x_{i,2},\ldots,x_{i,d})$. A method of adding a noise to the $j$th dimension of a data point $\mathbf{x}_{i}$ is as follows:
\begin{equation}
	x'_{i,j} = x_{i,j} + L(v_{i,j}) \hspace{8pt} (j = 1,2,\ldots,d),
	\label{eq:noisedData}
\end{equation}
where
\begin{equation}
	L(v_{i,j}) = \mu - \frac{\Delta f_{j}}{\epsilon}\text{sgn}(v_{i,j})\ln(1 - 2|v_{i,j}|),
	\label{eq:ICDF}
\end{equation}
here, $v_{i,j}$ is the random variable sampled from a uniform distribution $U(-0.5, 0.5)$, $\text{sgn}(\cdot)$ is a signum function, $\ln(\cdot)$ is a natural logarithm function, $\mu$ is the mean of a Laplace distribution, and $\Delta f_{j}$ is a local sensitivity of differential privacy for $x_{i,j}$.

The local sensitivity $\Delta f_{j}$ is defined as follows:
\begin{equation}
	\Delta f_{j} = \max_{n = 1,2,\ldots,N} (x_{n,j}) - \min_{n = 1,2,\ldots,N} (x_{n,j}).
	\label{eq:sensitivityDP}
\end{equation}

\subsection{Correntropy and Correntropy-induced Metric}
\label{sec:cimDefinition}
Correntropy \cite{liu07} provides a generalized similarity measure between two arbitrary data points $ \mathbf{x} = (x_{1},x_{2},\ldots,x_{d}) $ and $ \mathbf{y} = (y_{1},y_{2},\ldots,y_{d}) $ as follows:
\begin{equation}
	C(\mathbf{x}, \mathbf{y}) = \textbf{E} \left[ \kappa_{\sigma} \left( \mathbf{x}, \mathbf{y} \right) \right],
\end{equation}
where $ \textbf{E} \left[ \cdot \right] $ is the expectation operation, and $ \kappa_{\sigma} \left( \cdot \right) $ denotes a positive definite kernel with a bandwidth $ \sigma $. The correntropy is estimated as follows:
\begin{equation}
	\hat{C}(\mathbf{x}, \mathbf{y}, \sigma) = \frac{1}{d} \sum_{i=1}^{d} \kappa_{\sigma} \left( x_{i}, y_{i} \right).
	\label{eq:correntropy}
\end{equation}

In this paper, we use the following Gaussian kernel in the correntropy:
\begin{equation}
	\kappa_{\sigma} \left( x_{i}, y_{i} \right) = \exp \left[ - \frac{\left( x_{i} - y_{i} \right)^{2} }{2 \sigma^{2}} \right].
	\label{eq:gaussian}
\end{equation}

A nonlinear metric called CIM is derived from the correntropy \cite{liu07}. CIM quantifies the similarity between two data points $ \mathbf{x} $ and $ \mathbf{y} $ as follows:
\begin{equation}
	\mathrm{CIM}\left(\mathbf{x}, \mathbf{y}, \sigma \right) = \left[ 1 - \hat{C}(\mathbf{x}, \mathbf{y}, \sigma) \right]^{\frac{1}{2}},
	\label{eq:defcim}
\end{equation}
here, since the Gaussian kernel in (\ref{eq:gaussian}) does not have the coefficient $ \frac{1}{\sqrt{2\pi}\sigma} $, the range of CIM is limited to $ \left[0,1 \right] $.

In general, the Euclidean distance suffers from the curse of dimensionality. However, CIM reduces this drawback since the correntropy calculates the similarity between two data points by using a kernel function. Moreover, it has also been shown that CIM with the Gaussian kernel has a high outlier rejection ability \cite{liu07}.

\subsection{Kernel Density Estimator}
\label{sec:kde}
In general, the bandwidth of a kernel function can be estimated from $ \lambda $ instances belonging to a certain distribution \cite{henderson12}, which is defined as follows:
\begin{equation}
\bm{\Sigma} = U(F_{\nu})  \bm{\Gamma} \lambda^{-\frac{1}{2\nu+d}},
\label{eq:sigmaEst1}
\end{equation}
\begin{equation}
U(F_{\nu}) = \left( \frac{\pi ^{d/2} 2^{d+\nu-1}(\nu !)^{2}R(F)^{d}}{\nu \kappa_{\nu}^{2}(F)\left[(2\nu)!!+(d-1)(\nu!!)^{2}\right]} \right)^{\frac{1}{2\nu+d}},
\label{eq:sigmaEst2}
\end{equation}
where $ \bm{\Gamma} $ denotes a rescale operator ($ d $-dimensional vector) which is defined by a standard deviation of each of the $ d $ attributes among $ \lambda $ instances, $ \nu $ is the order of a kernel, the single factorial of $\nu$ is calculated by the product of integer numbers from 1 to $\nu$, the double factorial notation is defined as $ (2\nu)!!  = (2\nu-1) \cdot 5 \cdot 3 \cdot 1 $ (commonly known as the odd factorial), $ R(F) $ is a roughness function, and $ \kappa_{\nu}(F) $ is the moment of a kernel. The details of the derivation of (\ref{eq:sigmaEst1}) and (\ref{eq:sigmaEst2}) can be found in \cite{henderson12}. In this paper, we use the Gaussian kernel for CIM. Therefore, $ \nu = 2 $, $ R(F) = (2\sqrt{\pi})^{-1} $, and $ \kappa_{\nu}^{2}(F) = 1 $. Then, (\ref{eq:sigmaEst2}) is rewritten as follows:
\begin{equation}
	\mathbf{H} = \left( \frac{4}{2+d} \right)^{\frac{1}{4+d}}  \bm{\Gamma}  \lambda^{-\frac{1}{4+d}}.
	\label{eq:SIGMA}
\end{equation}

Equation (\ref{eq:SIGMA}) is known as the Silverman's rule \cite{silverman18}. Here, $ \mathbf{H} $ contains the bandwidth of a kernel function in CIM.


\subsection{CIM-based ART with Edge: CAE}
\label{sec:cae}
CAE is a parameter-free ART-based topological clustering algorithm capable of continual learning \cite{masuyama23}. In general, ART-based algorithms have a data-dependent parameter such as a vigilance parameter (similarity threshold). In CAE, a similarity threshold is calculated based on a pairwise similarities among a certain number of nodes. The sufficient number of nodes for calculating the similarity threshold is estimated by a Determinantal Point Processes (DPP)-based criterion \cite{kulesza12,holder20}. In addition, an edge deletion threshold is estimated based on the age of each edge, which is inspired by the edge deletion mechanism of SOINN+ \cite{wiwatcharakoses20}. Empirical studies with the synthetic and real-world datasets showed that the clustering performance of CAE is superior to state-of-the-art parameter-free/fixed algorithms \cite{masuyama23}.

The following sections provide the learning processes of CAE step by step. Algorithm \ref{alg:pseudocodeCAE} summarizes the entire learning procedure of CAE.

\begin{algorithm*}[htbp]
	\DontPrintSemicolon
	\KwIn{\\
		a set of data points $ \mathcal{X}_{c} $
	}
	\KwOut{\\
		a set of nodes $\mathcal{Y}$\\
		a set of winning counts $\mathcal{M}$\\
		a set of ages of edges $\mathcal{E}$ \\
	}
	\While{
		\normalfont{existing data points to be trained}
	}
	{
		Input a data point $\mathbf{x}$ ($\mathbf{x} \in \mathcal{X}_{c}$).\\

		\uIf{\normalfont{the number of active nodes $\lambda$ is not defined \textbf{or} the number of nodes $|\mathcal{Y}|$ is smaller than $\lambda/2$ \textbf{or} a similarity threshold $V_{\text{threshold}}$ is not calculated}}
		{
			Create a node as $\mathbf{y}_{|\mathcal{Y}|+1} = \mathbf{x}$, and update a set of nodes as $ \mathcal{Y} \leftarrow \mathcal{Y} \cup \{ \mathbf{y}_{|\mathcal{Y}|+1} \} $. \\
			\tcc{Estimation of Diversity of Nodes}
			Calculate a pairwise similarity matrix $\mathbf{R}$.  \tcp*{(\ref{eq:SimilarityMat})}
			Calculate the diversity as $D = \det( \mathbf{R} )$. \tcp*{(\ref{eq:div})}
			\uIf{$ D <  1.0\mathrm{e}{-6}$}
			{
				$\lambda = 2|\mathcal{A}|$ \\
				\tcc{Calculation of Similarity Threshold}
				Calculate a similarity threshold $V_{\text{threshold}}$.\\
			}
			\Else{
				$\lambda = \infty$
			}
		}
		\Else
		{
			Select the 1st and 2nd nearest nodes from $\mathbf{x}$ (i.e., $\mathbf{y}_{s_{1}}$ and $\mathbf{y}_{s_{2}}$) based on CIM. \tcp*{(\ref{eq:winnerCIM1}),(\ref{eq:winnerCIM2})}
			Calculate similarities between $ \mathbf{x} $ and $\mathbf{y}_{s_{1}}$, $\mathbf{y}_{s_{2}}$ (i.e., $V_{s_{1}}$ and $V_{s_{2}}$). \tcp*{(\ref{eq:cim1}),(\ref{eq:cim2})}
			\tcc{Vigilance Test, and Creation/Update of Nodes and Edges}
			\uIf{$V_{\mathrm{threshold}} < V_{s_{1}}$}
			{
				\tcc{Case I}
				$\mathbf{y}_{|\mathcal{Y}|+1} = \mathbf{x}$, and $ \mathcal{Y} \leftarrow \mathcal{Y} \cup \{ \mathbf{y}_{|\mathcal{Y}|+1} \} $ \\
				$M_{|\mathcal{Y}|+1} = 1$, and $ \mathcal{M} \leftarrow \mathcal{M} \cup \{ M_{|\mathcal{Y}|+1} \} $ \\
				Calculate $ \sigma_{|\mathcal{Y}|+1} $ by (\ref{eq:SIGMA}) and (\ref{eq:sigma}) with the active node set $\mathcal{A}$, and $ \mathcal{S} \leftarrow \mathcal{S} \cup \{ \sigma_{|\mathcal{Y}|+1} \} $. \\
				Update the active node set $\mathcal{A}$. 
			}
			\Else
			{
				\tcc{Case I\hspace{-.1em}I}
				$ M_{s_1} \leftarrow M_{s_1} + 1$ \tcp*{(\ref{eq:countNode})}
				$\mathbf{y}_{s_{1}} \leftarrow \mathbf{y}_{s_{1}} + \frac{1}{M_{s_1}}( \mathbf{x} - \mathbf{y}_{s_{1}})$. \tcp*{(\ref{eq:updateNodeWeight1})}
				Update the active node set $\mathcal{A}$ \\
				\For{$\mathbf{y}_{k} \in \mathcal{N}_{s_{1}}$}
				{
					$a{(\mathbf{y}_{s_{1}},\mathbf{y}_{k})} \leftarrow a{(\mathbf{y}_{s_{1}},\mathbf{y}_{k})} + 1$ \tcp*{(\ref{eq:edd_age})}
				}

				\If{$V_{s_{2}} \leq V_{\mathrm{threshold}}$}
				{
					\tcc{Case I\hspace{-.1em}I\hspace{-.1em}I}
					$a(\mathbf{y}_{s_{1}},\mathbf{y}_{s_{2}}) \leftarrow 1$ \tcp*{(\ref{eq:ageInit})}
					\For{$\mathbf{y}_{k} \in \mathcal{N}_{s_{1}}$}
					{
						$\mathbf{y}_{k}  \leftarrow \mathbf{y}_{k}  + \frac{1}{10M_{k}}( \mathbf{x}- \mathbf{y}_{k})$ \tcp*{(\ref{eq:updateNodeWeight2})}
					}
				}
			}
			\tcc{Estimation of Edge Deletion Threshold}
			$\alpha \leftarrow $ a set of ages of edges which connect to $\mathbf{y}_{s_{1}}$ \\
			$\alpha_{0.75} \leftarrow$ the 75th percentile of elements in $\alpha$ \\
			$a_{\text{thr}} =\alpha_{0.75} + \text{IQR}(\alpha)$. \tcp*{(\ref{eq:outliers})}
			$a_{\text{max}} = \bar{\alpha}_{\text{del}} \frac{|\alpha_{\text{del}}|}{|\alpha_{\text{del}}|+|\alpha|} + a_{\text{thr}}  \left(1 \!-\!  \frac{|\alpha_{\text{del}}|}{|\alpha_{\text{del}}|+|\alpha|} \right)$
			\tcp*{(\ref{eq:CulA_max})}
			\tcc{Deletion of Edges}
			\For{$\mathbf{y}_{k} \in \mathcal{N}_{s_{1}}$}{
			\If{$a(\mathbf{y}_{s_{1}},\mathbf{y}_{k}) > a_{\text{max}}$}
			{
				$\alpha_{\text{del}} \leftarrow \alpha_{\text{del}} \cup \{ a(\mathbf{y}_{s_{1}},\mathbf{y}_{k}) \}$ \\
				Delete the edge between $\mathbf{y}_{s_{1}}$ and $\mathbf{y}_{k}$.
			}
	}
		}
		\If{\normalfont{the number of presented data points is a multiple of $\lambda$}}
		{
			Delete isolated nodes.
		}
	}
	\caption{Learning procedure of CAE}
	\label{alg:pseudocodeCAE}
\end{algorithm*}

\subsubsection{Estimation of Diversity of Nodes}
\label{sec:estDiversity}

In CAE, a similarity threshold is defined by a pairwise similarities among nodes (i.e., $\mathcal{Y}$). Therefore, the diversity of nodes for calculating the similarity threshold is important to obtain an appropriate threshold value, which leads to good clustering performance.

The diversity $D$ of the active node set $\mathcal{A}$ is estimated by a DPP-based criterion \cite{kulesza12,holder20} incorporating CIM as follows:
\begin{equation}
	D =  \det( \mathbf{R} ),
	\label{eq:div}
\end{equation}
where
\begin{equation}
	\mathbf{R} =  \left[ \exp\left(1 - \mathrm{CIM}(\mathbf{y}_{i}, \mathbf{y}_{j}, \sigma)\right) \right]_{1 \leq i, j \leq |\mathcal{A}|}.
	\label{eq:SimilarityMat}
\end{equation}

Here, $\det( \mathbf{R} )$ is the determinant of the matrix $\mathbf{R}$, and $\mathbf{R}$ is a matrix of pairwise similarities between nodes in $\mathcal{A}$. A bandwidth $\sigma$ for CIM is calculated from $ \mathbf{H} $ in (\ref{eq:SIGMA}) by using the node set $\mathcal{A}$. As in (\ref{eq:SIGMA}), $ \mathbf{H} $ contains the bandwidth of a kernel function in CIM. In this paper, the median of $ \mathbf{H} $ is used as the bandwidth of the Gaussian kernel in CIM, i.e.,
\begin{equation}
	\sigma = \mathrm{median} \left( \mathbf{H} \right).
	\label{eq:sigma}
\end{equation}

In general, the diversity $D = 0$ means that the node set $\mathcal{A}$ is not diverse while $D > 0$ means $\mathcal{A}$ is diverse. In other words, the value of $D$ becomes close to zero when a new node is created around the existing nodes.

In CAE, the value of $\lambda$ is set as the two times of the number of nodes (i.e., $2|\mathcal{Y}|$) when the diversity $D$ satisfies $D < 1.0\mathrm{e}{-6}$. If the number of nodes becomes smaller than $\lambda/2$ after a node deletion process, $\lambda$ is calculated again.

As shown in lines 3-4 of Algorithm \ref{alg:pseudocodeCAE}, the first $\lambda/2$ data points (i.e., $\{ \mathbf{x}_{1}, \mathbf{x}_{2},\ldots \mathbf{x}_{\lambda/2} \}$) directly become nodes, i.e., $ \mathcal{Y} = \{\mathbf{y}_{1}, \mathbf{y}_{2}, \ldots, \mathbf{y}_{\lambda/2}\} $ where $ \mathbf{y}_{k} = \mathbf{x}_{k} $ $ ( k = 1, 2, \ldots, \lambda/2 ) $. In addition, the bandwidth for the Gaussian kernel in CIM is assigned to each node, i.e., $ \mathcal{S} = \{ \sigma_{1},\sigma_{2},\ldots, \sigma_{\lambda/2}\} $ where $ \sigma_{1} = \sigma_{2} = \cdots = \sigma_{\lambda/2} $. 

The value of $\lambda$ is automatically updated in the proposed CAE algorithm. In an active node set $\mathcal{A}$, $\lambda$ nodes are stored. When a new node is added to $\mathcal{A}$, an old node is removed to maintain the active node set size as $\lambda$. The addition of a new node and the removal of an old node are explained later.

\subsubsection{Calculation of Similarity Threshold}
\label{sec:calSimilarity}
The similarity threshold $ V_{\text{threshold}} $ is calculated by the average of the minimum pairwise CIM values in the active node set $ \mathcal{A}$ as follows:
\begin{equation}
	V_{\text{threshold}} = \frac{1}{\lambda} \sum_{\mathbf{y}_{i} \in \mathcal{A}} \min_{\mathbf{y}_{j} \in \mathcal{A} \setminus \mathbf{y}_{i}} \left[\mathrm{CIM}\left(\mathbf{y}_{i}, \mathbf{y}_{j}, \mathrm{mean}(\mathcal{S})\right)\right],
	\label{eq:pairwiseCIM}
\end{equation}
where $ \mathcal{S} $ is a set of bandwidths of the Gaussian kernel in CIM for $ \mathcal{A} $. The bandwidth of each node in $ \mathcal{A} $ is calculated by using (\ref{eq:SIGMA}) and (\ref{eq:sigma}) when a new node is created.

\subsubsection{Selection of Winner Nodes}
\label{sec:winnerNodes}

During the learning process of CAE, every time a data point $ \mathbf{x} $ is given, two nodes that have a similar state to $ \mathbf{x} $ are selected from $\mathcal{Y}$, namely the 1st winner node $ \mathbf{y}_{s_{1}} $ and the 2nd winner node $ \mathbf{y}_{s_{2}} $. The winner nodes are determined based on the value of CIM in line 14 of Algorithm \ref{alg:pseudocodeCAE} as follows:
\begin{equation}
	\hspace{-4.5mm} s_{1} = \argmin_{\mathbf{y}_{i} \in \mathcal{Y}}\left[ \mathrm{CIM}\left(\mathbf{x}, \mathbf{y}_{i}, \mathrm{mean}(\mathcal{S}) \right) \right],
	\label{eq:winnerCIM1}
\end{equation}
\vspace{-5pt}
\begin{equation}
	s_{2} = \argmin_{\mathbf{y}_{i} \in \mathcal{Y} \backslash \{\mathbf{y}_{s_{1}}\}}\left[ \mathrm{CIM}\left(\mathbf{x}, \mathbf{y}_{i}, \mathrm{mean}(\mathcal{S}) \right) \right],
	\label{eq:winnerCIM2}
\end{equation}
\noindent where $ s_{1} $ and $ s_{2} $ denote the indexes of the 1st and 2nd winner nodes, respectively. $ \mathcal{S} = \{ \sigma_{1},\sigma_{2},\ldots,\sigma_{|\mathcal{Y}|} \} $ is a set of bandwidths of the Gaussian kernel in CIM corresponding to a node set $ \mathcal{Y} $.

Note that the 1st winner node $\mathbf{y}_{s_{1}}$ becomes a new active node, and the oldest node in the active node set $\mathcal{A}$ (i.e., $\lambda$ nodes in $\mathcal{Y}$) is replaced by the new one.

\subsubsection{Vigilance Test}
\label{sec:vigilanceTest}
Similarities between the data point $ \mathbf{x} $ and each of the 1st and 2nd winner nodes are defined in lines 14-15 of Algorithm \ref{alg:pseudocodeCAE} as follows:
\begin{equation}
	V_{s_{1}} =  \mathrm{CIM}\left(\mathbf{x}, \mathbf{y}_{s_{1}}, \mathrm{mean}(\mathcal{S}) \right),
	\label{eq:cim1}
\end{equation}
\vspace{-8pt}
\begin{equation}
	V_{s_{2}} = \mathrm{CIM}\left(\mathbf{x}, \mathbf{y}_{s_{2}}, \mathrm{mean}(\mathcal{S}) \right).
	\label{eq:cim2}
\end{equation}

The vigilance test classifies the relationship between the data point $\mathbf{x}$ and the two winner nodes into three cases by using the similarity threshold $ V_{\text{threshold}} $, i.e.,

\begin{itemize}
	[ 
	\setlength{\IEEElabelindent}{\dimexpr-\labelwidth-\labelsep}
	\setlength{\itemindent}{\dimexpr\labelwidth+\labelsep}
	\setlength{\listparindent}{\parindent}
	] 
	
	\item Case I \\
	\indent The similarity between $ \mathbf{x} $ and the 1st winner node $ \mathbf{y}_{s_{1}} $ is larger (i.e., less similar) than $ V_{\text{threshold}} $, namely:
	\begin{equation}
		V_{\text{threshold}} < V_{s_{1}} \leq V_{s_{2}}.
		\label{eq:case1}
	\end{equation}
	
	\vspace{0.8mm}
	
	\item Case I\hspace{-.1em}I \\
	\indent The similarity between $ \mathbf{x} $ and the 1st winner node $ \mathbf{y}_{s_{1}} $ is smaller (i.e., more similar) than $ V_{\text{threshold}} $, and the similarity between $ \mathbf{x} $ and the 2nd winner node $ \mathbf{y}_{s_{2}} $ is larger (i.e., less similar) than $ V_{\text{threshold}} $, namely:
	\begin{equation}
		V_{s_{1}} \leq V_{\text{threshold}} < V_{s_{2}}.
		\label{eq:case2}
	\end{equation}
	
	\vspace{0.8mm}
	
	\item Case I\hspace{-.1em}I\hspace{-.1em}I \\
	\indent The similarities between $ \mathbf{x} $ and the 1st and 2nd winner nodes (i.e., $ \mathbf{y}_{s_{1}} $ and $ \mathbf{y}_{s_{2}} $) are both smaller (i.e., more similar) than $ V_{\text{threshold}} $, namely:
	\begin{equation}
		V_{s_{1}} \leq V_{s_{2}} \leq V_{\text{threshold}}.
		\label{eq:case3}
	\end{equation}
	
\end{itemize}

\subsubsection{Creation/Update of Nodes and Edges}
\label{sec:update}
Depending on the result of the vigilance test, a different operation is performed.

If the data point $ \mathbf{x} $ is classified as Case I by the vigilance test (i.e., (\ref{eq:case1}) is satisfied), a new node is created as $\mathbf{y}_{|\mathcal{Y}|+1} = \mathbf{x}$, and updated a node set as $ \mathcal{Y} \leftarrow \mathcal{Y} \cup \{ \mathbf{y}_{|\mathcal{Y}|+1} \} $. Here, the node $\mathbf{y}_{|\mathcal{Y}|+1}$ becomes a new active node, and the oldest node in the active node set $\mathcal{A}$ (i.e., $\lambda$ nodes in $\mathcal{Y}$) is replaced by the new one. In addition, a bandwidth $ \sigma_{|\mathcal{Y}|+1} $ for $ \mathbf{y}_{|\mathcal{Y}|+1} $ is calculated by (\ref{eq:SIGMA}) and (\ref{eq:sigma}) with the active node set $\mathcal{A}$, and the winning count of $ \mathbf{y}_{|\mathcal{Y}|+1} $ is initialized as $ M_{|\mathcal{Y}|+1} = 1 $.

If the data point $ \mathbf{x} $ is classified as Case I\hspace{-1pt}I by the vigilance test (i.e., (\ref{eq:case2}) is satisfied), first, the winning count of $ \mathbf{y}_{s_{1}} $ is updated as follows:
\begin{equation}
	M_{s_{1}} \leftarrow M_{s_{1}} + 1.
	\label{eq:countNode}
\end{equation}

Then, $ \mathbf{y}_{s_{1}} $ is updated as follows:
\begin{equation}
	\mathbf{y}_{s_{1}} \leftarrow \mathbf{y}_{s_{1}} + \frac{1}{M_{s_{1}}} \left( \mathbf{x} - \mathbf{y}_{s_{1}} \right),
	\label{eq:updateNodeWeight1}
\end{equation}
here, the node $\mathbf{y}_{s_{1}}$ becomes a new active node, and the oldest node in the active node set $\mathcal{A}$ (i.e., $\lambda$ nodes in $\mathcal{Y}$) is replaced by the new one.

When updating the node, the difference between $ \mathbf{x} $ and $ \mathbf{y}_{s_{1}} $ is divided by $ M_{s_{1}} $. Thus, the change of the node position is smaller when $ M_{s_{1}} $ is larger. This is because the information around the node, where data points are frequently given, is important and should be held by the node.

The age of each edge connected to the 1st winner node $ \mathbf{y}_{s_{1}} $ is also updated as follows:
\begin{equation}
	a{(\mathbf{y}_{s_{1}},\mathbf{y}_{k})} \leftarrow a{(\mathbf{y}_{s_{1}},\mathbf{y}_{k})} + 1 \quad (\mathbf{y}_{k} \in \mathcal{N}_{s_{1}}),
	\label{eq:edd_age}
\end{equation}
where $ \mathcal{N}_{s_{1}} $ is a set of all neighbor nodes of the node $ \mathbf{y}_{s_{1}} $.

If the data point $ {\mathbf{x}} $ is classified as Case I\hspace{-1pt}I\hspace{-1pt}I by the vigilance test (i.e., (\ref{eq:case3}) is satisfied), the same operations as Case I\hspace{-1pt}I (i.e., (\ref{eq:countNode})-(\ref{eq:edd_age})) are performed. In addition, if there is an edge between $ \mathbf{y}_{s_{1}} $ and $ \mathbf{y}_{s_{2}} $, an age of the edge is reset as follows:
\begin{equation}
	a{(\mathbf{y}_{s_{1}},\mathbf{y}_{s_{2}})} \leftarrow 1.
	\label{eq:ageInit}
\end{equation} 

In the case that there is no edge between $ \mathbf{y}_{s_{1}} $ and $ \mathbf{y}_{s_{2}} $, a new edge is defined with an age of the edge by (\ref{eq:ageInit}).

After updated the edge information, the neighbor nodes of $ \mathbf{y}_{s_{1}} $ are updated as follows:
\begin{equation}
	{\mathbf{y}}_{k} \gets {\mathbf{y}}_{k} + \frac{1}{10 M_{k}} \left({\mathbf{x}}-{\mathbf{y}}_{k}\right) \quad \left(\mathbf{y}_{k} \in \mathcal{N}_{s_{1}}\right).
	\label{eq:updateNodeWeight2}
\end{equation}

Apart from the above operations in Cases I-I\hspace{-1pt}I\hspace{-1pt}I, the nodes with no edges are deleted (and removed from the active node set $\mathcal{A}$) every $ \lambda $ data points for the noise reduction purpose (i.e., the node deletion interval is the presentation of $ \lambda $ data points), which is performed in lines 39-40 of Algorithm \ref{alg:pseudocodeCAE}.

With respect to the active node set $\mathcal{A}$, its update rules are summarized as follows. In Case I, a new node is directly created by the data point $ {\mathbf{x}} $ and added to $\mathcal{A}$. In Case I\hspace{-1pt}I and Case I\hspace{-1pt}I\hspace{-1pt}I, the updated winner node in (\ref{eq:updateNodeWeight1}) is added to $\mathcal{A}$. In all cases, the oldest active node is removed from $\mathcal{A}$. Then, in lines 39-40 of Algorithm \ref{alg:pseudocodeCAE}, all active nodes with no edges are removed. After this removal procedure, the number of active nodes can be smaller than $\lambda$.

\subsubsection{Estimation of Edge Deletion Threshold}
\label{sec:EstEdgeThreshold}

CAE estimates an edge deletion threshold based on the ages of the current edges and the deleted edges, which is inspired by the edge deletion mechanism of SOINN+ \cite{wiwatcharakoses20}.

The edge deletion threshold $a_{\text{max}}$ is defined as follows:
\begin{equation}
	a_{\text{max}} = \bar{\alpha}_{\text{del}} \frac{|\alpha_{\text{del}}|}{|\alpha_{\text{del}}|+|\alpha|} + a_{\text{thr}} \left(1-\frac{|\alpha_{\text{del}}|}{|\alpha_{\text{del}}|+|\alpha|} \right),
	\label{eq:CulA_max}
\end{equation}
where $\alpha_{\text{del}}$ is the set of ages of all the deleted edges during the learning process, $|\alpha_{\text{del}}|$ is the number of elements in $\alpha_{\text{del}}$, $\bar{\alpha}_{\text{del}}$ is the arithmetric mean of $\alpha_{\text{del}}$, $\alpha$ is the set of ages of edges which connect to $\mathbf{y}_{s_{1}}$ ($\alpha \subset \mathcal{E}$), and $|\alpha|$ is the number of elements in $\alpha$. The coefficient $a_{\text{thr}}$ is defined as follows:
\begin{equation}
	a_{\text{thr}} =\alpha_{0.75} + \text{IQR}(\alpha),
	\label{eq:outliers}
\end{equation}
where $\alpha_{0.75}$ is the 75th percentile of elements in $\alpha$, and $\text{IQR}(\alpha)$ is the interquartile range.

The edge deletion threshold $a_{\text{max}}$ is updated each time the age of an edge increases, which is performed in lines 31-34 of Algorithm \ref{alg:pseudocodeCAE}.

\subsubsection{Deletion of Edges}
\label{sec:deleteEdge}

If there is an edge whose age is greater than the edge deletion threshold $a_{\text{max}}$, the edge is deleted and the set of ages of deleted edges $\alpha_{\text{del}}$ is updated, which are performed in lines 35-38 of Algorithm \ref{alg:pseudocodeCAE}.

\section{Proposed Algorithms}
\label{sec:proposedAlgorithm}

This section explains the proposed algorithms in detail: CAE with a minor modification (i.e., CAE$_{\text{FC}}$), CA+, and FCAC.

\subsection{CAE$_{\text{FC}}$}
\label{sec:cae_fc}

Although CAE \cite{masuyama23} is a state-of-the-art parameter-free ART-based topological clustering algorithm, CAE tends to generate a large number of nodes. The main reason for this phenomenon is that the diversity $D$ of the node set $\mathcal{A}$ defined in (\ref{eq:div}) is unlikely to be $D < 1.0\mathrm{e}{-6}$. In other words, the pairwise similarity matrix $\mathbf{R}$ defined in (\ref{eq:SimilarityMat}) is not appropriate in the case where a large number of nodes is not preferable. In CAE$_{\text{FC}}$, a correntropy-based pairwise similarity matrix $\mathbf{R}$ is used, which is defined as follows:
\begin{equation}
	\mathbf{R} =  \left[ \exp\left(\hat{C}(\mathbf{y}_{i}, \mathbf{y}_{j}, \sigma)\right) \right]_{1 \leq i, j \leq |\mathcal{A}|},
	\label{eq:SimilarityMat_corr}
\end{equation}
where $\hat{C}(\cdot)$ is correntropy which is defined in (\ref{eq:correntropy}).

\begin{figure}[htbp]
	\vspace{-4mm}
	\centering
	\subfloat[Dataset]{
		\includegraphics[width=1.4in]{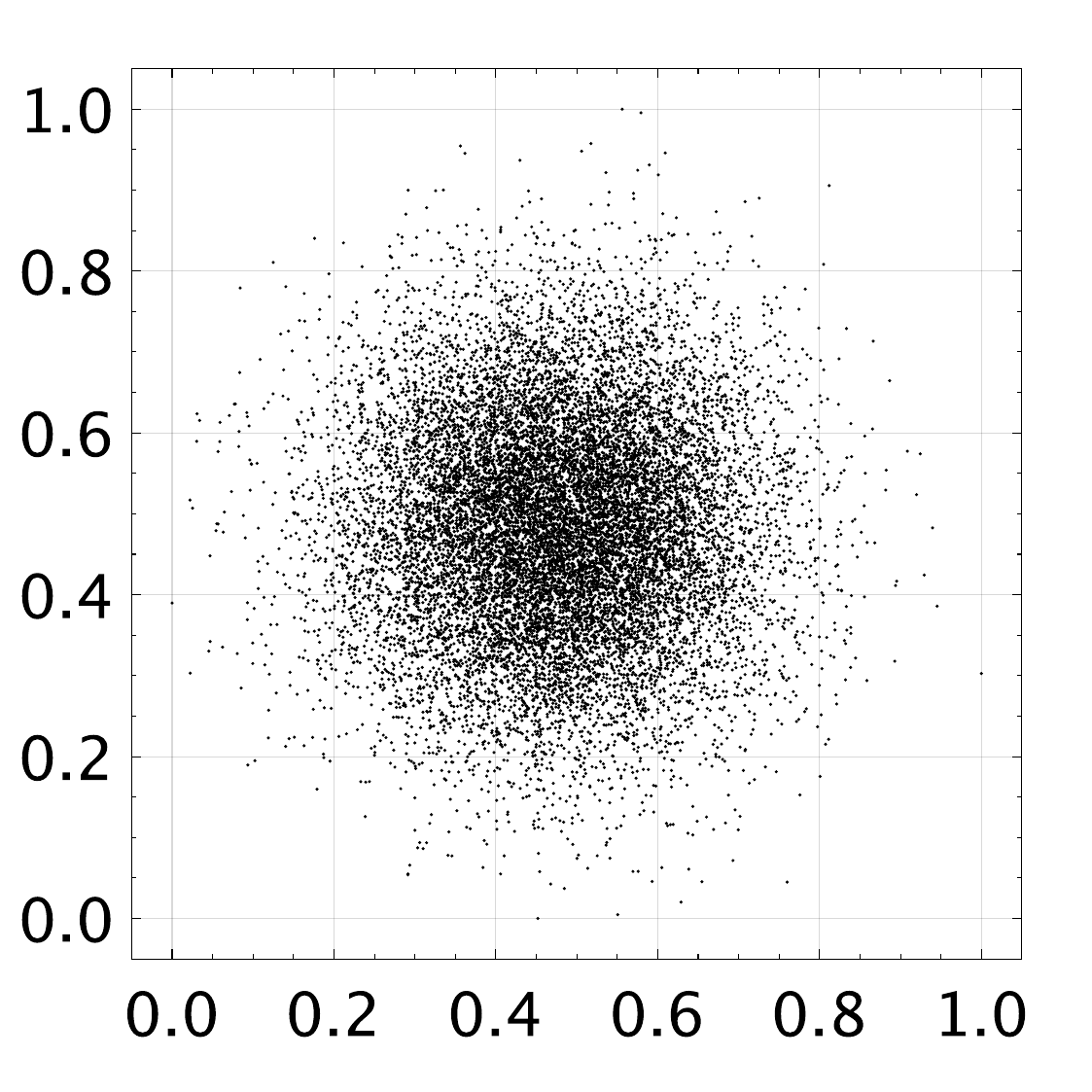}
		\label{fig:comparison_data}
	}\hfil
	\subfloat[CAE]{
		\includegraphics[width=1.4in]{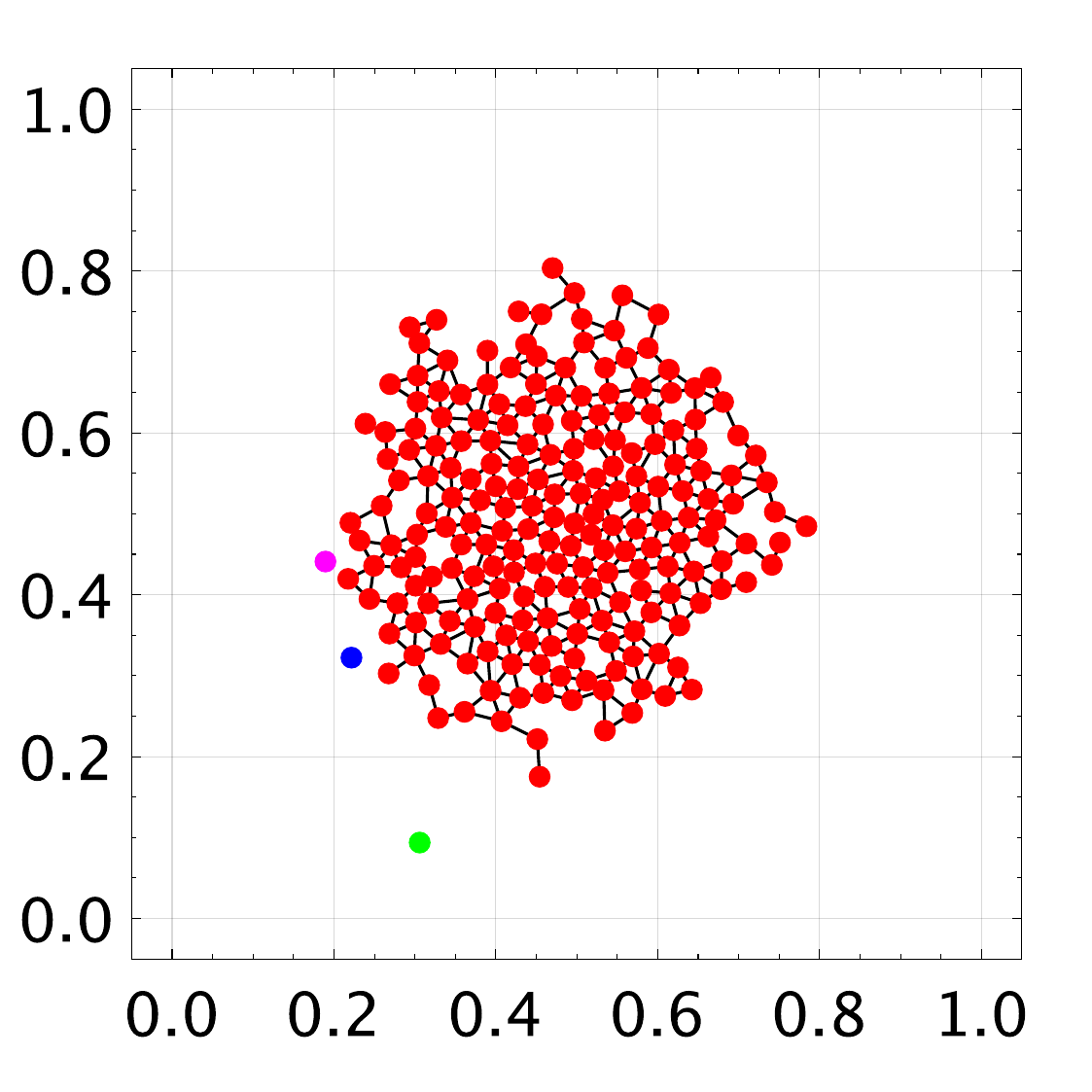}
		\label{fig:comparison_cae}
	}\hfil \\ \vspace{-2mm}
	\subfloat[CAE$_{\text{FC}}$]{
		\includegraphics[width=1.4in]{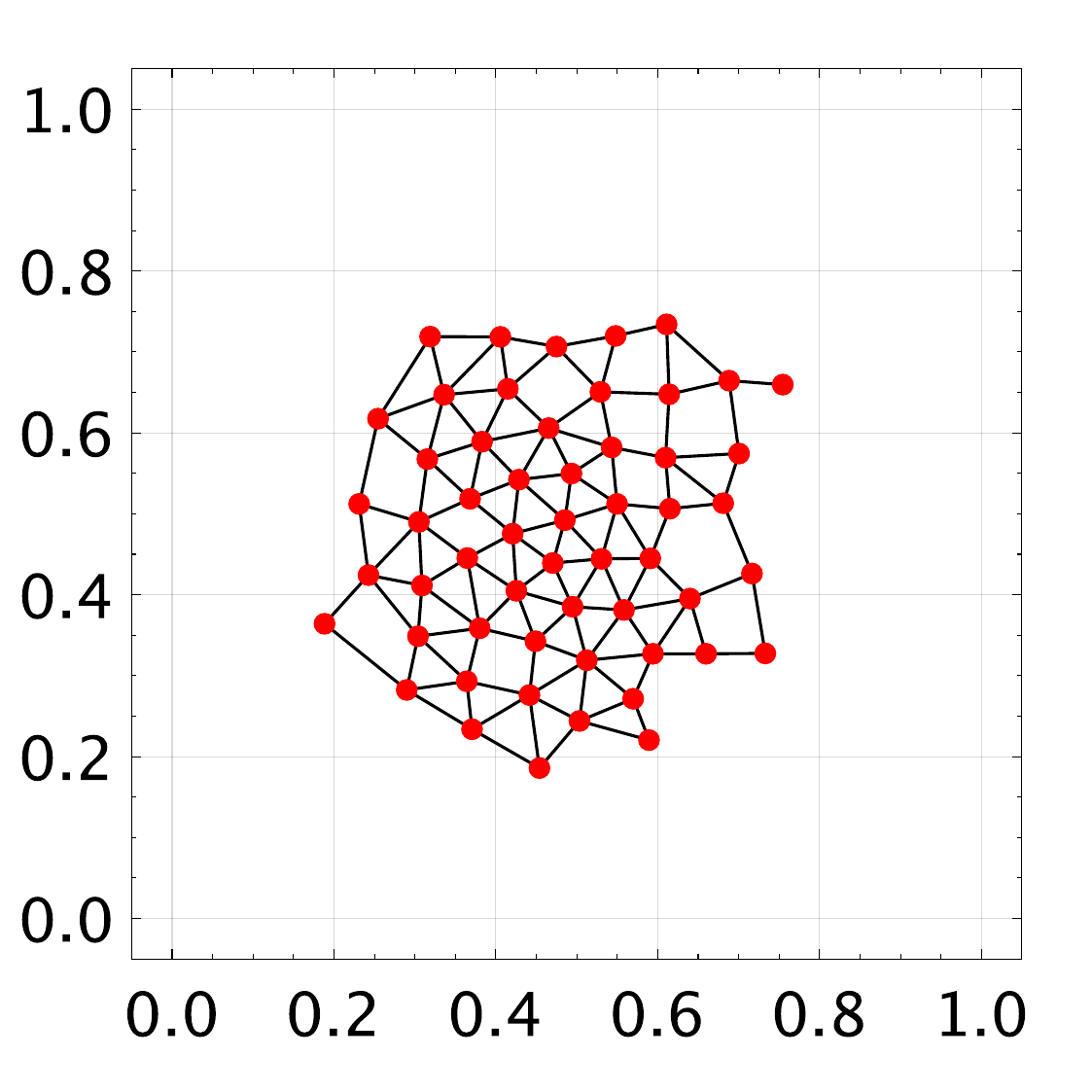}
		\label{fig:comparison_caefc}
	}\hfil
	\subfloat[CA+]{
		\includegraphics[width=1.4in]{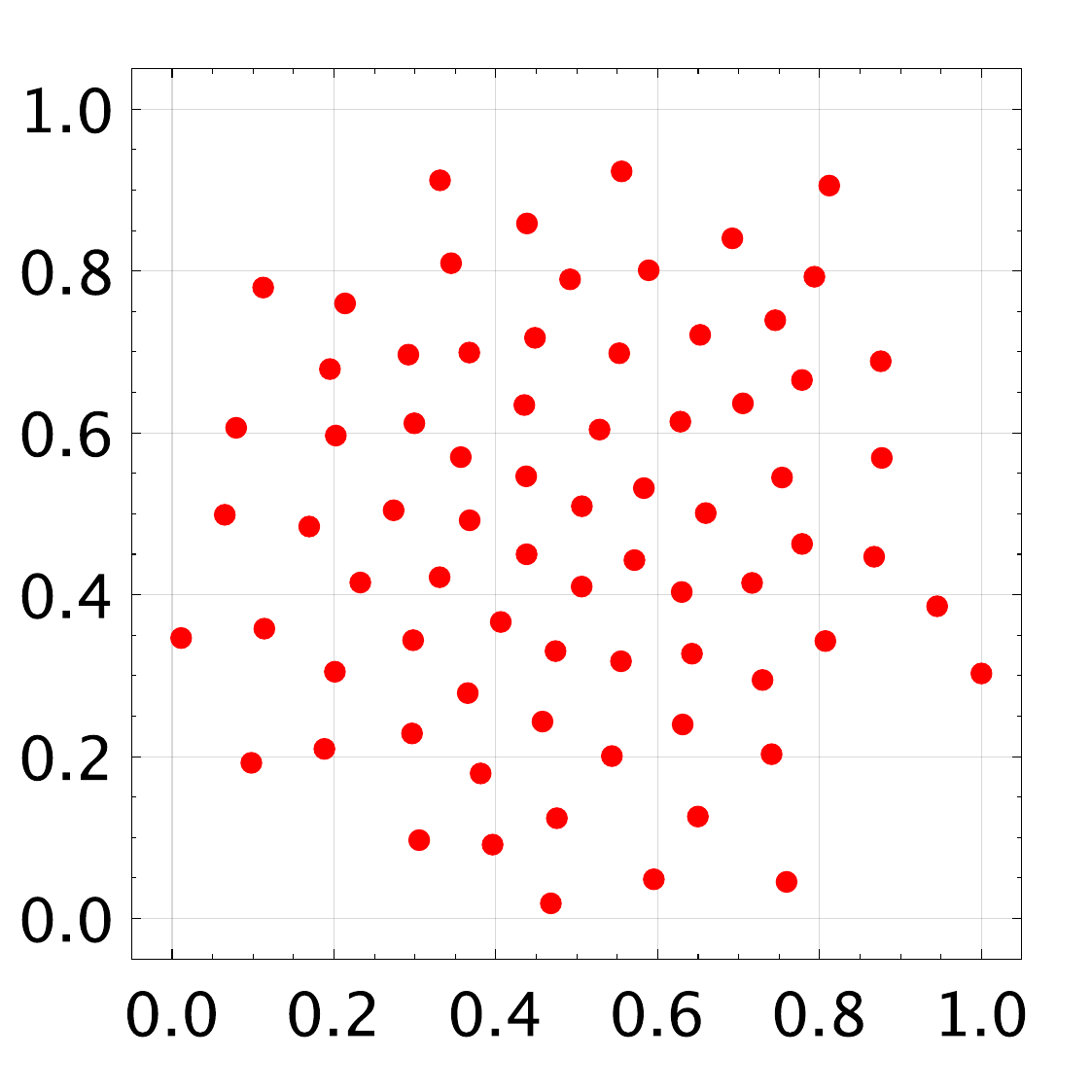}
		\label{fig:comparison_ca_plus}
	}
	\caption{Qualitative comparison of CAE, CAE$_{\text{FC}}$, and CA+.}
	\label{fig:comparison_cae_ca_plus}
\end{figure}

In comparison to the definition in (\ref{eq:SimilarityMat}), the definition in equation (\ref{eq:SimilarityMat_corr}) tends to have a larger difference in the values of the elements of $\mathbf{R}$. As a result, the value of $D$ is close to zero (i.e., $D < 1.0\mathrm{e}{-6}$) when the diversity of the node set $\mathcal{A}$ is small.

The rest of the learning procedure of CAE$_{\text{FC}}$ is completely the same as CAE (see Algorithm \ref{alg:pseudocodeCAE}).

For qualitative comparison between CAE and CAE$_{\text{FC}}$, we perform a clustering task by using a simple dataset in Fig. \ref{fig:comparison_data}. The dataset consists of 16,000 data points generated from a 2D Gaussian distribution ($\mu = (0, 0), \Sigma = [1, 0; 0, 1]$) and scaled to [0, 1]. Each data point is given to CAE and CAE$_{\text{FC}}$ only once. Figs. \ref{fig:comparison_cae} and \ref{fig:comparison_caefc} show the results of the clustering task by CAE and CAE$_{\text{FC}}$, respectively. Note that in Fig. \ref{fig:comparison_cae}, two isolated nodes are shown by different colors. Obviously, CAE generates a large number of nodes, while CAE$_{\text{FC}}$ generates a small number of nodes.

If the objective of the clustering task is efficient information extraction, it is required to approximate a given data set with fewer nodes. From this perspective, CAE$_{\text{FC}}$ is preferable to CAE. Note that the desired number of nodes depends on the purpose of the application and therefore CAE may be preferable in some other cases.

\subsection{CA+}
\label{sec:caPlus}

CA+ is a variant of CAE$_{\text{FC}}$, i.e., CAE$_{\text{FC}}$ without topology. Thus, the majority of the learning procedure is the same as CAE$_{\text{FC}}$. The differences between CA+ and CAE$_{\text{FC}}$ are summarized as follows:

\begin{itemize}
	\setlength{\leftskip}{-3mm}
	\vspace{1mm}
	\item For CA+, the learning processes related to the edge information are removed from Algorithm \ref{alg:pseudocodeCAE}, namely lines 25-26, line 28, and lines 31-38. As a result, the output of CA+ is only a node set $\mathcal{Y}$.

	\vspace{1mm}
	\item Since all the nodes of CA+ are isolated (i.e., no node has edges), a process for deleting isolated nodes (lines 39-40 in Algorithm \ref{alg:pseudocodeCAE}) is removed.
	
	\vspace{-1mm}
	\item In CAE$_{\text{FC}}$, the weight updating rule in (\ref{eq:updateNodeWeight2}) uses the edge information for defining the neighbor nodes $\mathcal{N}_{s_{1}}$ (see also lines 29-30 in Algorithm \ref{alg:pseudocodeCAE}). Since CA+ has no edges, CA+ updates only the 2nd winner node $\mathbf{y}_{s_{2}}$ as follows:
	\begin{equation}
		{\mathbf{y}}_{s_{2}} \gets {\mathbf{y}}_{s_{2}} + \frac{1}{100 M_{s_{2}}} \left({\mathbf{x}}-{\mathbf{y}}_{s_{2}}\right),
		\label{eq:updateNodeWeight2_ca_plus}
	\end{equation}
	where $M_{s_{2}}$ is the winning count of $\mathbf{y}_{s_{2}}$.

	In CA+, $\mathbf{y}_{s_{2}}$ does not need to be moved significantly because more nodes are generated than CAE$_{\text{FC}}$. Therefore, the coefficient in (\ref{eq:updateNodeWeight2_ca_plus}) is set to $1/100$ in CA+ whereas it is $1/10$ in CAE$_{\text{FC}}$.
\end{itemize}

Similar to the qualitative comparison between CAE and CAE$_{\text{FC}}$, CA+ and CAE$_{\text{FC}}$ are also compared by the same clustering task with the dataset in Fig. \ref{fig:comparison_data}. Each data point in Fig. \ref{fig:comparison_data} is given to CA+ and CAE$_{\text{FC}}$ only once. Figs. \ref{fig:comparison_caefc} and \ref{fig:comparison_ca_plus} show the results of the clustering task by CAE$_{\text{FC}}$ and CA+, respectively. CAE$_{\text{FC}}$ does not generate any nodes on the outer edges of the Gaussian distribution, while CA+ generates some nodes to cover the entire distribution. As a result, more nodes are generated by CA+ (75 nodes) than CAE$_{\text{FC}}$ (75 nodes) for the task in Fig. \ref{fig:comparison_data}. This property of CA+ is preferable for a client in federated clustering because the generated nodes in each client are utilized as training data for the server.

\begin{figure*}[htbp]
	\centering
	\includegraphics[width=6.0in]{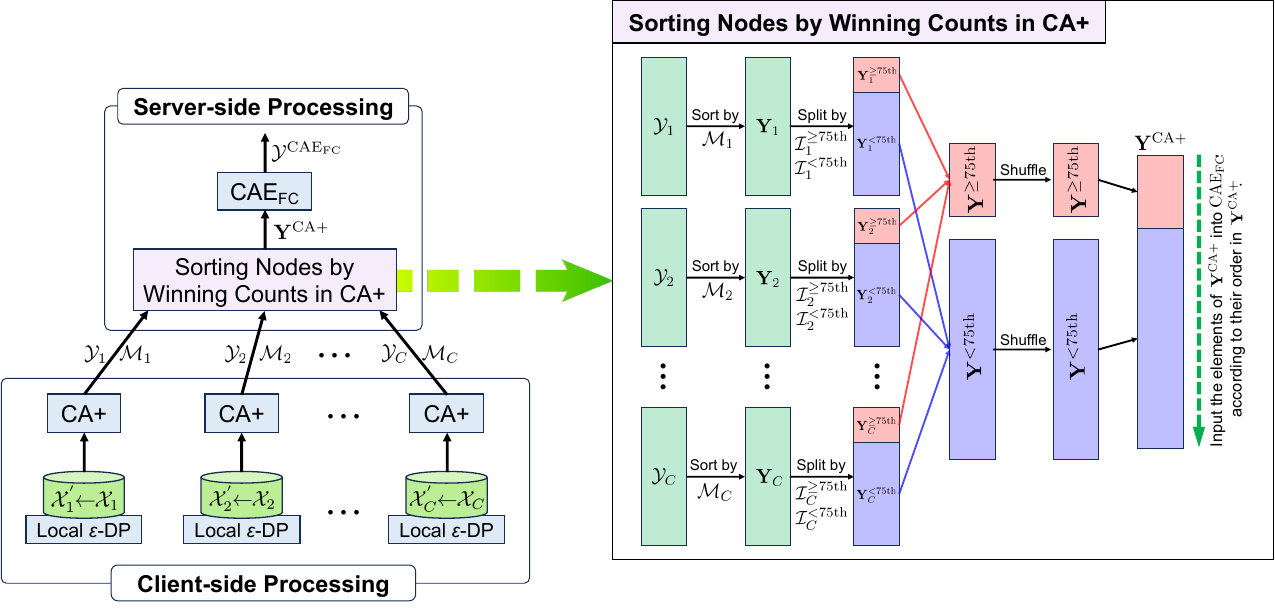}
	\caption{Overview of FCAC.}
	\vspace{-2mm}
	\label{fig:fcac_overview}
\end{figure*}

\subsection{FCAC}
\label{sec:fcac}

\subsubsection{Overview}
\label{sec:overviewFCAC}
Fig. \ref{fig:fcac_overview} shows the outline of FCAC. Similar to $k$-FED \cite{dennis21} and MUFC \cite{pan23}, FCAC is a one-shot federated clustering algorithm that does not require multiple communication rounds between the server and clients. In addition, since FCAC uses ART-based clustering algorithms, FCAC can perform continual learning while avoiding catastrophic forgetting. Furthermore, FCAC does not require any iterative learning process for convergence unlike $k$-FED and MUFC with $k$-means as a base clusterer.

As a unique learning process of FCAC, the training data of the server (i.e., generated nodes in each client) are re-ordered according to the importance of each node (i.e., a winning count $\mathcal{M}$), and then nodes with higher importance are fed to CAE$_{\text{FC}}$ first one by one in the order of importance from the most important node (see Sorting Nodes by Winning Counts in CA+ in Fig. \ref{fig:fcac_overview}). Since the similarity threshold $V_{\text{threshold}}$ in (\ref{eq:pairwiseCIM}), which is an important parameter for achieving high clustering performance, is calculated based on a certain number of initial training data points, this process can be expected to improve the stability of network creation for CAE$_{\text{FC}}$ in the early stage of learning.

\subsubsection{Learning Procedure}
\label{sec:learningFCAC}
Algorithm \ref{alg:pseudocodeFCAC} summarizes the learning procedure of FCAC. In the client-side processing, local $\epsilon$-differential privacy is applied to each data point, and then CA+ is performed in parallel for obtaining $\mathcal{Y}_{c}$ and $\mathcal{M}_{c}$ ($c = 1,2,\ldots,C$).

In the server-side processing, as a first step, the nodes from each client are re-ordered based on a winning count of each node. Algorithm \ref{alg:sortnodes} summarizes the sorting mechanism of Fig. \ref{fig:fcac_overview} in detail. First, the nodes in a client $\mathcal{Y}_{c}$ split into two ordered groups based on the winning counts $\mathcal{M}_{c}$, namely $\mathbf{Y}^{\geq75\text{th}}_{c}$ and $\mathbf{Y}^{<75\text{th}}_{c}$ (lines 2-6). Here, $\mathbf{Y}^{\geq75\text{th}}_{c}$ consists of the nodes above the 75th percentile of elements in $\mathcal{M}_{c}$, i.e., a group of nodes with high winning counts $\mathcal{M}_{c}$, while $\mathbf{Y}^{<75\text{th}}_{c}$ consists of the nodes below the 75th percentile of elements in $\mathcal{M}_{c}$, i.e., a group of nodes with low winning counts $\mathcal{M}_{c}$. Then, the order of the nodes is randomly shuffled in each group (lines 7-8). Last, the shuffled node groups $\mathbf{Y}^{\geq75\text{th}}_{c}$ and $\mathbf{Y}^{<75\text{th}}_{c}$ are marged into $\mathbf{Y}^{\text{CA+}}$ while preserving their ordering (line 9).

\begin{algorithm}[htbp]
	\DontPrintSemicolon
	\KwIn{\\
		a privacy budget $\epsilon$ \\
		the number of clients $C$ \\
		a set of data points for each client $ \mathcal{X}_{c} $ $(c = 1,2,\ldots,C)$
	}
	\KwOut{\\
		a set of nodes $\mathcal{Y}^{\text{CAE}_{\text{FC}}}$\\
	}
	\tcc{Client-side Processing}
	\SetKwBlock{DoParallel}{for $c = \{1,\ldots,C\}$ do in parallel}{end}
	\DoParallel{	
		Generate a set of privacy-preserving data points $\mathcal{X}_{c}^{'}$ by local $\epsilon$-differentical privacy. \tcp*{(\ref{eq:noisedData}),(\ref{eq:ICDF}),(\ref{eq:sensitivityDP})}
		$\mathcal{Y}_{c} \leftarrow \text{Perform CA+ by using } \mathcal{X}_{c}^{'}$ \\
	}
	\tcc{Server-side Processing}
	$\mathbf{Y}^{\text{CA+}} \leftarrow$ Sorting nodes by winning counts in CA+. \tcp*{Algorithm \ref{alg:sortnodes}}
	$\mathcal{Y}^{\text{CAE}_{\text{FC}}} \leftarrow \text{Perform CAE$_{\text{FC}}$ by using } \mathbf{Y}^{\text{CA+}}$ \\
	
	\caption{Learning procedure of FCAC}
	\label{alg:pseudocodeFCAC}
\end{algorithm}

After preparing sorted nodes $\mathbf{Y}^{\text{CA+}}$, the elements of $\mathbf{Y}^{\text{CA+}}$ is fed to CAE$_{\text{FC}}$ according to their order in $\mathbf{Y}^{\text{CA+}}$ (line 7 in Algorithm \ref{alg:pseudocodeFCAC}).

\section{Simulation Experiments}
\label{sec:experiment}
In this section, first, the effect of local $\epsilon$-differential privacy is qualitatively analyzed. Next, the continual learning ability of FCAC is demonstrated by using a synthetic dataset. Then, the clustering performance of FCAC is quantitatively evaluated by using real-world datasets and compared with state-of-the-art federated clustering algorithms. Last, we analyze the computational complexity of FCAC.

Note that all experiments are carried out on Matlab 2023b and Python 3.10 with the Apple M1 Ultra processor and 128GB RAM.

\subsection{Effect of Local $\epsilon$-Differential Privacy}
\label{sec:effectLDP}
Apart from the abilities of FCAC, an intuitive understanding of the effect of local $\epsilon$-differential privacy on a dataset is necessary to discuss the clustering performance of FCAC. In general, the value of $\epsilon$ controls the degree of data privacy protection, i.e., $\epsilon = 0$ means perfect privacy, while $\epsilon = \infty$ means no privacy guarantee. However, the relationship between the value of $\epsilon$ and the degree of data privacy protection is difficult to define quantitatively because it depends on the data sensitivity, the purpose of data analysis, and applications \cite{dwork14}. In this paper, therefore, we do not focus on the theoretical analysis of privacy protection.

\begin{figure}[htbp]
	\vspace{-5mm}
	\centering
	\subfloat[Dataset]{
		\includegraphics[width=1.38in]{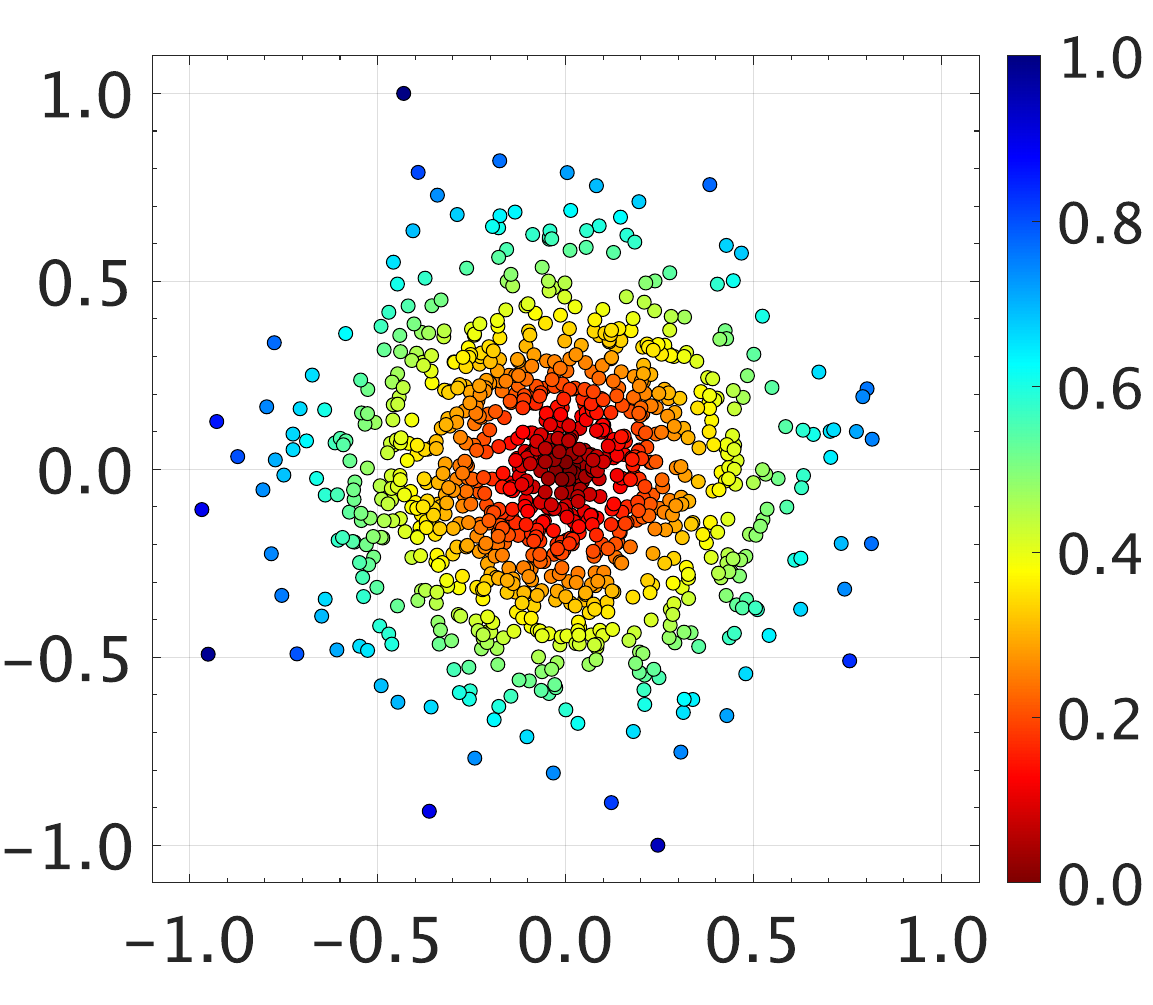}
		\label{fig:epsilon_original}
	}\hfil
	\subfloat[$\epsilon = 10$, $D_{\text{WS}} = 0.0793$]{
		\includegraphics[width=1.38in]{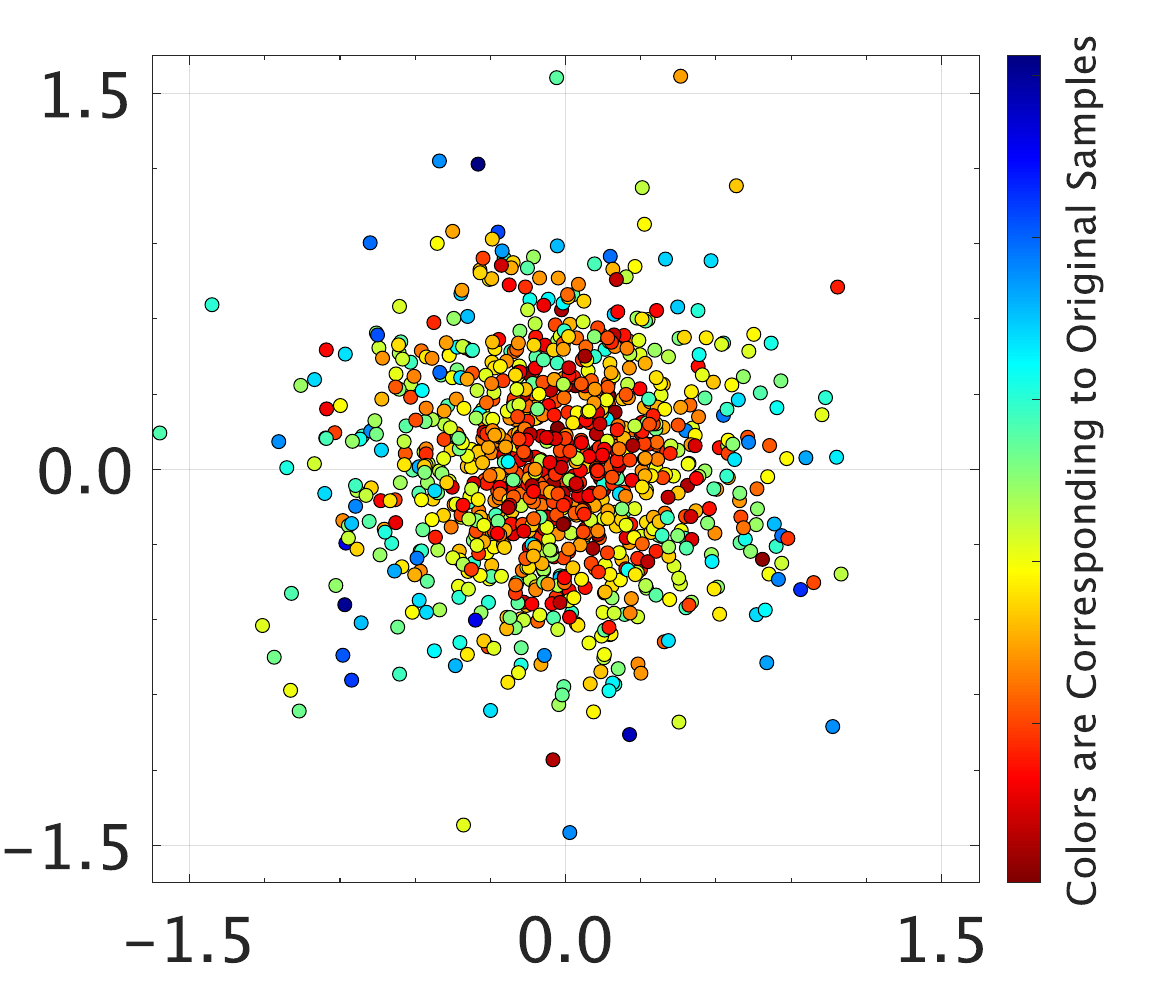}
		\label{fig:epsilon_10_noised}
	}\hfil \\ \vspace{-2mm}
	\subfloat[$\epsilon = 15$, $D_{\text{WS}} = 0.0416$]{
		\includegraphics[width=1.38in]{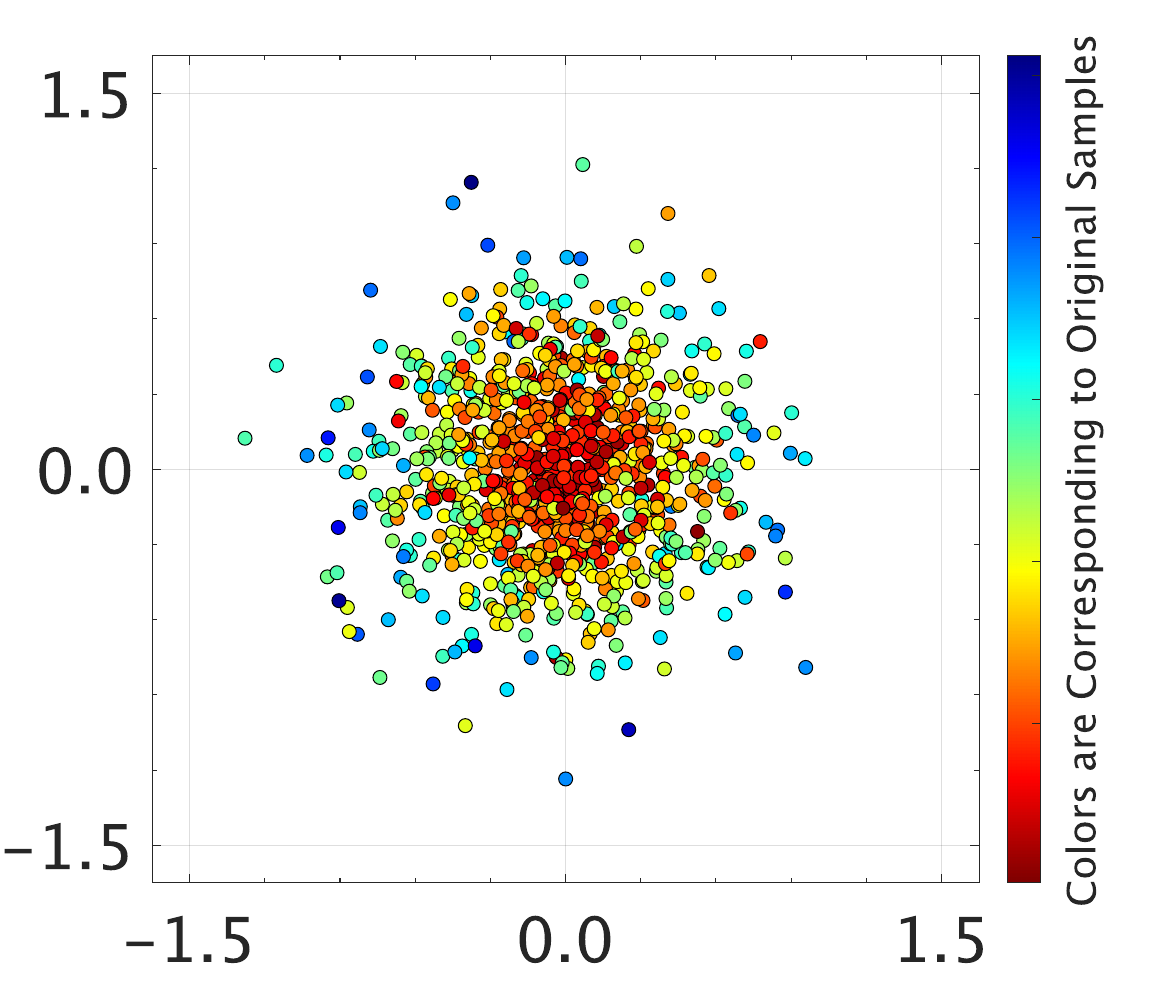}
		\label{fig:epsilon_15_noised}
	}\hfil
	\subfloat[$\epsilon = 25$, $D_{\text{WS}} = 0.0177$]{
		\includegraphics[width=1.38in]{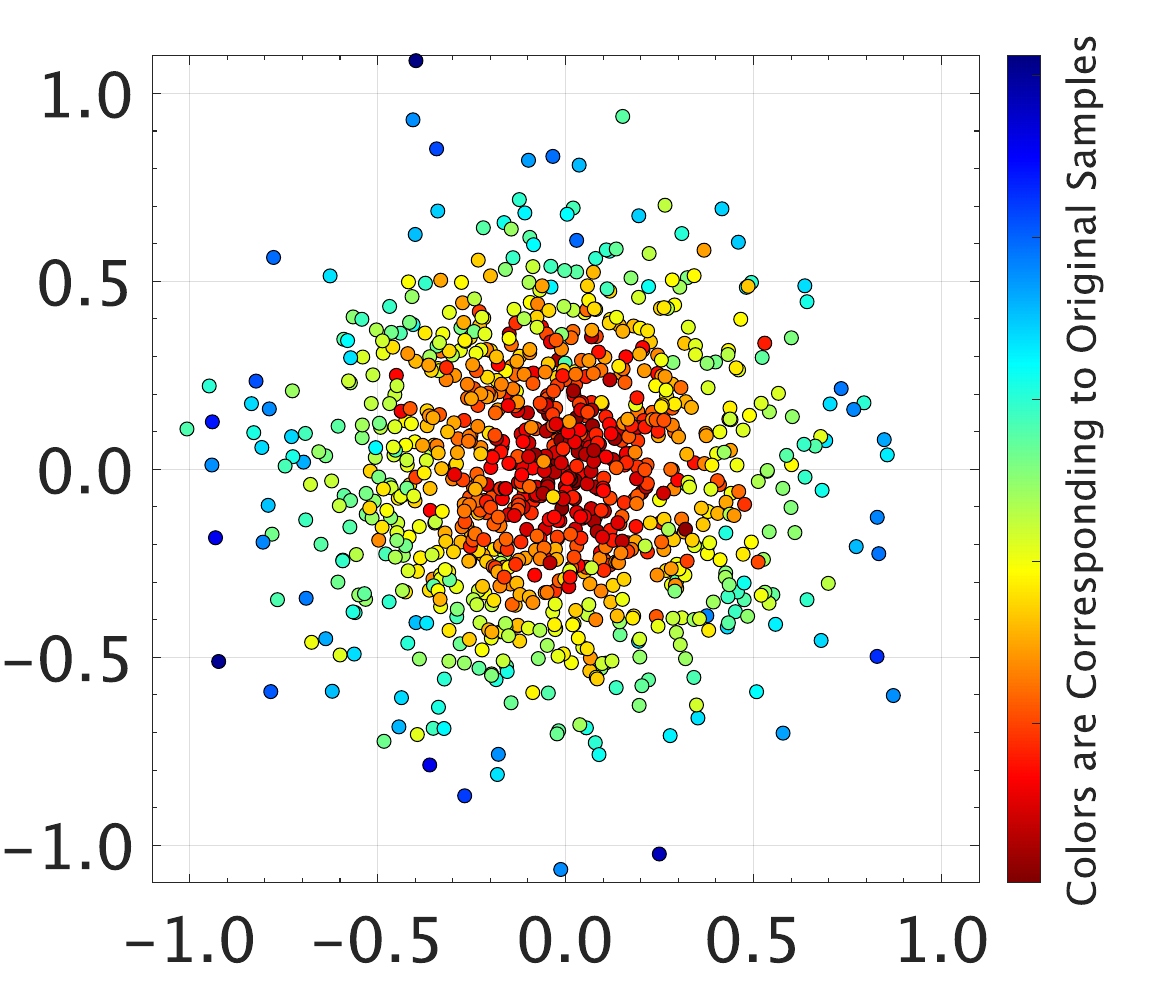}
		\label{fig:epsilon_25_noised}
	}\hfil \\ \vspace{-2mm}
	\subfloat[$\epsilon = 50$, $D_{\text{WS}} = 0.0060$]{
		\includegraphics[width=1.38in]{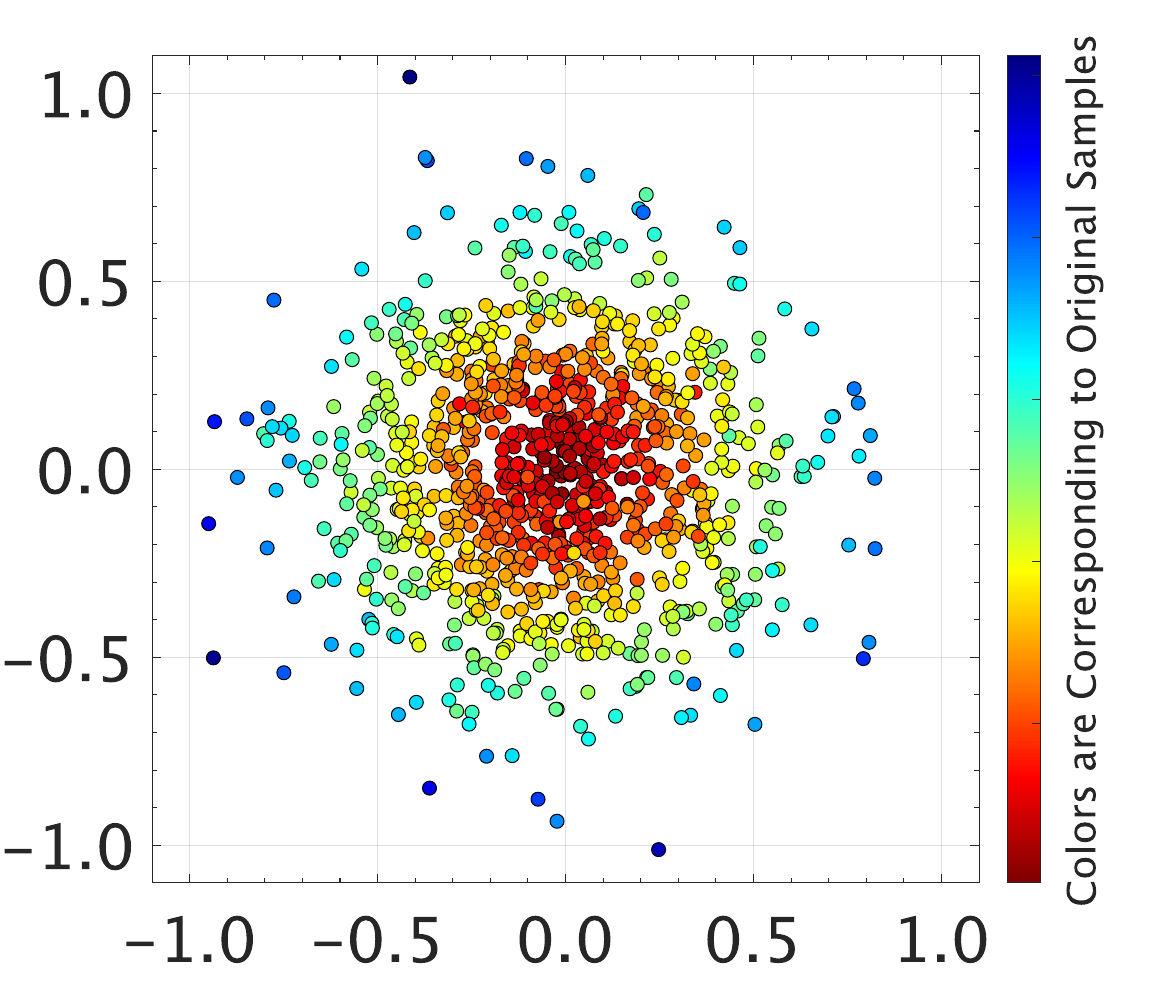}
		\label{fig:epsilon_50_noised}
	}\hfil
	\subfloat[$\epsilon = 75$, $D_{\text{WS}} = 0.0040$]{
		\includegraphics[width=1.38in]{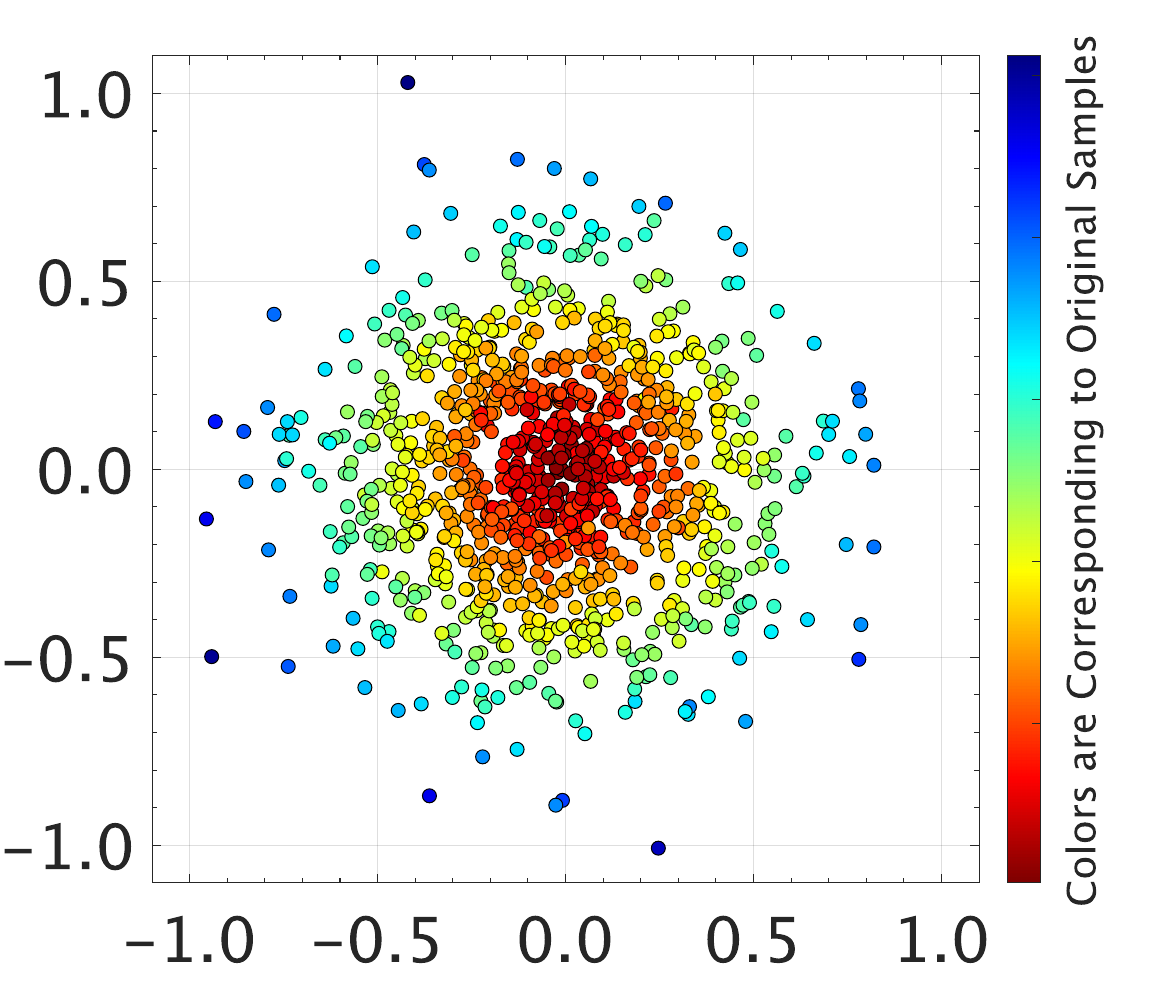}
		\label{fig:epsilon_75_noised}
	}
	\vspace{-1mm}
	\caption{Effects of local $\epsilon$-differential privacy.}
	\vspace{-4mm}
	\label{fig:epsilon_gaussian}
\end{figure}

Here, we qualitatively evaluate the effect of local $\epsilon$-differential privacy on a synthetic dataset which is shown in Fig. \ref{fig:epsilon_original}. The dataset consists of 1,000 data points generated from the 2D Gaussian distribution and it is scaled to [-1, 1]. In Fig. \ref{fig:epsilon_original}, the color of a node is red if the node position is close to (0, 0), while the color of a node is blue if the node position is far from (0, 0). The noise as local $\epsilon$-differential privacy for each data point is generated by using (\ref{eq:ICDF}) with $\mu = 0$. The value of $\epsilon$ is set as $\{ 10, 15, 25, 50, 75 \}$. To quantitatively measure the changes in the distribution from the original dataset to a noised one, we apply the 1-Wasserstein Distance ($D_{\text{WS}}$) which arises from the idea of optimal transport \cite{peyre19}. Note that $D_{\text{WS}}$ represents the amount of change in the data distribution, not represents the degree of data privacy protection.

\begin{algorithm}[htbp]
	\DontPrintSemicolon
	\KwIn{\\
		node sets $\{\mathcal{Y}_{1},\mathcal{Y}_{2},\ldots,\mathcal{Y}_{C}\}$ \\
		winning count sets $\{\mathcal{M}_{1},\mathcal{M}_{2},\ldots,\mathcal{M}_{C}\}$
	}
	\KwOut{\\
		sorted nodes $\mathbf{Y}^{\text{CA+}}$\\
	}
	Initialize $\mathbf{Y}^{\geq75\text{th}} \leftarrow \text{null} $, and $\mathbf{Y}^{<75\text{th}} \leftarrow \text{null}$. \\
	\For{$c = 1,\ldots,C$}
	{
		Extract indices $\mathcal{I}^{\geq75\text{th}}_{c}$ and $\mathcal{I}^{<75\text{th}}_{c}$ which are above and below the 75th percentile of elements in $\mathcal{M}_{c}$, respectively. \\
		Extract $\mathcal{Y}^{\geq75\text{th}}_{c}$ and $\mathcal{Y}^{<75\text{th}}_{c}$ from $\mathcal{Y}_{c}$ corresponding to $\mathcal{I}^{\geq75\text{th}}_{c}$ and $\mathcal{I}^{<75\text{th}}_{c}$, respectively. \\
		
		$\mathbf{Y}^{\geq75\text{th}} \leftarrow \mathbf{Y}^{\geq75\text{th}}.\text{append}(\mathcal{Y}^{\geq75\text{th}}_{c})$ \\
		$\mathbf{Y}^{<75\text{th}} \leftarrow \mathbf{Y}^{<75\text{th}}.\text{append}(\mathcal{Y}^{<75\text{th}}_{c})$ \\
	}
	$\mathbf{Y}^{\geq75\text{th}} \leftarrow \mathrm{Shuffle}(\mathbf{Y}^{\geq75\text{th}})$ \\
	$\mathbf{Y}^{<75\text{th}} \leftarrow \mathrm{Shuffle}(\mathbf{Y}^{<75\text{th}})$ \\
	$\mathbf{Y}^{\text{CA+}} \leftarrow [\mathbf{Y}^{\geq75\text{th}}; \mathbf{Y}^{<75\text{th}}]$ \\
	
	\caption{Sorting nodes by winning counts in CA+}
	\label{alg:sortnodes}
\end{algorithm}

Figs. \ref{fig:epsilon_10_noised}-\ref{fig:epsilon_75_noised} show the effect of local $\epsilon$-differential privacy on a 2D Gaussian distribution corresponding to each $\epsilon$ value. The corresponding $D_{\text{WS}}$ value for each distribution is shown next to the $\epsilon$ value. In Figs. \ref{fig:epsilon_10_noised}-\ref{fig:epsilon_75_noised}, the color of each node corresponds to the color of the node before adding noise (i.e., the node in Fig. \ref{fig:epsilon_original}). The effect of local $\epsilon$-difference privacy is quite limited in Figs. \ref{fig:epsilon_50_noised} and \ref{fig:epsilon_75_noised} ($\epsilon = 50, 75$). The disturbance of the distribution is observed in Figs. \ref{fig:epsilon_15_noised} and \ref{fig:epsilon_25_noised} ($\epsilon = 15, 25$) but each figure also shows clear similarity to the distribution of the original dataset (i.e., red color nodes are placed in the central region of the distribution, while the green and blue color nodes are placed in the edge region of the distribution). In contrast, the distribution in Fig. \ref{fig:epsilon_10_noised} ($\epsilon = 10$) differs significantly from the one in Fig. \ref{fig:epsilon_original}. This indicates that the data utility is too low although the data privacy protection is high.

From the above-mentioned observations, we only consider $\epsilon = 15, 25, 50, 75$ for local $\epsilon$-differential privacy in the subsequent sections.

\subsection{Continual Learning Ability}
\label{sec:continual}
\begin{figure}[htbp]
	\centering
	\includegraphics[width=2.45in]{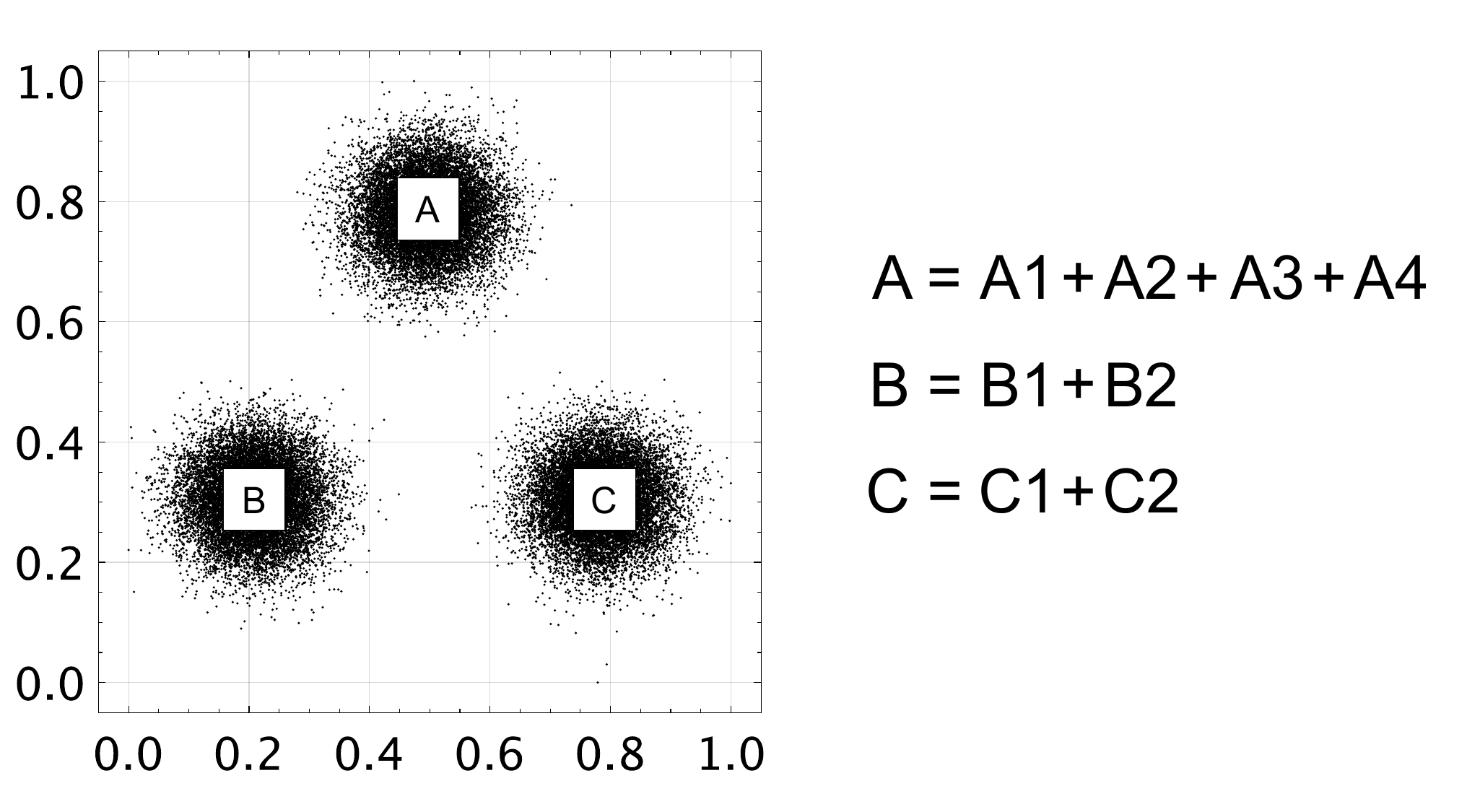}
	\vspace{-1mm}
	\caption{Entire dataset for the continual learning.}
	\label{fig:synthetic_gaussian_all}
\end{figure}

\begin{figure}[htbp]
	\vspace{-4mm}
	\centering
	\subfloat[Round \#1]{
		\includegraphics[width=1.05in]{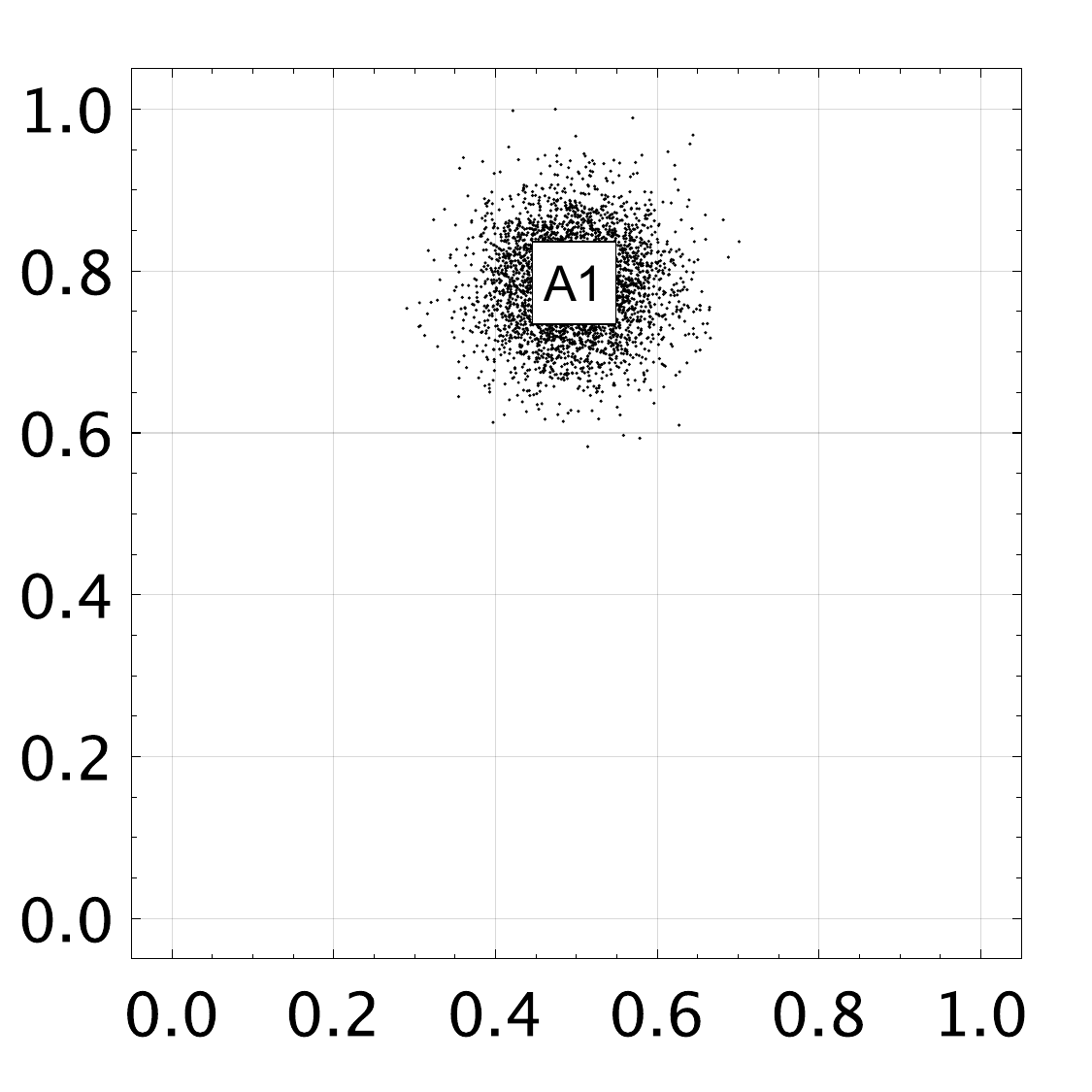}
		\label{fig:synthetic_gaussian_1_round_1}
	}\hfil
	\subfloat[Round \#2]{
		\includegraphics[width=1.05in]{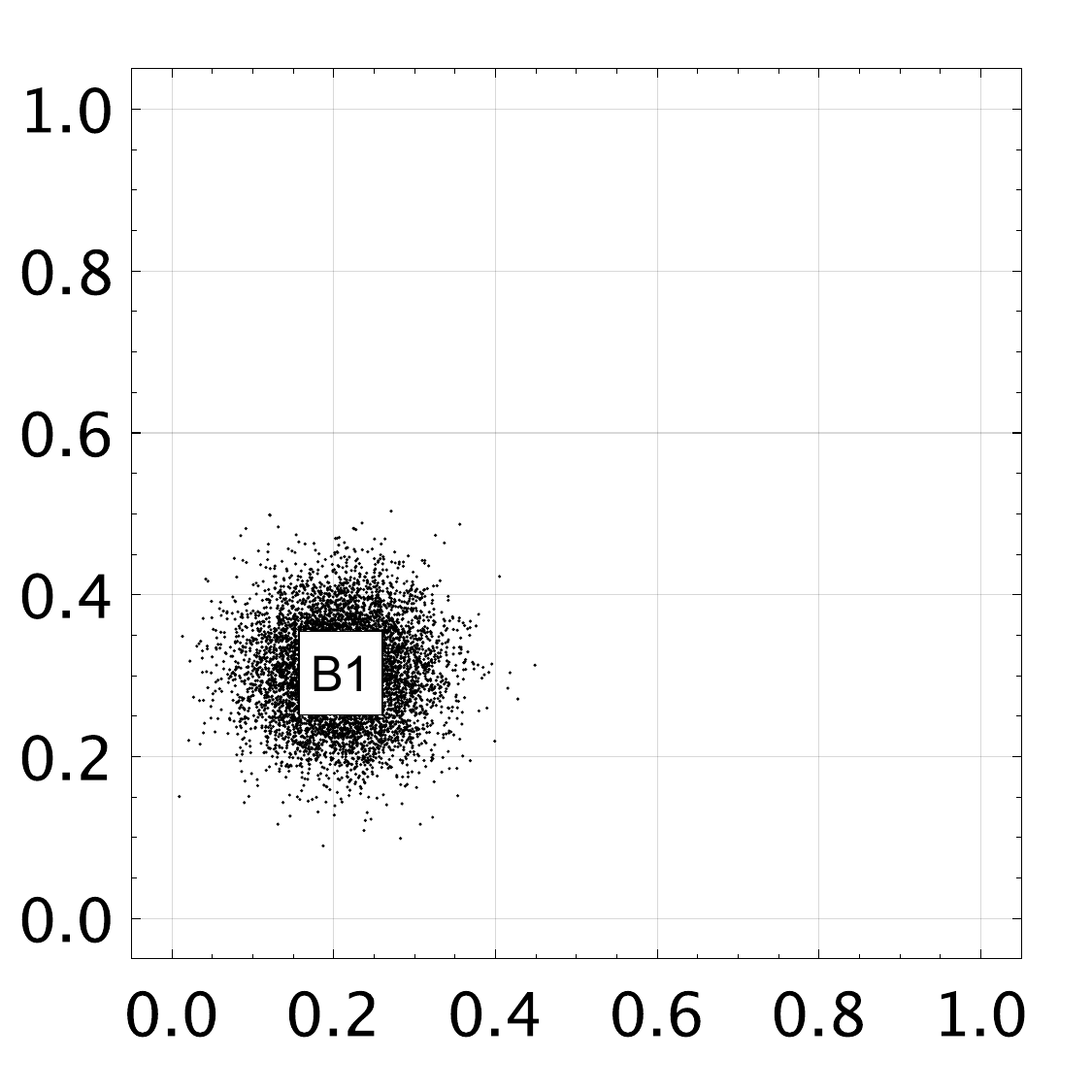}
		\label{fig:synthetic_gaussian_1_round_2}
	}\hfil
	\subfloat[Round \#3]{
		\includegraphics[width=1.05in]{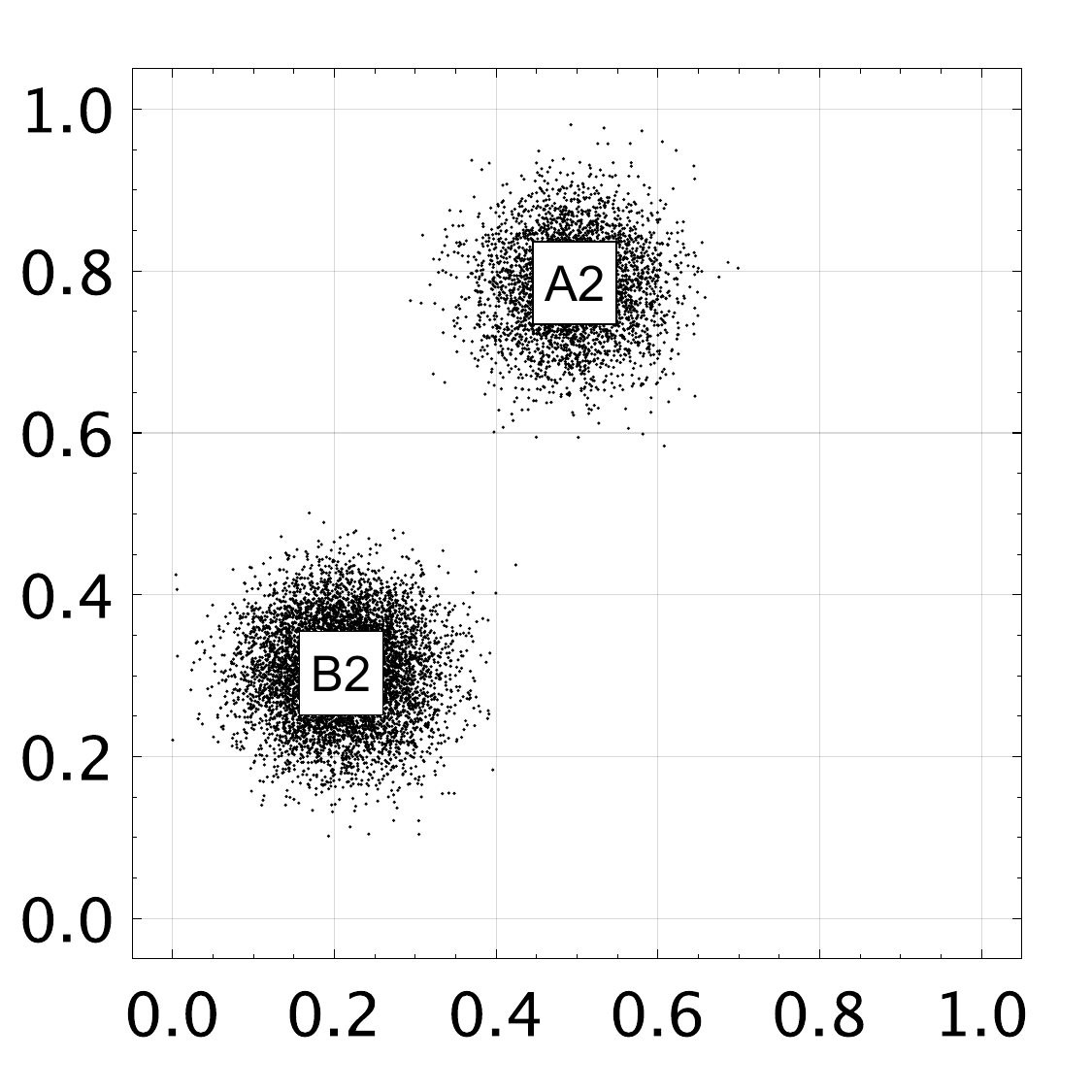}
		\label{fig:synthetic_gaussian_1_round_3}
	}
	\vspace{-1mm}
	\caption{Visualization of the synthetic dataset for the client \#1.}
	\label{fig:synthetic_gaussian_1}
\end{figure}

\begin{figure}[htbp]
	\vspace{-4mm}
	\centering
	\subfloat[Round \#1]{
		\includegraphics[width=1.05in]{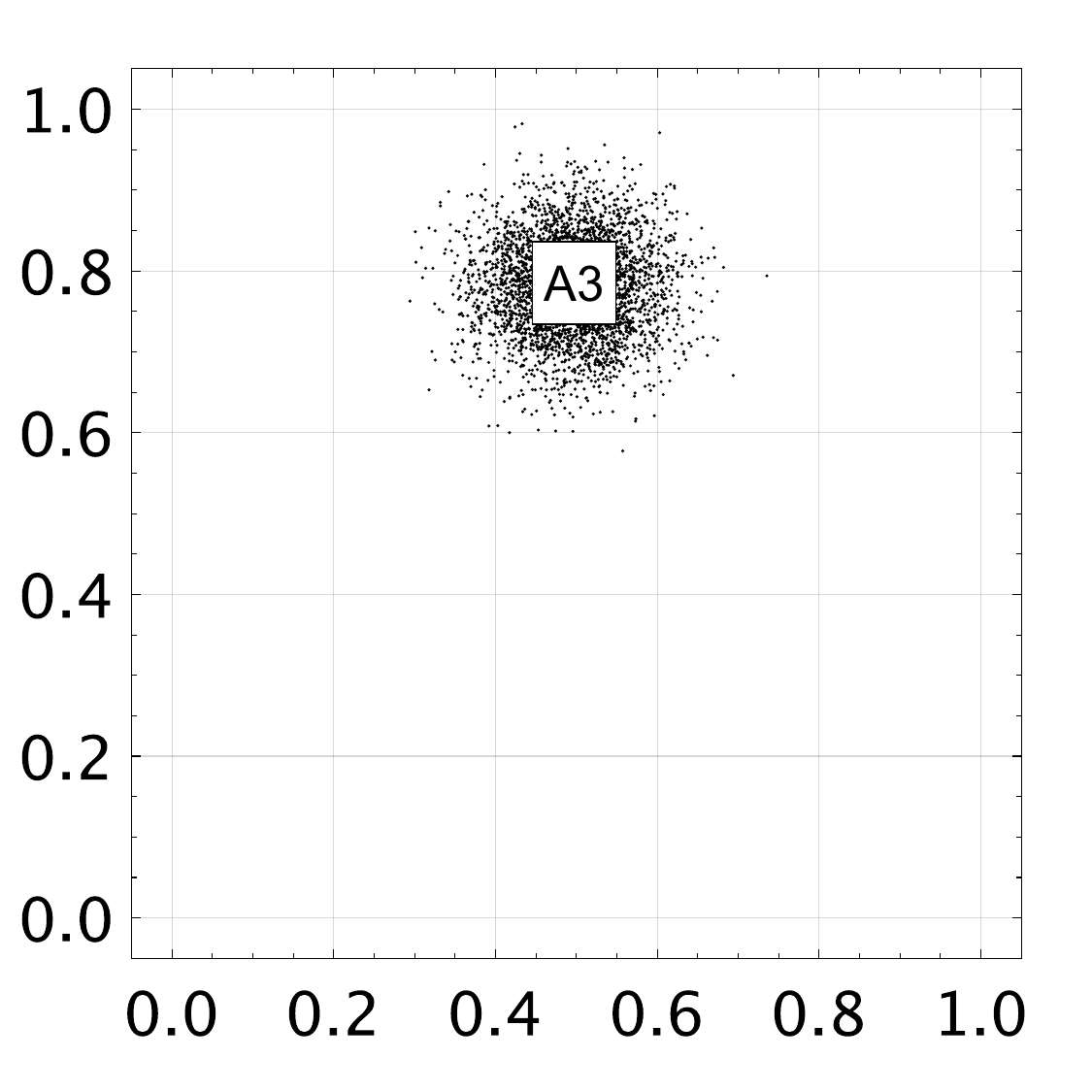}
		\label{fig:synthetic_gaussian_2_round_1}
	}\hfil
	\subfloat[Round \#2]{
		\includegraphics[width=1.05in]{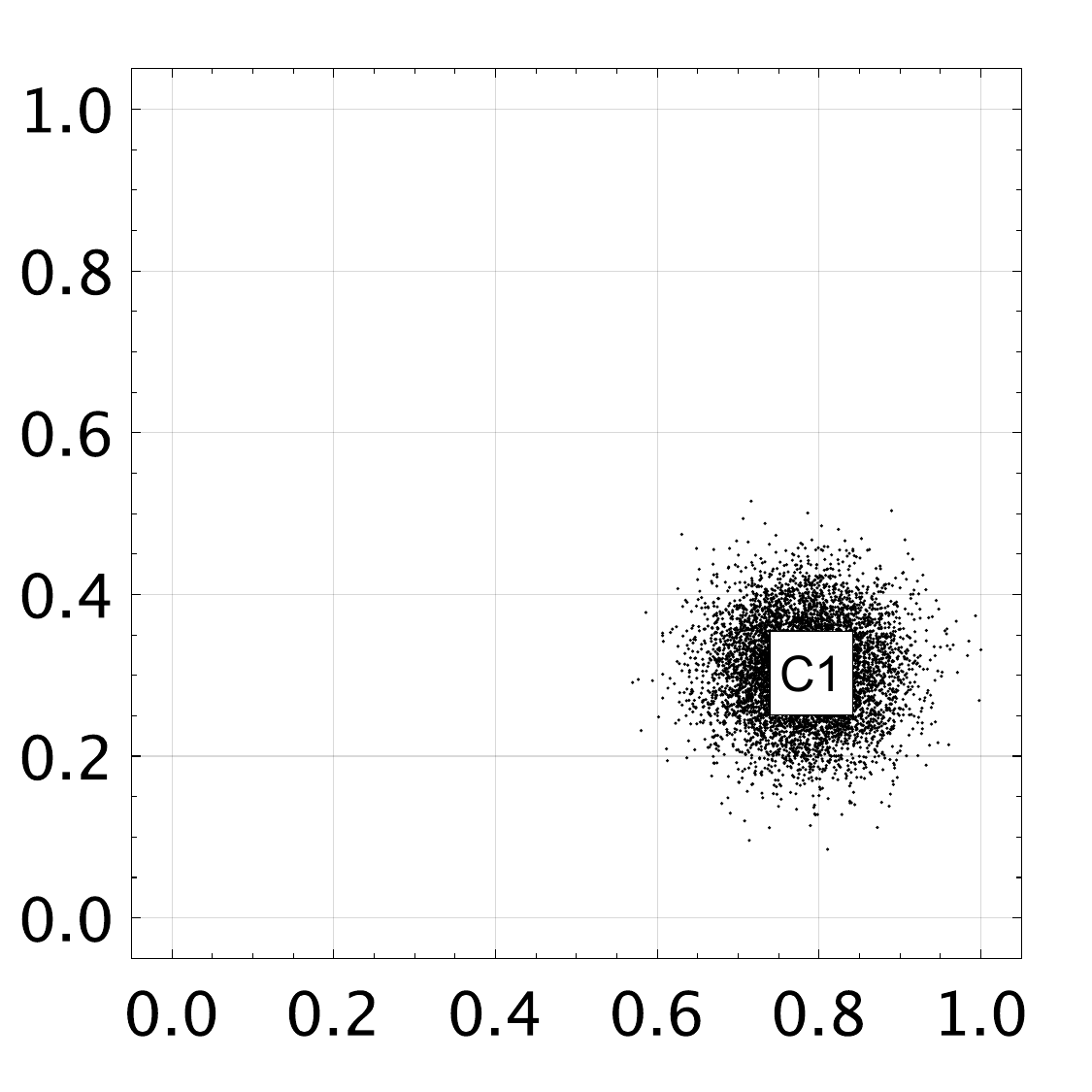}
		\label{fig:synthetic_gaussian_2_round_2}
	}\hfil
	\subfloat[Round \#3]{
		\includegraphics[width=1.05in]{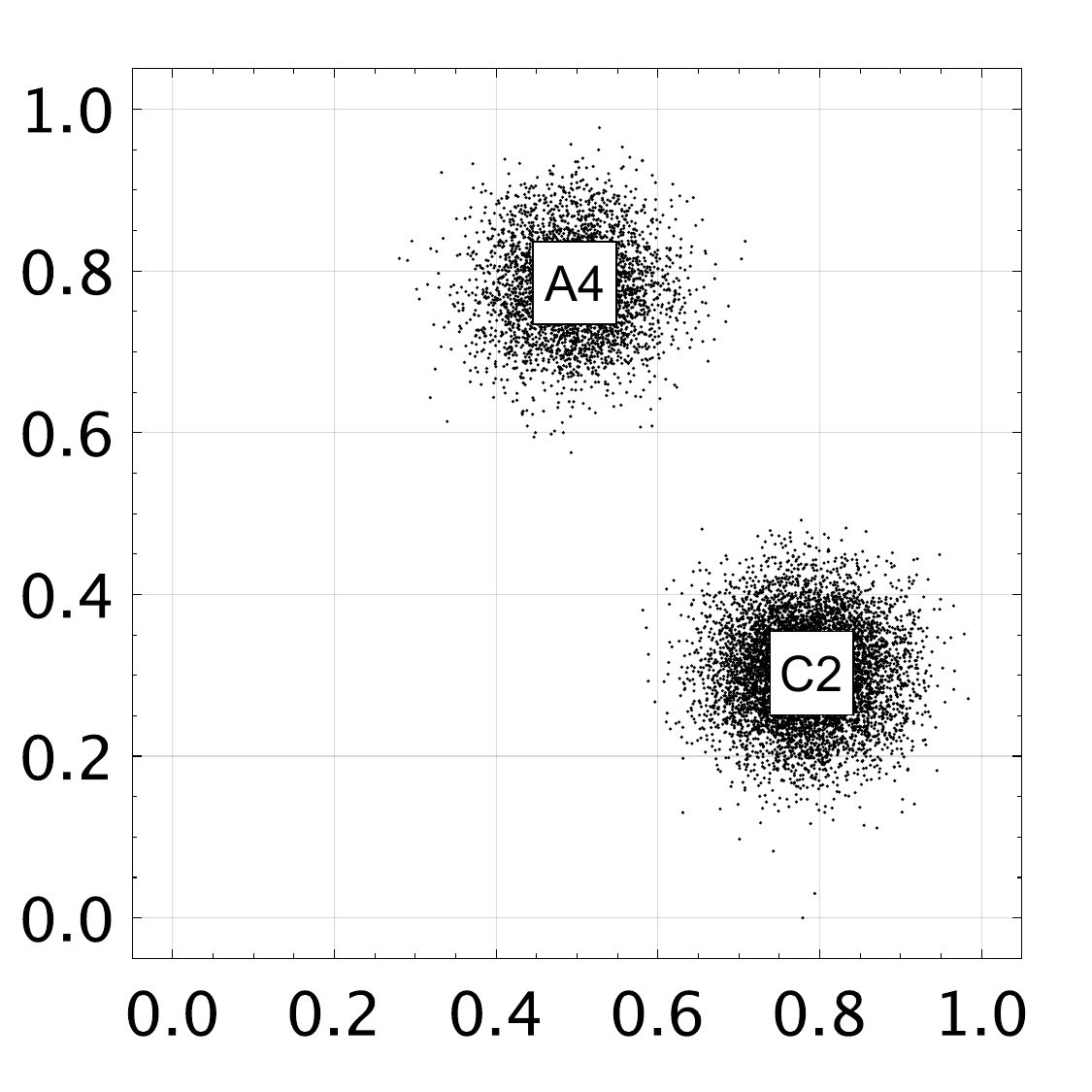}
		\label{fig:synthetic_gaussian_2_round_3}
	}
	\vspace{-1mm}
	\caption{Visualization of the synthetic dataset for the client \#2.}
	\label{fig:synthetic_gaussian_2}
\end{figure}

This section demonstrates the continual learning ability of FCAC. Here, we consider that FCAC has two clients (i.e., client \#1, client \#2) and a server. The entire dataset used in this experiment is shown in Fig. \ref{fig:synthetic_gaussian_all}. Each distribution in the entire dataset consists of 15,000 data points generated from the 2D Gaussian distribution, and it is scaled to [0, 1]. In order to perform continual learning in the practical environment, the entire dataset is divided into eight subsets without duplication as shown in Figs. \ref{fig:synthetic_gaussian_1} and \ref{fig:synthetic_gaussian_2} (i.e., A1, A2, A3, A4, B1, B2, C1, C2). The subsets A1, A2, A3, and A4 consist of 3,750 data points each, while the subsets B1, B2, C1, and C2 consist of 7,500 data points each. Each data point in each subset is given to FCAC only once. Throughout the round \#1 to \#3, CA+ in each client and CAE$_{\text{FC}}$ in the server are continually updated without being initialized.

\begin{figure}[htbp]
	\vspace{-4mm}
	\centering
	\subfloat[Client \#1 (CA+)]{
		\includegraphics[width=1.05in]{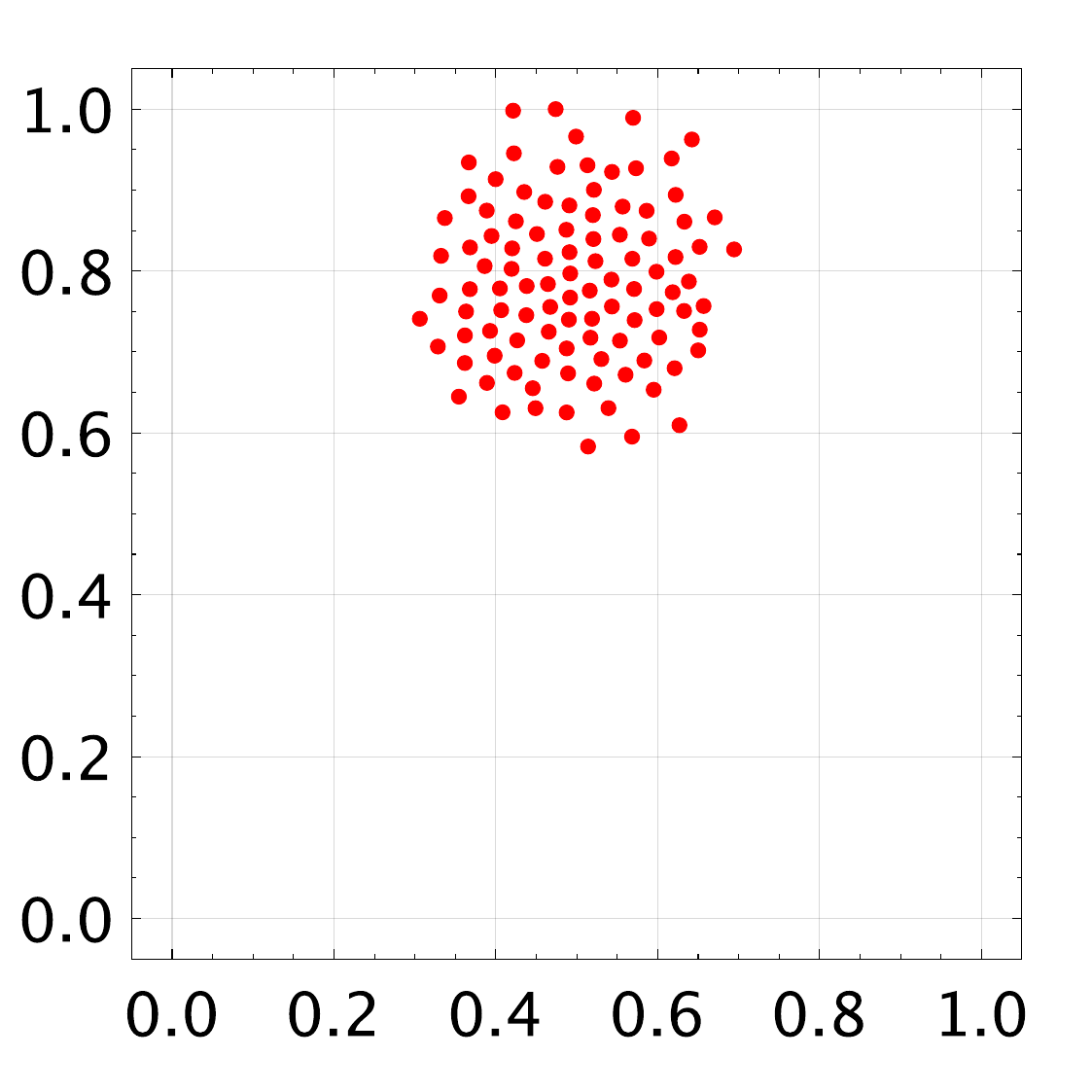}
		\label{fig:client_1_round_1}
	}\hfil
	\subfloat[Client \#2 (CA+)]{
		\includegraphics[width=1.05in]{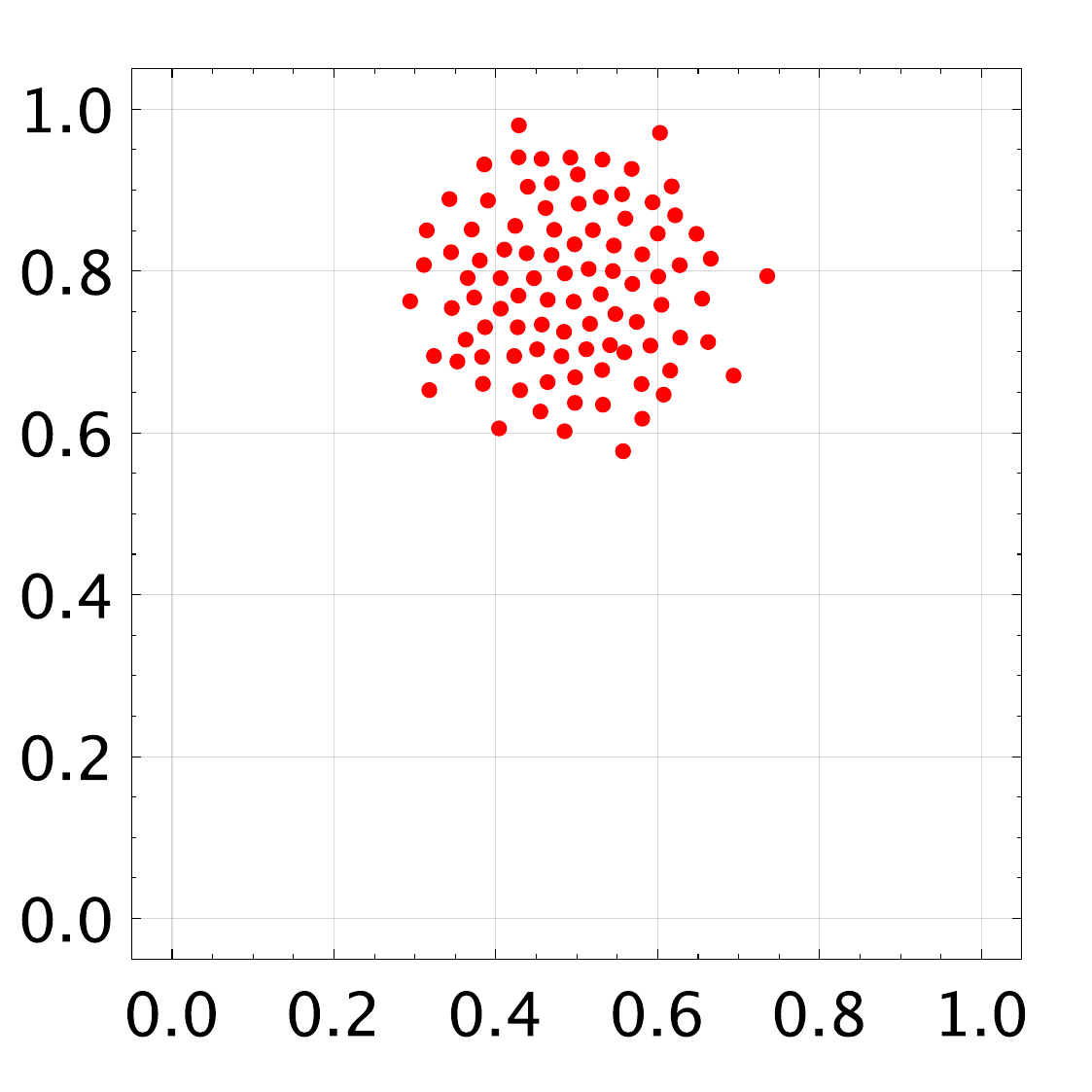}
		\label{fig:client_2_round_1}
	}\hfil
	\subfloat[Server (CAE$_{\text{FC}}$)]{
		\includegraphics[width=1.05in]{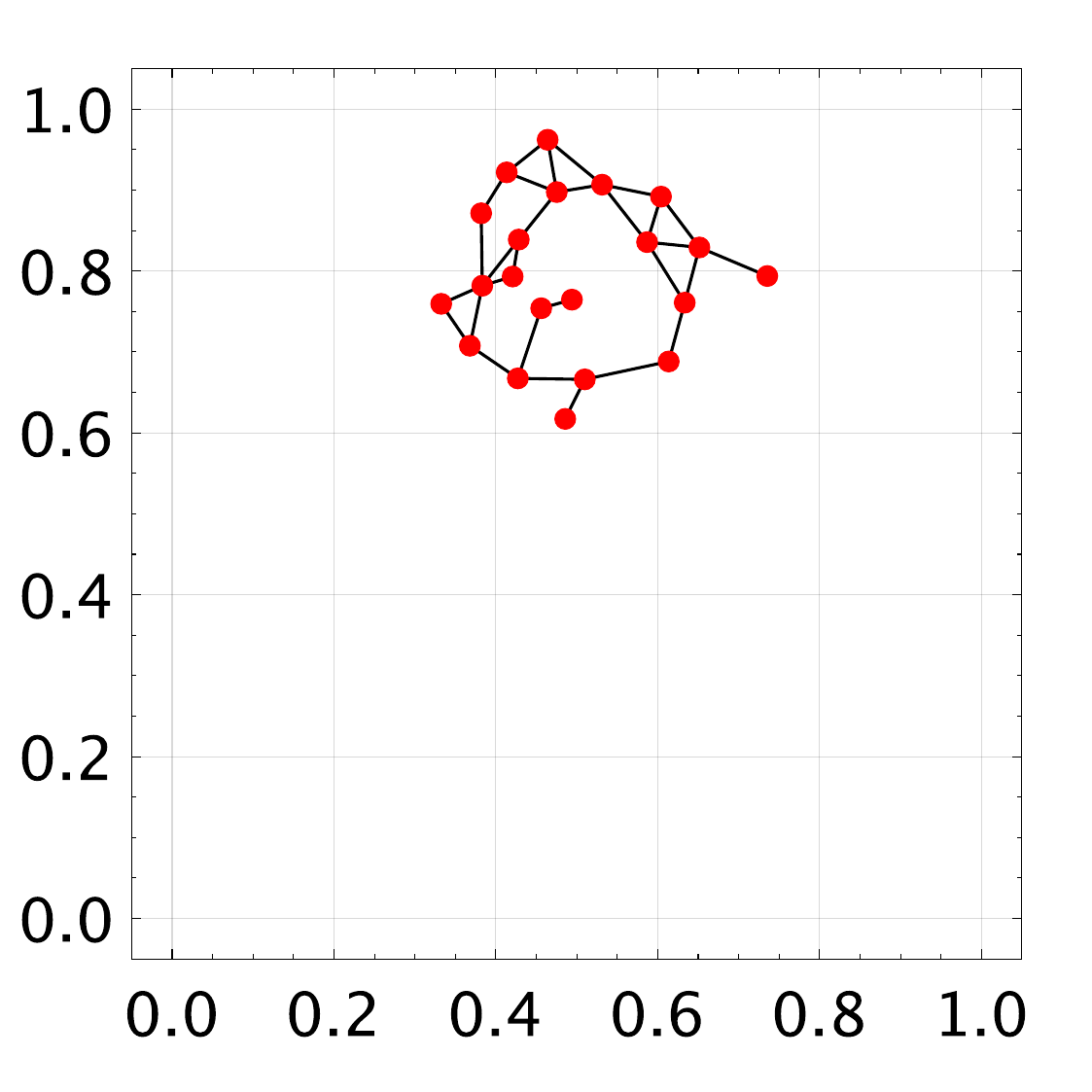}
		\label{fig:server_net_1}
	}
	\vspace{-1mm}
	\caption{Clustering results in the round \#1.}
	\label{fig:round_1}
\end{figure}

Fig. \ref{fig:round_1} shows the clustering results of the client \#1, the client \#2, and the server in the round \#1. It can be seen that the server generates networks by using the generated nodes in clients \#1 and \#2.

\begin{figure}[htbp]
	\vspace{-4mm}
	\centering
	\subfloat[Client \#1 (CA+)]{
		\includegraphics[width=1.05in]{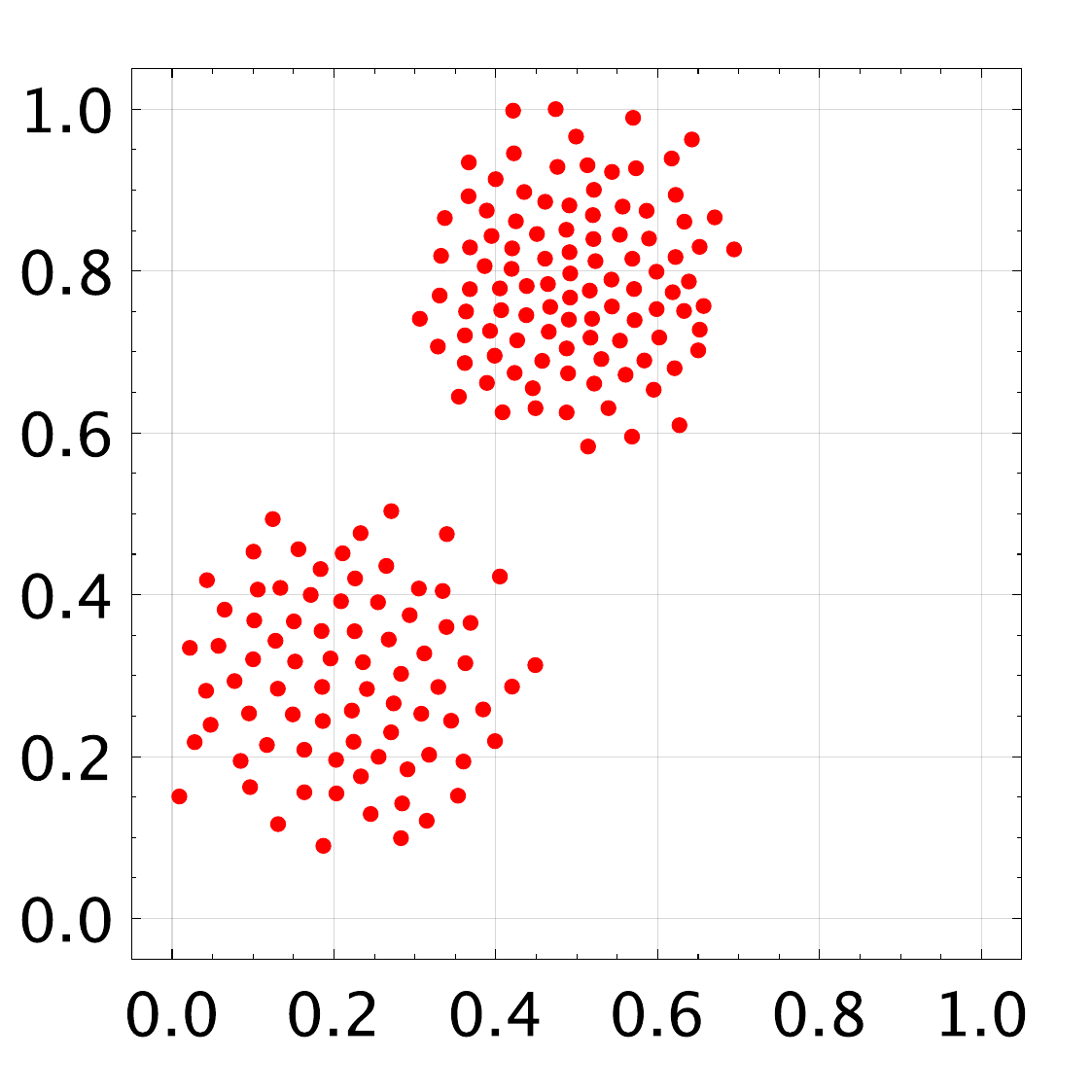}
		\label{fig:client_1_round_2}
	}\hfil
	\subfloat[Client \#2 (CA+)]{
		\includegraphics[width=1.05in]{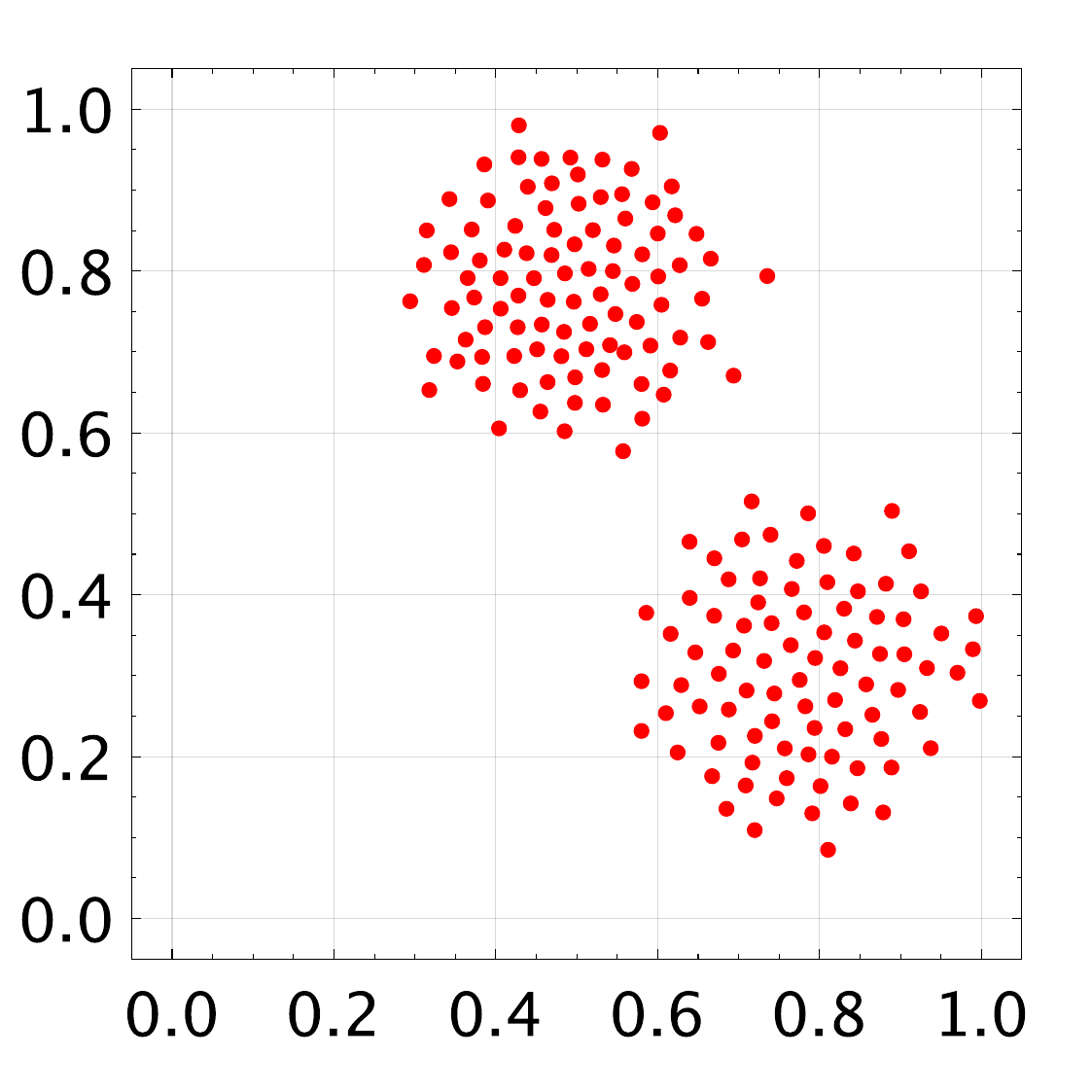}
		\label{fig:client_2_round_2}
	}\hfil
	\subfloat[Server (CAE$_{\text{FC}}$)]{
		\includegraphics[width=1.05in]{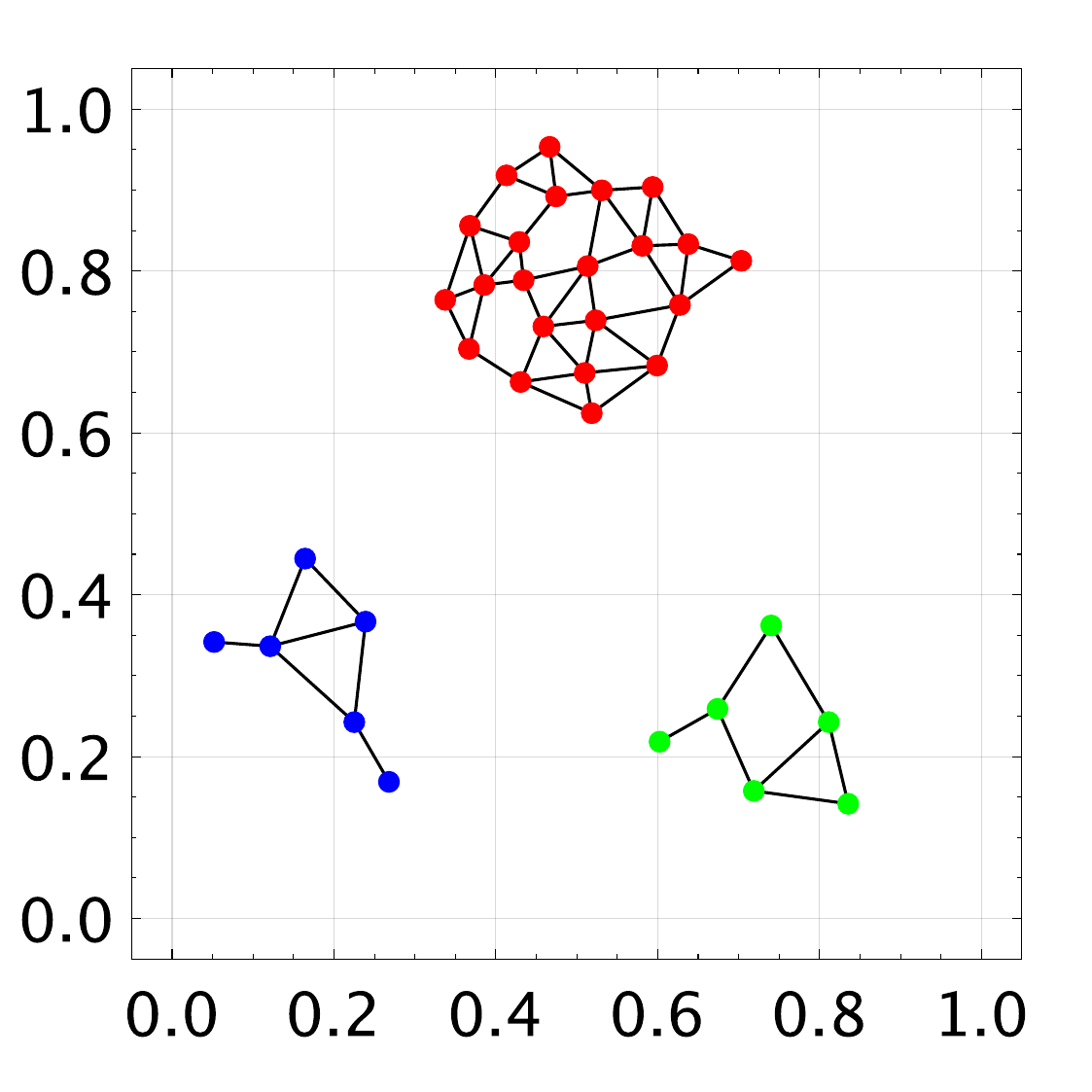}
		\label{fig:server_net_2}
	}
	\vspace{-1mm}
	\caption{Clustering results in the round \#2.}
	\label{fig:round_2}
\end{figure}

Fig. \ref{fig:round_2} shows the results in the round \#2. Since each client and the server perform continual learning, the result of each client in the round \#1 remains to the round \#2 (see Figs. \ref{fig:client_1_round_2} and \ref{fig:client_2_round_2}). In Fig. \ref{fig:server_net_2}, therefore, the cluster with red-colored nodes in the server is organized by updating the small clusters of the server in the round \#1 (i.e., Fig. \ref{fig:server_net_1}).

\begin{figure}[htbp]
	\vspace{-4mm}
	\centering
	\subfloat[Client \#1 (CA+)]{
		\includegraphics[width=1.05in]{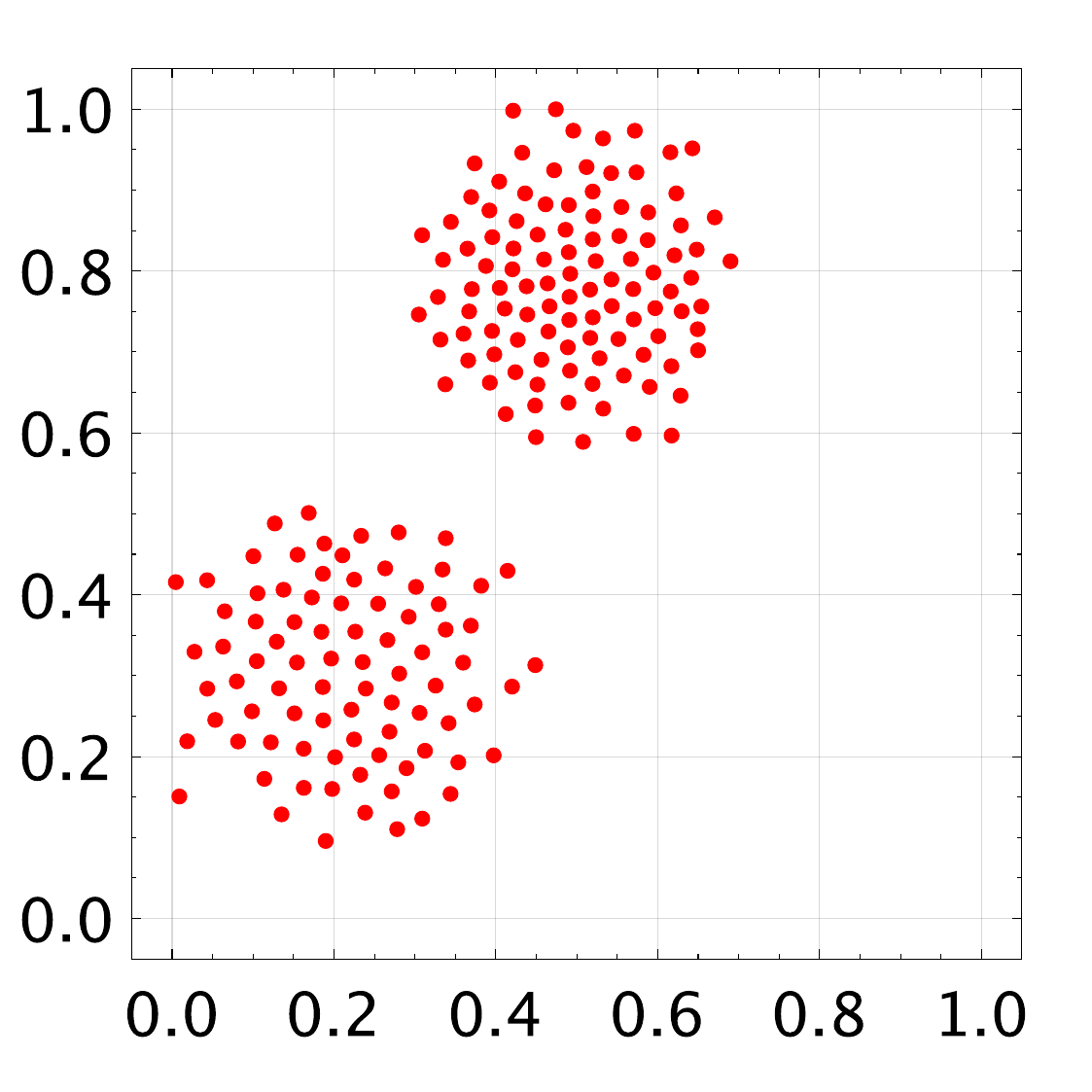}
		\label{fig:client_1_round_3}
	}\hfil
	\subfloat[Client \#2 (CA+)]{
		\includegraphics[width=1.05in]{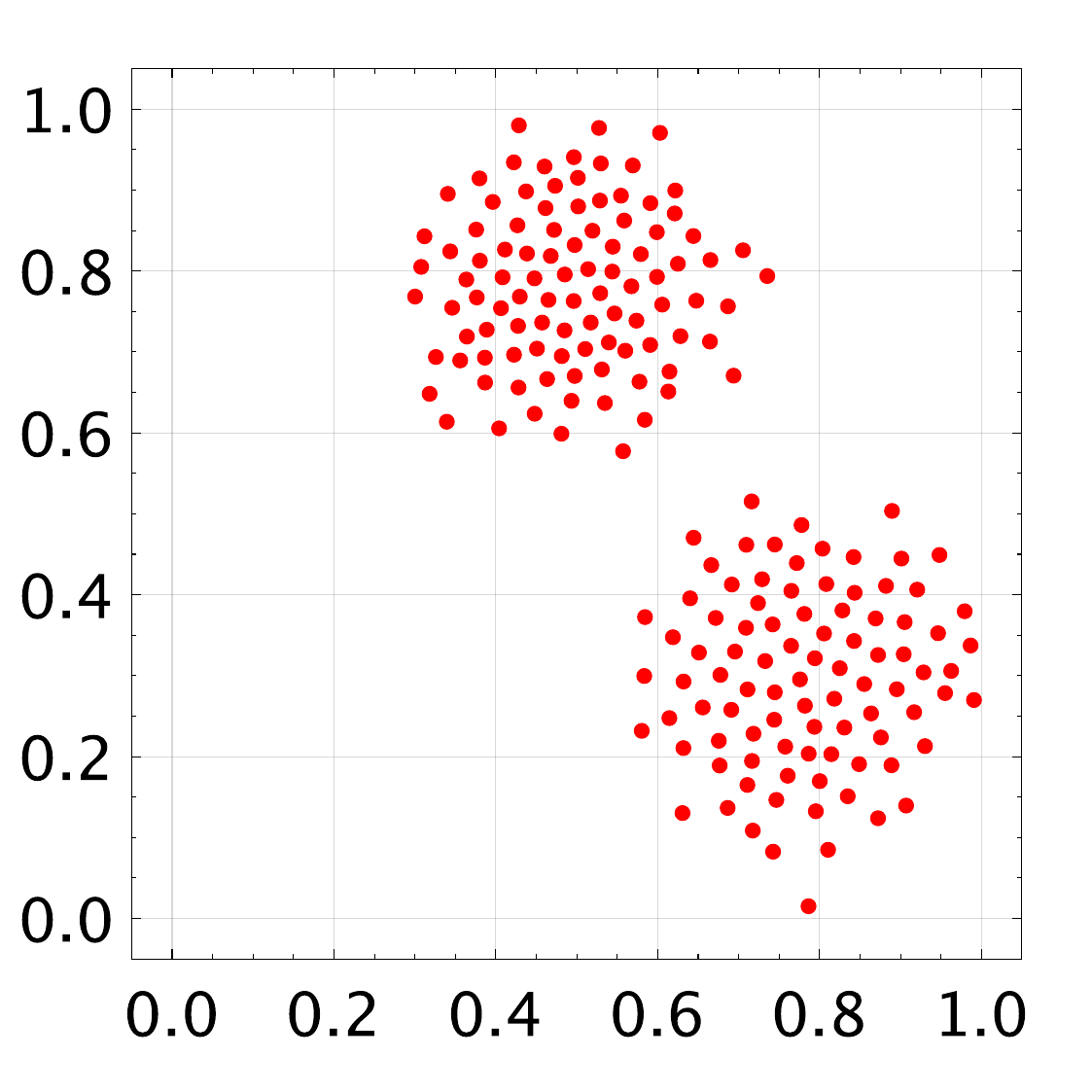}
		\label{fig:client_2_round_3}
	}\hfil
	\subfloat[Server (CAE$_{\text{FC}}$)]{
		\includegraphics[width=1.05in]{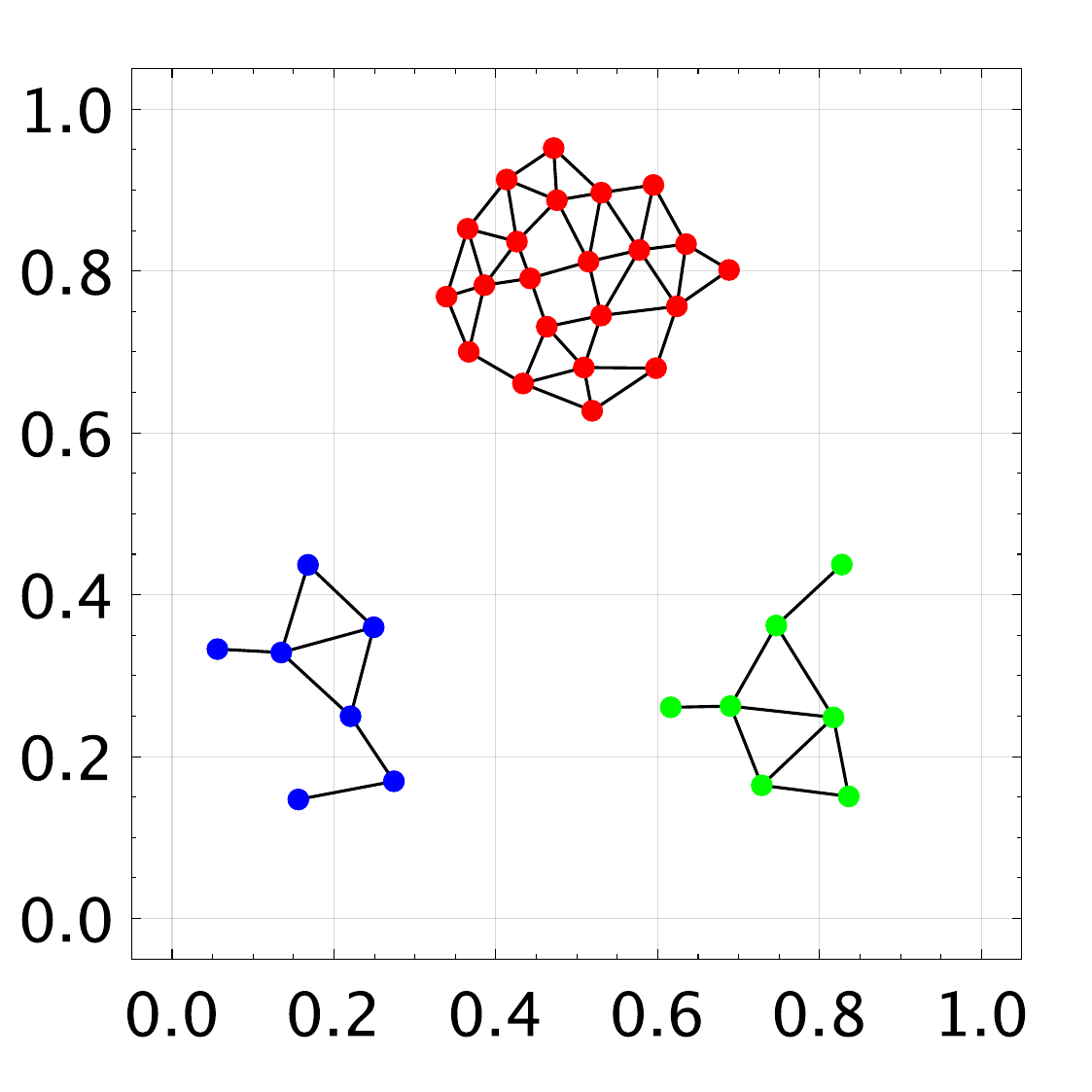}
		\label{fig:server_net_3}
	}
	\vspace{-1mm}
	\caption{Clustering results in the round \#3.}
	\label{fig:round_3}
\end{figure}

Fig. \ref{fig:round_3} shows the results in the round \#3. It can be seen that the results in the round \#3 are well-organized by updating the results in the round \#2. It is noteworthy that there is no significant change in the respective distributions of nodes between rounds \#2 and \#3 (i.e., Figs. \ref{fig:round_2} and \ref{fig:round_3}). This indicates that the distribution of the training data points is well-approximated with only a single learning epoch in rounds \#1 and \#2.

From the above-mentioned observations, we regard that FCAC can continually perform federated clustering without catastrophic forgetting.

\subsection{Clustering Performance on Real-world Datasets}
\label{sec:performance}
In this section, we evaluate the clustering performance of FCAC compared to state-of-the-art federated clustering algorithms by using real-world datasets.

\subsubsection{Compared Algorithms}
\label{sec:algorithm_real}
As federated clustering algorithms, $k$-FED \cite{dennis21}, FedFCM \cite{stallmann22}, and MUFC \cite{pan23} are selected as compared algorithms. Note that although MUFC is originally proposed in the machine unlearning domain, MUFC is treated as a compared algorithm because it has a federated clustering mechanism \cite{pan23}. In addition, we include $k$-means \cite{lloyd82} as a reference although $k$-means is not a federated clustering algorithm.

Since a base clusterer of all the compared algorithms is a centroid-based algorithm, the number of clusters and the number of iterations for convergence are required. Throughout our experiments in this section, both in clients and a server, we set the number of clusters as equal to the number of classes in each dataset, and the number of iterations for convergence is 100. Moreover, the initialization of centroids in $k$-means is performed 10 times in the same way as in $k$-means++. In contrast, FCAC has no parameters to be specified in advance, the number of iterations for convergence is 1, and no centroid initialization process is need. Note that, therefore, FCAC has much higher applicability than the other compared algorithms.

The source code of $k$-FED\footnote{\url{https://github.com/metastableB/kfed/}}, FedFCM\footnote{\url{https://github.com/stallmo/federated_clustering}}, and MUFC\footnote{\url{https://github.com/thupchnsky/mufc}} are obtained from the publicly available implementations. The source code of FCAC is available at GitHub\footnote{\url{https://github.com/Masuyama-lab/FCAC}}.

\subsubsection{Dataset}
\label{sec:dataset_real}
We use 10 real-world datasets from public repositories \cite{vanschoren14}. Table \ref{tab:datasets} summarizes statistics of the 10 real-world datasets.

To perform federated clustering, each dataset is split into an arbitrary number of clients by using codes from the personalized federated learning platform repository\footnote{\url{https://github.com/TsingZ0/PFL-Non-IID}}. In our experiments, each algorithm is evaluated by two conditions, namely an Independent and Identically Distributed (IID) scenario and a non-IID scenario. In the IID scenario, the number of data points in each client is the same, and the data distribution for each client is consistent with the entire dataset. In the non-IID scenario, on the other hand, the number of data points in each client is different. Moreover, the data distribution for each client is not consistent with the entire dataset (as a result, each client has a different data distribution). In the case of the non-IID scenario, each of the 10 datasets in Table \ref{tab:datasets} is divided into the same number of subsets as the number of clients by using Dirichlet distribution-based splitting approach with a parameter $\alpha = 0.5$ \cite{lin20}.

After splitting each of the 10 datasets, local $\epsilon$-differential privacy is applied to each data point in each client. In our experiments, we first examine $\epsilon = \infty$ to evaluate the general clustering performance of each algorithm although $\epsilon = \infty$ means no privacy guarantee (i.e., a data point has no noise). Then, we examine four values of $\epsilon$ (i.e., 15, 25, 50, 75) which are determined based on the observations in Section \ref{sec:effectLDP}. 

During experiments, all data points in each dataset are presented to each algorithm in random order. Since all algorithms are clustering algorithms, we use the same data points for training and testing, i.e., an algorithm is trained by all data points in each dataset and tested by the same data points as the training data.

Note that since $k$-means is not a federated clustering algorithm, the datasets split into each client are merged, and then $k$-means performs clustering on the merged dataset.

\begin{table}[htbp]
	\vspace{2mm}
	\centering
	\renewcommand{\arraystretch}{1.2}
	\caption{Statistics of real-world datasets}
	\label{tab:datasets}
	\scalebox{0.89}{ 
		\begin{tabular}{l|r|r|r|r}
			\hline\hline
			Dataset         & \# of Instances & \# of Features & \# of Classes & \# of Clients \\
			\hline
			Hill-Valley     & 1,212           & 100            & 2             & 5             \\
			Ozone 			& 2,534           & 73             & 2             & 5             \\
			Bioresponse     & 3,751           & 1,777          & 2             & 10            \\
			Phoneme         & 5,404           & 5              & 2             & 10            \\
			Optdigits       & 5,620           & 64             & 10            & 50            \\
			Pendigits       & 10,992          & 16             & 10            & 50            \\
			Mozilla4        & 15,545          & 5              & 2             & 10            \\
			Magic 			& 19,020          & 11             & 2             & 50            \\
			FMNIST   & 70,000          & 784            & 10            & 100           \\
			Skin            & 245,057         & 3              & 2             & 100           \\         
			\hline\hline
		\end{tabular}
	}
	\\
	\vspace{1mm}
	\footnotesize \raggedright
	\hspace{0.5mm} Ozone stands for Ozone-Level-8hr. Magic stands for Magic-Telescope.\\
    \hspace{0.5mm} FMNIST stands for Fashion-MNIST.\\
\end{table}

\begin{table*}[htbp]
	\centering
	\caption{Results of quantitative comparisons on 10 real-world datasets in the IID scenario with $\epsilon = \infty$}
	\label{tab:ResultsIID}
	\footnotesize
	\renewcommand{\arraystretch}{1.1}
	\scalebox{0.92}{
	\begin{tabular}{ll|r|r|r|r|r} \hline\hline
	Dataset     & Metric        & \multicolumn{1}{c|}{$k$-means} & \multicolumn{1}{c|}{$k$-FED}  & \multicolumn{1}{c|}{FedFCM}   & \multicolumn{1}{c|}{MUFC}    & \multicolumn{1}{c}{FCAC}   \\
	\hline
	Hill-Valley & ARI           & -0.0001 (0.0000) \colorbox{gray!35}{3} & -0.0001 (0.0000) \colorbox{gray!10}{4} & -0.0001 (0.0001) \colorbox{gray!0}{5} & 0.0000 (0.0003) \colorbox{gray!65}{2}  & 0.0011 (0.0014)   \colorbox{gray!90}{1}  \\
				& AMI           & -0.0003 (0.0000) \colorbox{gray!35}{3} & -0.0003 (0.0001) \colorbox{gray!0}{5} & -0.0003 (0.0002) \colorbox{gray!10}{4} & -0.0002 (0.0009) \colorbox{gray!65}{2} & 0.0025 (0.0022)   \colorbox{gray!90}{1}  \\
				& NMI           & 0.0005 (0.0000) \colorbox{gray!35}{3}  & 0.0005 (0.0001) \colorbox{gray!0}{5}  & 0.0005 (0.0002) \colorbox{gray!10}{4}  & 0.0007 (0.0008) \colorbox{gray!65}{2}  & 0.0076 (0.0037)   \colorbox{gray!90}{1}  \\
				& \# of Nodes    & 2.0 (0.0) \hspace{3.5mm}          & 2.0 (0.0) \hspace{3.5mm}          & 2.0 (0.0) \hspace{3.5mm}          & 2.0 (0.0) \hspace{3.5mm}          & 19.9 (4.0) \hspace{3.5mm}           \\
				& \# of Clusters & \multicolumn{1}{c|}{---}          & \multicolumn{1}{c|}{---}          & \multicolumn{1}{c|}{---}          & \multicolumn{1}{c|}{---}          & 11.6 (4.2) \hspace{3.5mm}           \\
				\hline
	Ozone       & ARI           & -0.0311 (0.0000) \colorbox{gray!0}{5} & -0.0308 (0.0008) \colorbox{gray!10}{4} & -0.0098 (0.0012) \colorbox{gray!90}{1} & -0.0219 (0.0224) \colorbox{gray!65}{2} & -0.0246 (0.0173)   \colorbox{gray!35}{3} \\
				& AMI           & 0.0187 (0.0000) \colorbox{gray!35}{3}  & 0.0186 (0.0005) \colorbox{gray!10}{4}  & 0.0189 (0.0045) \colorbox{gray!65}{2}  & 0.0175 (0.0055) \colorbox{gray!0}{5}  & 0.0221 (0.0049)   \colorbox{gray!90}{1}  \\
				& NMI           & 0.0191 (0.0000) \colorbox{gray!35}{3}  & 0.0190 (0.0005) \colorbox{gray!10}{4}  & 0.0194 (0.0045) \colorbox{gray!65}{2}  & 0.0180 (0.0055) \colorbox{gray!0}{5}  & 0.0268 (0.0039)   \colorbox{gray!90}{1}  \\
				& \# of Nodes    & 2.0 (0.0) \hspace{3.5mm}          & 2.0 (0.0) \hspace{3.5mm}          & 2.0 (0.0) \hspace{3.5mm}          & 2.0 (0.0) \hspace{3.5mm}          & 45.9 (13.1) \hspace{3.5mm}          \\
				& \# of Clusters & \multicolumn{1}{c|}{---}          & \multicolumn{1}{c|}{---}          & \multicolumn{1}{c|}{---}          & \multicolumn{1}{c|}{---}          & 29.5 (11.6) \hspace{3.5mm}          \\
				\hline
	Bioresponse & ARI           & -0.0010 (0.0000) \colorbox{gray!0}{5} & -0.0010 (0.0001) \colorbox{gray!10}{4} & 0.0101 (0.0025) \colorbox{gray!90}{1}  & 0.0079 (0.0067) \colorbox{gray!65}{2}  & 0.0003 (0.0063)   \colorbox{gray!35}{3}  \\
				& AMI           & 0.0033 (0.0000) \colorbox{gray!0}{5}  & 0.0033 (0.0001) \colorbox{gray!10}{4}  & 0.0060 (0.0006) \colorbox{gray!65}{2}  & 0.0056 (0.0027) \colorbox{gray!35}{3}  & 0.0117 (0.0067)   \colorbox{gray!90}{1}  \\
				& NMI           & 0.0035 (0.0000) \colorbox{gray!0}{5}  & 0.0035 (0.0001) \colorbox{gray!10}{4}  & 0.0062 (0.0006) \colorbox{gray!65}{2}  & 0.0058 (0.0027) \colorbox{gray!35}{3}  & 0.0151 (0.0075)   \colorbox{gray!90}{1}  \\
				& \# of Nodes    & 2.0 (0.0) \hspace{3.5mm}          & 2.0 (0.0) \hspace{3.5mm}          & 2.0 (0.0) \hspace{3.5mm}          & 2.0 (0.0) \hspace{3.5mm}          & 215.0 (14.9) \hspace{3.5mm}         \\
				& \# of Clusters & \multicolumn{1}{c|}{---}          & \multicolumn{1}{c|}{---}          & \multicolumn{1}{c|}{---}          & \multicolumn{1}{c|}{---}          & 18.6 (11.1) \hspace{3.5mm}          \\
				\hline
	Phoneme     & ARI           & 0.1195 (0.0008) \colorbox{gray!35}{3}  & 0.1155 (0.0320) \colorbox{gray!10}{4}  & 0.1397 (0.0019) \colorbox{gray!65}{2}  & 0.1536 (0.0631) \colorbox{gray!90}{1}  & 0.0659 (0.0312)   \colorbox{gray!0}{5}  \\
				& AMI           & 0.1756 (0.0007) \colorbox{gray!90}{1}  & 0.1658 (0.0147) \colorbox{gray!65}{2}  & 0.1608 (0.0016) \colorbox{gray!35}{3}  & 0.1417 (0.0462) \colorbox{gray!10}{4}  & 0.1136 (0.0115)   \colorbox{gray!0}{5}  \\
				& NMI           & 0.1758 (0.0007) \colorbox{gray!90}{1}  & 0.1659 (0.0147) \colorbox{gray!65}{2}  & 0.1610 (0.0016) \colorbox{gray!35}{3}  & 0.1418 (0.0462) \colorbox{gray!10}{4}  & 0.1149 (0.0113)   \colorbox{gray!0}{5}  \\
				& \# of Nodes    & 2.0 (0.0) \hspace{3.5mm}          & 2.0 (0.0) \hspace{3.5mm}          & 2.0 (0.0) \hspace{3.5mm}          & 2.0 (0.0) \hspace{3.5mm}          & 46.8 (10.5) \hspace{3.5mm}          \\
				& \# of Clusters & \multicolumn{1}{c|}{---}          & \multicolumn{1}{c|}{---}          & \multicolumn{1}{c|}{---}          & \multicolumn{1}{c|}{---}          & 26.9 (10.2) \hspace{3.5mm}          \\
				\hline
	Optdigits   & ARI           & 0.6719 (0.0028) \colorbox{gray!65}{2}  & 0.5836 (0.0279) \colorbox{gray!35}{3}  & 0.2224 (0.0322) \colorbox{gray!0}{5}  & 0.6922 (0.0355) \colorbox{gray!90}{1}  & 0.4425 (0.0625)   \colorbox{gray!10}{4}  \\
				& AMI           & 0.7547 (0.0035) \colorbox{gray!90}{1}  & 0.6957 (0.0140) \colorbox{gray!35}{3}  & 0.3737 (0.0278) \colorbox{gray!0}{5}  & 0.7506 (0.0188) \colorbox{gray!65}{2}  & 0.5937 (0.0352)   \colorbox{gray!10}{4}  \\
				& NMI           & 0.7555 (0.0035) \colorbox{gray!90}{1}  & 0.6966 (0.0139) \colorbox{gray!35}{3}  & 0.3755 (0.0277) \colorbox{gray!0}{5}  & 0.7514 (0.0187) \colorbox{gray!65}{2}  & 0.5986 (0.0351)   \colorbox{gray!10}{4}  \\
				& \# of Nodes    & 10.0 (0.0) \hspace{3.5mm}         & 10.0 (0.0) \hspace{3.5mm}         & 10.0 (0.0) \hspace{3.5mm}         & 10.0 (0.0) \hspace{3.5mm}         & 72.4 (21.1) \hspace{3.5mm}          \\
				& \# of Clusters & \multicolumn{1}{c|}{---}          & \multicolumn{1}{c|}{---}          & \multicolumn{1}{c|}{---}          & \multicolumn{1}{c|}{---}          & 41.0 (18.9) \hspace{3.5mm}          \\
				\hline
	Pendigits   & ARI           & 0.5449 (0.0268) \colorbox{gray!35}{3}  & 0.5245 (0.0298) \colorbox{gray!10}{4}  & 0.3428 (0.0321) \colorbox{gray!0}{5}  & 0.5645 (0.0284) \colorbox{gray!65}{2}  & 0.6185 (0.0507)   \colorbox{gray!90}{1}  \\
				& AMI           & 0.6833 (0.0045) \colorbox{gray!65}{2}  & 0.6613 (0.0135) \colorbox{gray!10}{4}  & 0.5192 (0.0301) \colorbox{gray!0}{5}  & 0.6766 (0.0151) \colorbox{gray!35}{3}  & 0.7180 (0.0275)   \colorbox{gray!90}{1}  \\
				& NMI           & 0.6838 (0.0045) \colorbox{gray!65}{2}  & 0.6619 (0.0135) \colorbox{gray!10}{4}  & 0.5201 (0.0300) \colorbox{gray!0}{5}  & 0.6772 (0.0150) \colorbox{gray!35}{3}  & 0.7195 (0.0275)   \colorbox{gray!90}{1}  \\
				& \# of Nodes    & 10.0 (0.0) \hspace{3.5mm}         & 10.0 (0.0) \hspace{3.5mm}         & 10.0 (0.0) \hspace{3.5mm}         & 10.0 (0.0) \hspace{3.5mm}         & 106.2 (16.6) \hspace{3.5mm}         \\
				& \# of Clusters & \multicolumn{1}{c|}{---}          & \multicolumn{1}{c|}{---}          & \multicolumn{1}{c|}{---}          & \multicolumn{1}{c|}{---}          & 34.4 (13.4) \hspace{3.5mm}          \\
				\hline
	Mozilla4    & ARI           & -0.0065 (0.0003) \colorbox{gray!90}{1} & -0.0078 (0.0004) \colorbox{gray!35}{3} & -0.0075 (0.0000) \colorbox{gray!65}{2} & -0.0109 (0.0168) \colorbox{gray!0}{5} & -0.0106 (0.0312)   \colorbox{gray!10}{4} \\
				& AMI           & 0.0456 (0.0001) \colorbox{gray!10}{4}  & 0.0456 (0.0023) \colorbox{gray!0}{5}  & 0.0575 (0.0000) \colorbox{gray!65}{2}  & 0.0526 (0.0099) \colorbox{gray!35}{3}  & 0.1003 (0.0156)   \colorbox{gray!90}{1}  \\
				& NMI           & 0.0457 (0.0001) \colorbox{gray!10}{4}  & 0.0457 (0.0023) \colorbox{gray!0}{5}  & 0.0576 (0.0000) \colorbox{gray!65}{2}  & 0.0526 (0.0099) \colorbox{gray!35}{3}  & 0.1026 (0.0162)   \colorbox{gray!90}{1}  \\
				& \# of Nodes    & 2.0 (0.0) \hspace{3.5mm}          & 2.0 (0.0) \hspace{3.5mm}          & 2.0 (0.0) \hspace{3.5mm}          & 2.0 (0.0) \hspace{3.5mm}          & 297.3 (91.7) \hspace{3.5mm}         \\
				& \# of Clusters & \multicolumn{1}{c|}{---}          & \multicolumn{1}{c|}{---}          & \multicolumn{1}{c|}{---}          & \multicolumn{1}{c|}{---}          & 162.6 (72.8) \hspace{3.5mm}         \\
				\hline
	Magic       & ARI           & 0.0594 (0.0000) \colorbox{gray!65}{2}  & 0.0543 (0.0018) \colorbox{gray!35}{3}  & 0.0201 (0.0005) \colorbox{gray!0}{5}  & 0.0217 (0.0145) \colorbox{gray!10}{4}  & 0.0993 (0.0442)   \colorbox{gray!90}{1}  \\
				& AMI           & 0.0209 (0.0000) \colorbox{gray!65}{2}  & 0.0182 (0.0009) \colorbox{gray!35}{3}  & 0.0070 (0.0001) \colorbox{gray!0}{5}  & 0.0070 (0.0038) \colorbox{gray!10}{4}  & 0.0900 (0.0299)   \colorbox{gray!90}{1}  \\
				& NMI           & 0.0210 (0.0000) \colorbox{gray!65}{2}  & 0.0183 (0.0009) \colorbox{gray!35}{3}  & 0.0071 (0.0001) \colorbox{gray!0}{5}  & 0.0071 (0.0038) \colorbox{gray!10}{4}  & 0.0908 (0.0302)   \colorbox{gray!90}{1}  \\
				& \# of Nodes    & 2.0 (0.0) \hspace{3.5mm}          & 2.0 (0.0) \hspace{3.5mm}          & 2.0 (0.0) \hspace{3.5mm}          & 2.0 (0.0) \hspace{3.5mm}          & 158.1 (42.1) \hspace{3.5mm}         \\
				& \# of Clusters & \multicolumn{1}{c|}{---}          & \multicolumn{1}{c|}{---}          & \multicolumn{1}{c|}{---}          & \multicolumn{1}{c|}{---}          & 27.1 (15.2) \hspace{3.5mm}          \\
				\hline
	FMNIST      & ARI           & 0.3596 (0.0155) \colorbox{gray!35}{3}  & 0.3620 (0.0193) \colorbox{gray!65}{2}  & 0.1701 (0.0029) \colorbox{gray!0}{5}  & 0.3655 (0.0249) \colorbox{gray!90}{1}  & 0.2415 (0.0675)   \colorbox{gray!10}{4}  \\
				& AMI           & 0.5154 (0.0067) \colorbox{gray!90}{1}  & 0.5115 (0.0120) \colorbox{gray!65}{2}  & 0.2549 (0.0043) \colorbox{gray!0}{5}  & 0.5092 (0.0133) \colorbox{gray!35}{3}  & 0.4607 (0.0375)   \colorbox{gray!10}{4}  \\
				& NMI           & 0.5155 (0.0067) \colorbox{gray!90}{1}  & 0.5117 (0.0120) \colorbox{gray!65}{2}  & 0.2551 (0.0043) \colorbox{gray!0}{5}  & 0.5093 (0.0133) \colorbox{gray!35}{3}  & 0.4613 (0.0375)   \colorbox{gray!10}{4}  \\
				& \# of Nodes    & 10.0 (0.0) \hspace{3.5mm}         & 10.0 (0.0) \hspace{3.5mm}         & 10.0 (0.0) \hspace{3.5mm}         & 10.0 (0.0) \hspace{3.5mm}         & 83.6 (20.8) \hspace{3.5mm}          \\
				& \# of Clusters & \multicolumn{1}{c|}{---}          & \multicolumn{1}{c|}{---}          & \multicolumn{1}{c|}{---}          & \multicolumn{1}{c|}{---}          & 35.3 (14.9) \hspace{3.5mm}          \\
				\hline
	Skin        & ARI           & -0.0397 (0.0001) \colorbox{gray!10}{4} & -0.0396 (0.0001) \colorbox{gray!35}{3} & -0.0329 (0.0155) \colorbox{gray!65}{2} & -0.0409 (0.0094) \colorbox{gray!0}{5} & 0.1276 (0.2579)   \colorbox{gray!90}{1}  \\
				& AMI           & 0.0234 (0.0001) \colorbox{gray!35}{3}  & 0.0233 (0.0001) \colorbox{gray!10}{4}  & 0.0210 (0.0019) \colorbox{gray!0}{5}  & 0.0265 (0.0107) \colorbox{gray!65}{2}  & 0.2127 (0.2002)   \colorbox{gray!90}{1}  \\
				& NMI           & 0.0234 (0.0001) \colorbox{gray!35}{3}  & 0.0233 (0.0001) \colorbox{gray!10}{4}  & 0.0210 (0.0019) \colorbox{gray!0}{5}  & 0.0265 (0.0107) \colorbox{gray!65}{2}  & 0.2127 (0.2002)   \colorbox{gray!90}{1}  \\
				& \# of Nodes    & 2.0 (0.0) \hspace{3.5mm}                              & 2.0 (0.0) \hspace{3.5mm}                              & 2.0 (0.0) \hspace{3.5mm}          & 2.0 (0.0) \hspace{3.5mm}          & 80.3 (12.2) \hspace{3.5mm}                              \\
				& \# of Clusters & \multicolumn{1}{c|}{---}                              & \multicolumn{1}{c|}{---}                              & \multicolumn{1}{c|}{---}          & \multicolumn{1}{c|}{---}          & 14.9 (8.4) \hspace{3.5mm}      \\
				\hline
				& Average Rank  & \multicolumn{1}{c|}{\cellcolor{gray!60}{2.700}}   & \multicolumn{1}{c|}{\cellcolor{gray!10}{3.533}}   & \multicolumn{1}{c|}{\cellcolor{gray!0}{3.633}}             & \multicolumn{1}{c|}{\cellcolor{gray!35}{2.900}}    & \multicolumn{1}{c}{\cellcolor{gray!90}{2.233}}          \\
				\hline\hline
	\end{tabular}
	}
	\\
	\vspace{1mm}
	\footnotesize \raggedright
	\hspace{14mm} Ozone stands for Ozone-Level-8hr. Magic stands for Magic-Telescope. FMNIST stands for Fashion-MNIST.\\
	\hspace{14mm} The best value in each metric is indicated in bold. The values in parentheses indicate the standard deviation. \\
	\hspace{14mm} A number to the right of a metric value is the rank of an algorithm corresponding to the metric value.\\
	\hspace{14mm} The smaller the rank, the better the metric score. A darker tone in a cell corresponds to a smaller rank.\\
\end{table*}

\begin{table*}[htbp]
	\centering
	\caption{Results of quantitative comparisons on 10 real-world datasets in the non-IID scenario with $\epsilon = \infty$}
	\label{tab:ResultsNonIID}
	\footnotesize
	\renewcommand{\arraystretch}{1.1}
	\scalebox{0.92}{
	\begin{tabular}{ll|r|r|r|r|r} \hline\hline
		Dataset     & Metric        & \multicolumn{1}{c|}{$k$-means}            & \multicolumn{1}{c|}{$k$-FED}              & \multicolumn{1}{c|}{FedFCM}             & \multicolumn{1}{c|}{MUFC}               & \multicolumn{1}{c}{FCAC}               \\
		\hline
		Hill-Valley & ARI           & -0.0001 (0.0000) \colorbox{gray!0}{5} & -0.0001 (0.0001) \colorbox{gray!35}{3} & -0.0001 (0.0001) \colorbox{gray!10}{4} & 0.0000 (0.0003) \colorbox{gray!65}{2}  & 0.0006 (0.0011) \colorbox{gray!90}{1}  \\
					& AMI           & -0.0003 (0.0000) \colorbox{gray!0}{5} & -0.0002 (0.0003) \colorbox{gray!35}{3} & -0.0003 (0.0002) \colorbox{gray!10}{4} & -0.0002 (0.0008) \colorbox{gray!65}{2} & 0.0013 (0.0024) \colorbox{gray!90}{1}  \\
					& NMI           & 0.0005 (0.0000) \colorbox{gray!0}{5}  & 0.0006 (0.0003) \colorbox{gray!35}{3}  & 0.0006 (0.0002) \colorbox{gray!10}{4}  & 0.0006 (0.0007) \colorbox{gray!65}{2}  & 0.0057 (0.0031) \colorbox{gray!90}{1}  \\
					& \# of Nodes    & 2.0 (0.0) \hspace{3.5mm}          & 2.0 (0.0) \hspace{3.5mm}          & 2.0 (0.0) \hspace{3.5mm}          & 2.0 (0.0) \hspace{3.5mm}          & 18.2 (3.9) \hspace{3.5mm}         \\
					& \# of Clusters & \multicolumn{1}{c|}{---}          & \multicolumn{1}{c|}{---}          & \multicolumn{1}{c|}{---}          & \multicolumn{1}{c|}{---}          & 9.3 (3.3) \hspace{3.5mm}          \\
					\hline
		Ozone       & ARI           & -0.0311 (0.0000) \colorbox{gray!0}{5} & -0.0271 (0.0021) \colorbox{gray!35}{3} & -0.0046 (0.0001) \colorbox{gray!90}{1} & -0.0149 (0.0230) \colorbox{gray!65}{2} & -0.0283 (0.0189) \colorbox{gray!10}{4} \\
					& AMI           & 0.0187 (0.0000) \colorbox{gray!65}{2}  & 0.0177 (0.0007) \colorbox{gray!10}{4}  & 0.0048 (0.0003) \colorbox{gray!0}{5}  & 0.0180 (0.0063) \colorbox{gray!35}{3}  & 0.0202 (0.0049) \colorbox{gray!90}{1}  \\
					& NMI           & 0.0192 (0.0000) \colorbox{gray!65}{2}  & 0.0182 (0.0007) \colorbox{gray!10}{4}  & 0.0052 (0.0003) \colorbox{gray!0}{5}  & 0.0185 (0.0063) \colorbox{gray!35}{3}  & 0.0254 (0.0045) \colorbox{gray!90}{1}  \\
					& \# of Nodes    & 2.0 (0.0) \hspace{3.5mm}          & 2.0 (0.0) \hspace{3.5mm}          & 2.0 (0.0) \hspace{3.5mm}          & 2.0 (0.0) \hspace{3.5mm}          & 48.8 (20.0) \hspace{3.5mm}        \\
					& \# of Clusters & \multicolumn{1}{c|}{---}          & \multicolumn{1}{c|}{---}          & \multicolumn{1}{c|}{---}          & \multicolumn{1}{c|}{---}          & 32.4 (17.5) \hspace{3.5mm}        \\
					\hline
		Bioresponse & ARI           & -0.0010 (0.0000) \colorbox{gray!10}{4} & -0.0010 (0.0000) \colorbox{gray!35}{3} & 0.0111 (0.0003) \colorbox{gray!90}{1}  & 0.0059 (0.0075) \colorbox{gray!65}{2}  & -0.0014 (0.0024) \colorbox{gray!0}{5} \\
					& AMI           & 0.0033 (0.0000) \colorbox{gray!10}{4}  & 0.0033 (0.0000) \colorbox{gray!0}{5}  & 0.0064 (0.0002) \colorbox{gray!35}{3}  & 0.0068 (0.0078) \colorbox{gray!65}{2}  & 0.0122 (0.0070) \colorbox{gray!90}{1}  \\
					& NMI           & 0.0035 (0.0000) \colorbox{gray!10}{4}  & 0.0035 (0.0000) \colorbox{gray!0}{5}  & 0.0066 (0.0002) \colorbox{gray!35}{3}  & 0.0071 (0.0078) \colorbox{gray!65}{2}  & 0.0157 (0.0080) \colorbox{gray!90}{1}  \\
					& \# of Nodes    & 2.0 (0.0) \hspace{3.5mm}          & 2.0 (0.0) \hspace{3.5mm}          & 2.0 (0.0) \hspace{3.5mm}          & 2.0 (0.0) \hspace{3.5mm}          & 185.6 (20.1) \hspace{3.5mm}       \\
					& \# of Clusters & \multicolumn{1}{c|}{---}          & \multicolumn{1}{c|}{---}          & \multicolumn{1}{c|}{---}          & \multicolumn{1}{c|}{---}          & 20.8 (12.0) \hspace{3.5mm}        \\
					\hline
		Phoneme     & ARI           & 0.1194 (0.0008) \colorbox{gray!65}{2}  & 0.1574 (0.0566) \colorbox{gray!90}{1}  & 0.0775 (0.0214) \colorbox{gray!10}{4}  & 0.0918 (0.1104) \colorbox{gray!35}{3}  & 0.0513 (0.0204) \colorbox{gray!0}{5}  \\
					& AMI           & 0.1757 (0.0007) \colorbox{gray!90}{1}  & 0.0764 (0.0282) \colorbox{gray!0}{5}  & 0.1030 (0.0260) \colorbox{gray!10}{4}  & 0.1066 (0.0520) \colorbox{gray!65}{2}  & 0.1050 (0.0084) \colorbox{gray!35}{3}  \\
					& NMI           & 0.1758 (0.0007) \colorbox{gray!90}{1}  & 0.0766 (0.0282) \colorbox{gray!0}{5}  & 0.1031 (0.0260) \colorbox{gray!10}{4}  & 0.1067 (0.0520) \colorbox{gray!65}{2}  & 0.1063 (0.0082) \colorbox{gray!35}{3}  \\
					& \# of Nodes    & 2.0 (0.0) \hspace{3.5mm}          & 2.0 (0.0) \hspace{3.5mm}          & 2.0 (0.0) \hspace{3.5mm}          & 2.0 (0.0) \hspace{3.5mm}          & 45.3 (9.0) \hspace{3.5mm}         \\
					& \# of Clusters & \multicolumn{1}{c|}{---}          & \multicolumn{1}{c|}{---}          & \multicolumn{1}{c|}{---}          & \multicolumn{1}{c|}{---}          & 27.5 (10.3) \hspace{3.5mm}       \\
					\hline
		Optdigits   & ARI           & 0.6709 (0.0008) \colorbox{gray!65}{2}  & 0.5360 (0.0421) \colorbox{gray!35}{3}  & 0.2647 (0.0269) \colorbox{gray!0}{5}  & 0.6754 (0.0311) \colorbox{gray!90}{1}  & 0.4089 (0.0469) \colorbox{gray!10}{4}  \\
					& AMI           & 0.7563 (0.0005) \colorbox{gray!90}{1}  & 0.6739 (0.0273) \colorbox{gray!35}{3}  & 0.4036 (0.0223) \colorbox{gray!0}{5}  & 0.7449 (0.0167) \colorbox{gray!65}{2}  & 0.5669 (0.0281) \colorbox{gray!10}{4}  \\
					& NMI           & 0.7570 (0.0005) \colorbox{gray!90}{1}  & 0.6750 (0.0272) \colorbox{gray!35}{3}  & 0.4057 (0.0222) \colorbox{gray!0}{5}  & 0.7457 (0.0167) \colorbox{gray!65}{2}  & 0.5725 (0.0288) \colorbox{gray!10}{4}  \\
					& \# of Nodes    & 10.0 (0.0) \hspace{3.5mm}         & 10.0 (0.0) \hspace{3.5mm}         & 10.0 (0.0) \hspace{3.5mm}         & 10.0 (0.0) \hspace{3.5mm}         & 69.3 (22.8) \hspace{3.5mm}        \\
					& \# of Clusters & \multicolumn{1}{c|}{---}          & \multicolumn{1}{c|}{---}          & \multicolumn{1}{c|}{---}          & \multicolumn{1}{c|}{---}          & 44.9 (23.6) \hspace{3.5mm}        \\
					\hline
		Pendigits   & ARI           & 0.5478 (0.0280) \colorbox{gray!35}{3}  & 0.5040 (0.0236) \colorbox{gray!10}{4}  & 0.3178 (0.0180) \colorbox{gray!0}{5}  & 0.5672 (0.0290) \colorbox{gray!65}{2}  & 0.5836 (0.0656) \colorbox{gray!90}{1}  \\
					& AMI           & 0.6837 (0.0045) \colorbox{gray!65}{2}  & 0.6512 (0.0105) \colorbox{gray!10}{4}  & 0.4867 (0.0135) \colorbox{gray!0}{5}  & 0.6820 (0.0115) \colorbox{gray!35}{3}  & 0.6994 (0.0323) \colorbox{gray!90}{1}  \\
					& NMI           & 0.6842 (0.0045) \colorbox{gray!65}{2}  & 0.6517 (0.0105) \colorbox{gray!10}{4}  & 0.4876 (0.0135) \colorbox{gray!0}{5}  & 0.6825 (0.0115) \colorbox{gray!35}{3}  & 0.7010 (0.0322) \colorbox{gray!90}{1}  \\
					& \# of Nodes    & 10.0 (0.0) \hspace{3.5mm}         & 10.0 (0.0) \hspace{3.5mm}         & 10.0 (0.0) \hspace{3.5mm}         & 10.0 (0.0) \hspace{3.5mm}         & 101.2 (19.1)  \hspace{3.5mm}      \\
					& \# of Clusters & \multicolumn{1}{c|}{---}          & \multicolumn{1}{c|}{---}          & \multicolumn{1}{c|}{---}          & \multicolumn{1}{c|}{---}          & 32.7 (11.2) \hspace{3.5mm}        \\
					\hline
		Mozilla4    & ARI           & -0.0065 (0.0002) \colorbox{gray!35}{3} & -0.0021 (0.0011) \colorbox{gray!90}{1} & -0.0078 (0.0004) \colorbox{gray!10}{4} & -0.0047 (0.0322) \colorbox{gray!65}{2} & -0.0269 (0.0421) \colorbox{gray!0}{5} \\
					& AMI           & 0.0457 (0.0001) \colorbox{gray!35}{3}  & 0.0444 (0.0005) \colorbox{gray!0}{5}  & 0.0447 (0.0003) \colorbox{gray!10}{4}  & 0.0557 (0.0111) \colorbox{gray!65}{2}  & 0.0945 (0.0198) \colorbox{gray!90}{1}  \\
					& NMI           & 0.0457 (0.0001) \colorbox{gray!35}{3}  & 0.0444 (0.0005) \colorbox{gray!0}{5}  & 0.0448 (0.0003) \colorbox{gray!10}{4}  & 0.0557 (0.0111) \colorbox{gray!65}{2}  & 0.0969 (0.0200) \colorbox{gray!90}{1}  \\
					& \# of Nodes    & 2.0 (0.0) \hspace{3.5mm}          & 2.0 (0.0) \hspace{3.5mm}          & 2.0 (0.0) \hspace{3.5mm}          & 2.0 (0.0) \hspace{3.5mm}          & 295.8 (66.2) \hspace{3.5mm}       \\
					& \# of Clusters & \multicolumn{1}{c|}{---}          & \multicolumn{1}{c|}{---}          & \multicolumn{1}{c|}{---}          & \multicolumn{1}{c|}{---}          & 151.8 (60.1) \hspace{3.5mm}       \\
					\hline
		Magic       & ARI           & 0.0594 (0.0000) \colorbox{gray!65}{2}  & 0.0582 (0.0276) \colorbox{gray!35}{3}  & 0.0234 (0.0002) \colorbox{gray!0}{5}  & 0.0255 (0.0203) \colorbox{gray!10}{4}  & 0.1268 (0.0259) \colorbox{gray!90}{1}  \\
					& AMI           & 0.0209 (0.0000) \colorbox{gray!35}{3}  & 0.0289 (0.0197) \colorbox{gray!65}{2}  & 0.0075 (0.0001) \colorbox{gray!0}{5}  & 0.0117 (0.0150) \colorbox{gray!10}{4}  & 0.1072 (0.0139) \colorbox{gray!90}{1}  \\
					& NMI           & 0.0210 (0.0000) \colorbox{gray!35}{3}  & 0.0290 (0.0197) \colorbox{gray!65}{2}  & 0.0076 (0.0001) \colorbox{gray!0}{5}  & 0.0117 (0.0150) \colorbox{gray!10}{4}  & 0.1079 (0.0140) \colorbox{gray!90}{1}  \\
					& \# of Nodes    & 2.0 (0.0) \hspace{3.5mm}          & 2.0 (0.0) \hspace{3.5mm}          & 2.0 (0.0) \hspace{3.5mm}          & 2.0 (0.0) \hspace{3.5mm}          & 143.3 (24.3) \hspace{3.5mm}       \\
					& \# of Clusters & \multicolumn{1}{c|}{---}          & \multicolumn{1}{c|}{---}          & \multicolumn{1}{c|}{---}          & \multicolumn{1}{c|}{---}          & 22.3 (9.8) \hspace{3.5mm}         \\
					\hline
		FMNIST      & ARI           & 0.3535 (0.0103) \colorbox{gray!35}{3}  & 0.3602 (0.0227) \colorbox{gray!65}{2}  & 0.2080 (0.0062) \colorbox{gray!0}{5}  & 0.3616 (0.0240) \colorbox{gray!90}{1}  & 0.2877 (0.0590) \colorbox{gray!10}{4}  \\
					& AMI           & 0.5117 (0.0010) \colorbox{gray!65}{2}  & 0.5047 (0.0255) \colorbox{gray!35}{3}  & 0.3313 (0.0054) \colorbox{gray!0}{5}  & 0.5143 (0.0167) \colorbox{gray!90}{1}  & 0.4846 (0.0305) \colorbox{gray!10}{4}  \\
					& NMI           & 0.5118 (0.0010) \colorbox{gray!65}{2}  & 0.5049 (0.0255) \colorbox{gray!35}{3}  & 0.3315 (0.0054) \colorbox{gray!0}{5}  & 0.5144 (0.0167) \colorbox{gray!90}{1}  & 0.4852 (0.0305) \colorbox{gray!10}{4}  \\
					& \# of Nodes    & 10.0 (0.0) \hspace{3.5mm}         & 10.0 (0.0) \hspace{3.5mm}         & 10.0 (0.0) \hspace{3.5mm}         & 10.0 (0.0) \hspace{3.5mm}         & 77.4 (23.3) \hspace{3.5mm}        \\
					& \# of Clusters & \multicolumn{1}{c|}{---}          & \multicolumn{1}{c|}{---}          & \multicolumn{1}{c|}{---}          & \multicolumn{1}{c|}{---}          & 41.9 (18.1) \hspace{3.5mm}        \\
					\hline
		Skin        & ARI           & -0.0397 (0.0001) \colorbox{gray!10}{4} & -0.0308 (0.0007) \colorbox{gray!35}{3} & -0.0286 (0.0018) \colorbox{gray!65}{2} & -0.0435 (0.0122) \colorbox{gray!0}{5} & 0.2245 (0.2451) \colorbox{gray!90}{1}  \\
					& AMI           & 0.0234 (0.0001) \colorbox{gray!35}{3}  & 0.0179 (0.0004) \colorbox{gray!10}{4}  & 0.0161 (0.0007) \colorbox{gray!0}{5}  & 0.0299 (0.0180) \colorbox{gray!65}{2}  & 0.3054 (0.1916) \colorbox{gray!90}{1}  \\
					& NMI           & 0.0234 (0.0001) \colorbox{gray!35}{3}  & 0.0179 (0.0004) \colorbox{gray!10}{4}  & 0.0161 (0.0007) \colorbox{gray!0}{5}  & 0.0299 (0.0180) \colorbox{gray!65}{2}  & 0.3054 (0.1916) \colorbox{gray!90}{1}  \\
					& \# of Nodes    & 2.0 (0.0) \hspace{3.5mm}                              & 2.0 (0.0) \hspace{3.5mm}                              & 2.0 (0.0) \hspace{3.5mm}                              & 2.0 (0.0) \hspace{3.5mm}                              & 72.2 (11.9) \hspace{3.5mm}                            \\
					& \# of Clusters & \multicolumn{1}{c|}{---}                              & \multicolumn{1}{c|}{---}                              & \multicolumn{1}{c|}{---}                              & \multicolumn{1}{c|}{---}                              & 18.7 (10.2) \hspace{3.5mm}                            \\
					\hline
					& Average Rank  & \multicolumn{1}{c|}{\cellcolor{gray!35}{2.833}}           & \multicolumn{1}{c|}{\cellcolor{gray!10}{3.400}}             & \multicolumn{1}{c|}{\cellcolor{gray!0}{4.200}}         & \multicolumn{1}{c|}{\cellcolor{gray!60}{2.333}}          & \multicolumn{1}{c}{\cellcolor{gray!90}{2.233}}       \\
				\hline\hline
	\end{tabular}
	}
	\\
	\vspace{1mm}
	\footnotesize \raggedright
	\hspace{14mm} Ozone stands for Ozone-Level-8hr. Magic stands for Magic-Telescope. FMNIST stands for Fashion-MNIST.\\
	\hspace{14mm} The best value in each metric is indicated in bold. The values in parentheses indicate the standard deviation. \\
	\hspace{14mm} A number to the right of a metric value is the rank of an algorithm corresponding to the metric value.\\
	\hspace{14mm} The smaller the rank, the better the metric score. A darker tone in a cell corresponds to a smaller rank.\\
\end{table*}

\subsubsection{Results of General Clustering Performance}
\label{sec:results_inf}
Tables \ref{tab:ResultsIID} and \ref{tab:ResultsNonIID} show the results of clustering performance on the 10 real-world datasets in the IID and non-IID scenarios with $\epsilon = \infty$, respectively. The clustering performance of each algorithm is measured by the Adjusted Rand Index (ARI) \cite{hubert85}, the Adjusted Mutual Information (AMI) \cite{vinh10}, and the Normalized Mutual Information (NMI) \cite{strehl02}. With respect to FCAC, the number of nodes and clusters in the server (i.e., CAE$_{\text{FC}}$) are also shown. We repeat the evaluation 20 times with different random seeds for obtaining consistent averaging results. The best value in each metric is indicated in bold, and the values in parentheses indicate the standard deviation. A number to the right of each evaluation metric is the rank of an algorithm corresponding to the metric value. The smaller the rank, the better the metric score. In addition, a darker tone in a cell corresponds to a smaller rank (i.e., better evaluation).

As general trends, FCAC shows better clustering performance than the other algorithms in both scenarios, and MUFC and $k$-means show are better clustering performance than $k$-FED and FedFCM. As mentioned in Section \ref{sec:algorithm_real}, FCAC has no parameters to be specified in advance, the number of iterations for convergence is 1, and no centroid initialization process is needed, while all the compared algorithms have a parameter to be specified in advance, and require a number of iterative processes for good clustering performance. This clearly highlights the advantages of FCAC for situations where the distribution of a dataset is unknown and/or the size of a dataset is large.

With respect to the number of clusters of FCAC, FCAC tends to generate a large number of clusters compared to the true number of classes in each dataset as shown in Tables \ref{tab:ResultsIID} and \ref{tab:ResultsNonIID}. This property has positive impacts on clustering performance in many cases. In particular, FCAC shows high clustering performance in the Magic and Skin datasets. In addition, although FCAC shows a low rank in the Phoneme, Optdigits, and FMNIST datasets, the values of ARI, AMI, and NMI are not extremely low compared to the other algorithms. Note that, in general, it is difficult to discuss the relation between the number of clusters and clustering performance in the case of self-organizing algorithms that adaptively generate nodes corresponding to data points sampled from an unknown data distribution, such as GNG-, and ART-based clustering algorithms.

For statistical comparisons of the results in Tables \ref{tab:ResultsIID} and \ref{tab:ResultsNonIID}, the Friedman test and Nemenyi post-hoc analysis \cite{demvsar06} are used. The Friedman test is used to test the null hypothesis that all algorithms perform equally. If the null hypothesis is rejected, the Nemenyi post-hoc analysis is then conducted. The Nemenyi post-hoc analysis is used for all pairwise comparisons based on the ranks of results on each evaluation metric over all datasets. The difference in the performance between two algorithms is treated as statistically significant if the $ p $-value defined by the Nemenyi post-hoc analysis is smaller than the significance level. Here, the null hypothesis is rejected at the significance level of $ 0.05 $ both in the Friedman test and the Nemenyi post-hoc analysis.

\begin{figure}[htbp]
	\centering
	\subfloat[Overall]{
		\includegraphics[width=3.3in]{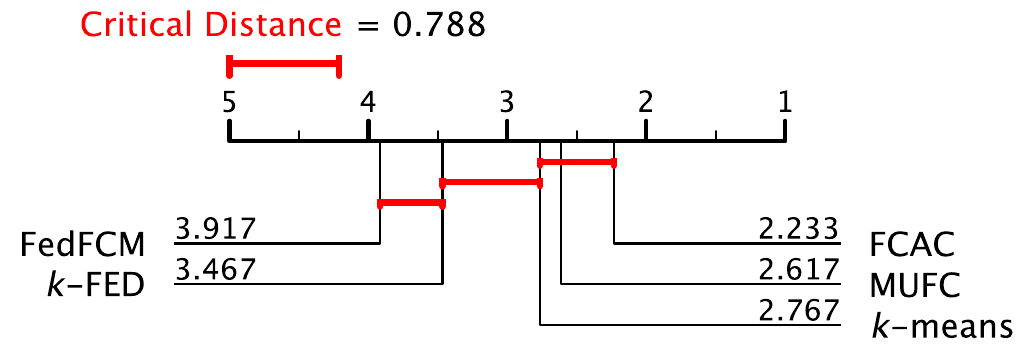}
		\label{fig:cd_all_inf}
	}\hfil \vspace{-1mm}
	\subfloat[IID scenario]{
		\includegraphics[width=3.3in]{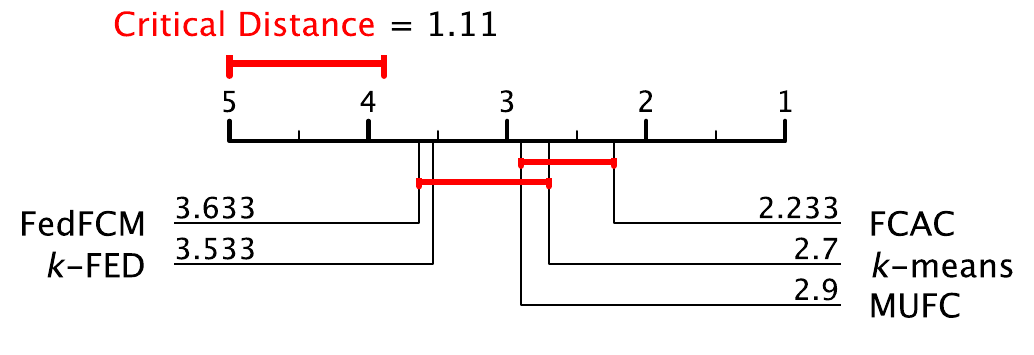}
		\label{fig:cd_iid_inf}
	}\hfil \vspace{1mm}
	\subfloat[non-IID scenario]{
		\includegraphics[width=3.3in]{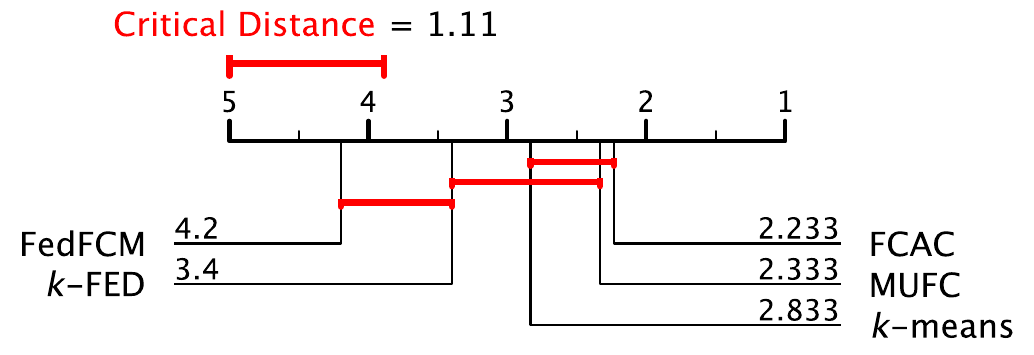}
		\label{fig:cd_niid_inf}
	}\hfil \vspace{-1mm}
	\caption{Critical difference diagram based on the results of ARI, AMI, and NMI with $\epsilon = \infty$.}
	\label{fig:cd_inf}
\end{figure}

Fig. \ref{fig:cd_inf} shows critical difference diagrams based on the results of ARI, AMI, and NMI by each algorithm, which are defined by the Nemenyi post-hoc analysis. A better result has a lower average rank, i.e., on the right side of each diagram. In theory, algorithms within a critical distance (i.e., a red line) do not have a statistically significance difference \cite{demvsar06}. Fig. \ref{fig:cd_all_inf} shows a critical difference diagram based on the overall results (i.e., all the results of ARI, AMI, and NMI in the IID and non-IID scenarios). FCAC is the lowest rank (i.e., best) algorithm with a statistically significant difference from $k$-FED and FedFCM. Figs. \ref{fig:cd_iid_inf} and \ref{fig:cd_niid_inf} are critical difference diagrams correspond to the results in Tables \ref{tab:ResultsIID} and \ref{tab:ResultsNonIID}, respectively. The ranks of MUFC and FedFCM differ depending on the scenario (i.e., IID or non-IID), which implies the instability of their learning.

The above-mentioned observations suggest that FCAC has superior clustering performance to state-of-the-art algorithms on various datasets with $\epsilon = \infty$.

\subsubsection{Results of Clustering Performance on Datasets with Local $\epsilon$-Differential Privacy}
\label{sec:results_dp}
For the comparisons of clustering performance on the 10 datasets with local $\epsilon$-differential privacy, we set $\epsilon = \{15, 25, 50, 75\}$ and then conduct the same experiments as Section \ref{sec:results_inf} for obtaining ARI, AMI, and NMI. As mentioned in Section \ref{sec:localDP}, the value of $\epsilon$ controls the degree of data privacy protection, i.e., the smaller $\epsilon$ value provides higher data privacy, while the larger $\epsilon$ value provides lower data privacy.

Similar to Section \ref{sec:results_inf}, the Friedman test and Nemenyi post-hoc analysis are used. The Friedman test is used to test the null hypothesis that all algorithms perform equally. If the null hypothesis is rejected, the Nemenyi post-hoc analysis is then conducted. Here, the null hypothesis is rejected at the significance level of $ 0.05 $ both in the Friedman test and the Nemenyi post-hoc analysis.

\begin{figure*}[htbp]
	\centering
	\subfloat[$\epsilon = 15$]{
		\includegraphics[width=3.3in]{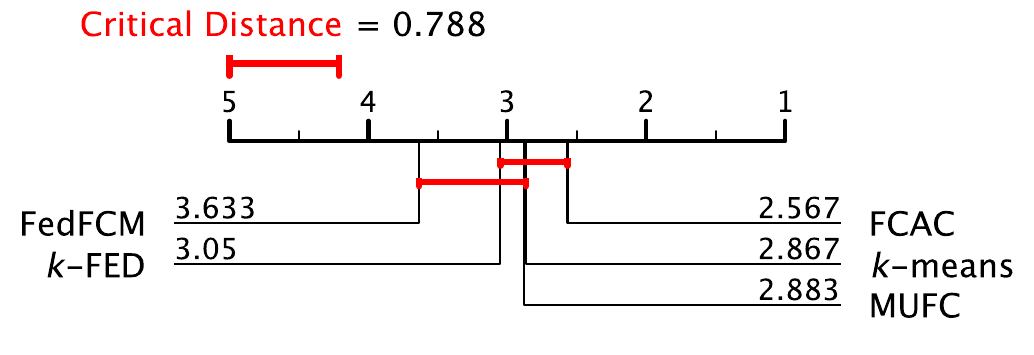}
		\label{fig:cd_all_15}
	}\hfil
	\subfloat[$\epsilon = 25$]{
		\includegraphics[width=3.3in]{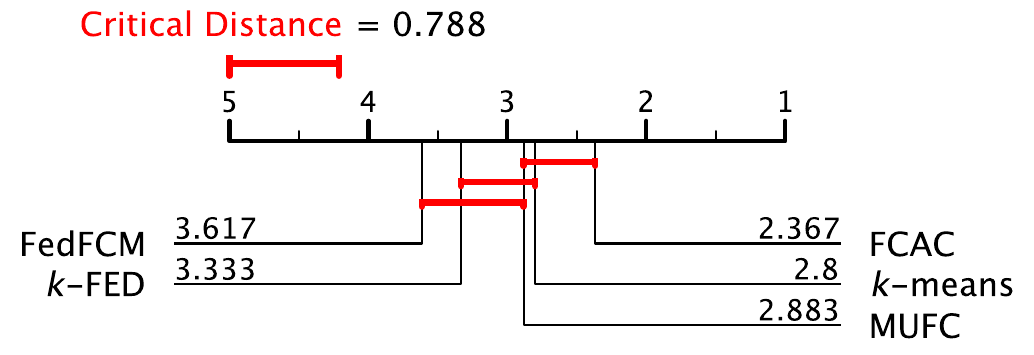}
		\label{fig:cd_all_25}
	}\hfil \\
	\subfloat[$\epsilon = 50$]{
		\includegraphics[width=3.3in]{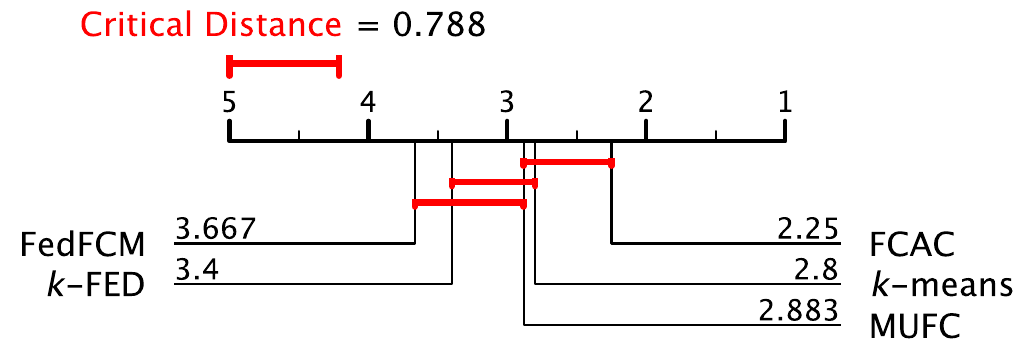}
		\label{fig:cd_all_50}
	}\hfil
	\subfloat[$\epsilon = 75$]{
		\includegraphics[width=3.3in]{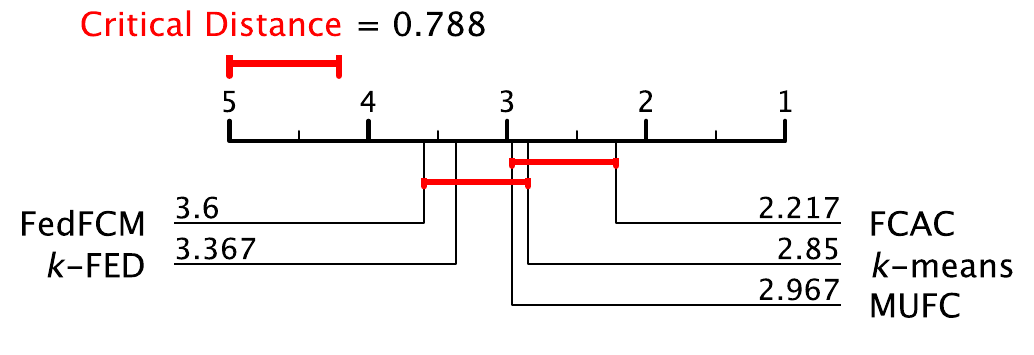}
		\label{fig:cd_all_75}
	}
	\vspace{-1mm}
	\caption{Critical difference diagram based on the overall results of ARI, AMI, and NMI with $\epsilon = \{15, 25, 50, 75\}$.}
	\label{fig:cd_all_15_25_50_75}
\end{figure*}

In this section, due to page limitations, we only show the critical difference diagrams based on the overall results of ARI, AMI, and NMI. Fig. \ref{fig:cd_all_15_25_50_75} shows the critical difference diagram corresponding to $\epsilon = 15, 25, 50, 75$. Except for $\epsilon = 15$, FCAC is the lowest rank (i.e., best) algorithm with a statistically significant difference from $k$-FED and FedFCM. This indicates that FCAC can maintain higher clustering performance for various privacy-preserving datasets than the other state-of-the-art algorithms.




\subsection{Computational Complexity}
\label{sec:compComplexity}
In FCAC, the computations on client-side can be performed in parallel, and therefore computational complexity is defined by the learning procedure of CA+, the re-ordering of training data points for CAE$_{\text{FC}}$, and the learning procedure of CAE$_{\text{FC}}$. Furthermore, since CA+ is a variant of CAE$_{\text{FC}}$, i.e., CAE$_{\text{FC}}$ without topology, we only consider the computational complexity of the re-ordering of training data points for CAE$_{\text{FC}}$ and the learning procedure of CAE$_{\text{FC}}$.

For computational complexity analysis, we use the notations in Table \ref{tab:notations}, namely $ d $ is the dimensionality of a data point, $ n $ is the number of data points, $ K $ is the number of nodes, $\mathcal{M}$ is a set of winning counts, $ \lambda $ is the number of active nodes, and $ |\mathcal{E}| $ is the number of elements in the ages of edges set $\mathcal{E}$.

The computational complexity of the re-ordering of training data points for CAE$_{\text{FC}}$ is as follows: for finding the 75th percentile of elements in $\mathcal{M}$ is $ \mathcal{O}(K) $ (line 3 in Alg. \ref{alg:sortnodes}), for splitting a node set is $ \mathcal{O}(K) $ (line 4 in Alg. \ref{alg:sortnodes}), and for shuffling the splitted nodes is $ \mathcal{O}(K^{\geq75\text{th}}) $ and $ \mathcal{O}(K^{<75\text{th}}) $ (lines 7-8 in Alg. \ref{alg:sortnodes}).

The computational complexity of the learning procedure of CAE$_{\text{FC}}$ is as follows: for computing a bandwidth of a kernel function in CIM is $ \mathcal{O}(d) $, for calculating a pairwise similarity matrix by using CIM is $ \mathcal{O}((\frac{\lambda}{2})^{2}dK) $ (line 5 in Alg. \ref{alg:pseudocodeCAE}), for calculating determinant of the pairwise similarity matrix is $ \mathcal{O}((\frac{\lambda}{2})^{3}) $ (line 6 in Alg. \ref{alg:pseudocodeCAE}), for computing CIM is $ \mathcal{O}(ndK) $ (line 14 in Alg. \ref{alg:pseudocodeCAE}), for finding nodes which have the 1st and 2nd smallest CIM value is $ \mathcal{O}(K) $ (line 14 in Alg. \ref{alg:pseudocodeCAE}), and for estimating the edge deletion threshold is $ \mathcal{O}(|\mathcal{E}|\log{|\mathcal{E}|}) $ (lines 31-34 in Alg. \ref{alg:pseudocodeCAE}). 

In general, $ n < \lambda^{3} $, $ K \ll n $, and $ \lambda < K $. As a result, the computational complexity of FCAC is $ \mathcal{O}( \lambda^{3}dK ) $.

\section{Concluding Remarks}
\label{sec:conclusion}
This paper proposed a new privacy-preserving continual federated clustering algorithm, called FCAC. FCAC uses an ART-based clustering algorithm capable of continual learning as a base clusterer for both clients and a server, and therefore FCAC is also capable of continual learning. Moreover, FCAC applies local differential privacy along with a federated learning framework to explicitly consider the data privacy protection. Empirical studies with the synthetic and real-world datasets showed that the clustering performance of FCAC is superior to state-of-the-art federated clustering algorithms while maintaining data privacy protection and the continual learning ability.

A future research topic is to incorporate deep learning techniques into FCAC in order to achieve further improvements in clustering performance.


%

%

  \section*{Acknowledgment}
This work was supported by the Japan Society for the Promotion of Science (JSPS) KAKENHI Grant Number JP19K20358 and 22H03664, National Natural Science Foundation of China (Grant No. 62250710163, 62250710682), Guangdong Provincial Key Laboratory (Grant No. 2020B121201001), the Program for Guangdong Introducing Innovative and Enterpreneurial Teams (Grant No. 2017ZT07X386), The Stable Support Plan Program of Shenzhen Natural Science Fund (Grant No. 20200925174447003), Shenzhen Science and Technology Program (Grant No. KQTD2016112514355531), JST [Moonshot RnD][Grant Number JP- MJMS2034], and TMU local 5G research support.

\ifCLASSOPTIONcaptionsoff
  \newpage
\fi



\bibliographystyle{IEEEtran}
\bibliography{myref}

%
%

%

\begin{IEEEbiography}[{\includegraphics[width=1in,height=1.25in,clip,keepaspectratio]{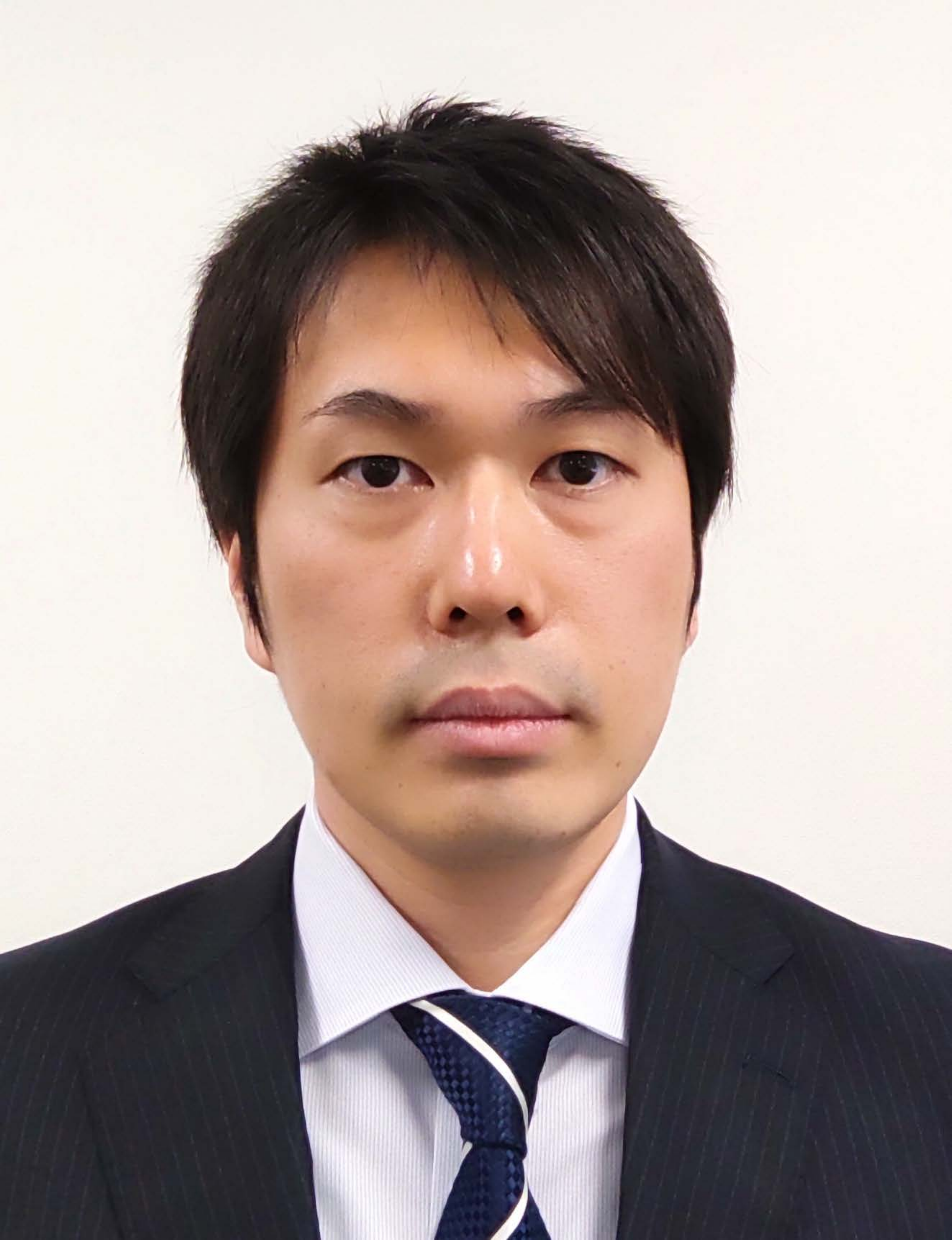}}]{Naoki Masuyama}
	(S'12--M'16) received the B.Eng. degree from Nihon University, Funabashi, Japan, in 2010, the M.E. degree from Tokyo Metropolitan University, Hino, Japan in 2012, and the Ph.D. degree from the Faculty of Computer Science and Information Technology, University of Malaya, Kuala Lumpur, Malaysia, in 2016. He is currently an Associate Professor with the Department of Core Informatics, Graduate School of Informatics, Osaka Metropolitan University, Sakai, Japan. 
	
	His current research interests include clustering, data mining, and continual learning.
\end{IEEEbiography}
\begin{IEEEbiography}[{\includegraphics[width=1in,height=1.25in,clip,keepaspectratio]{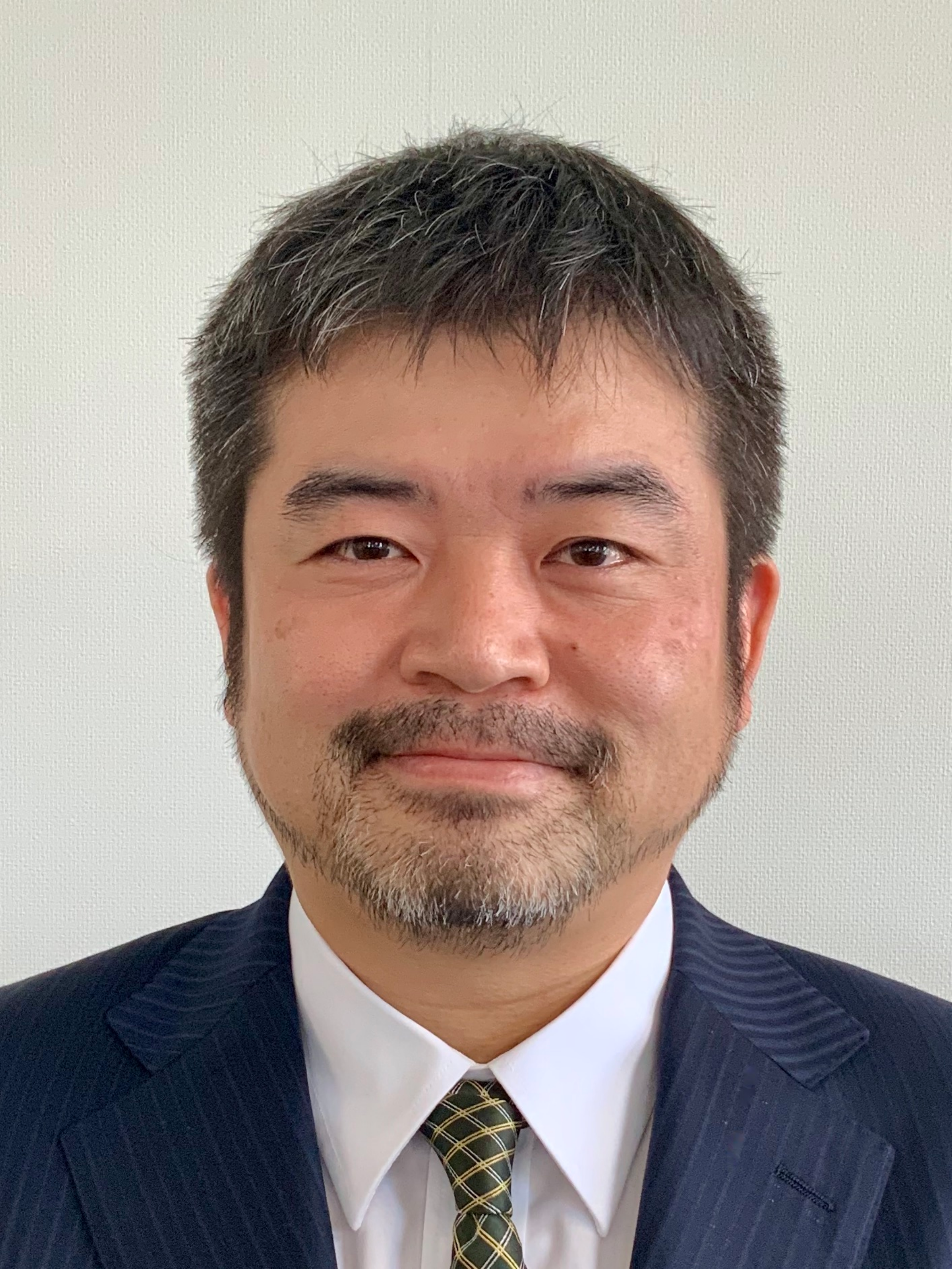}}]{Yusuke Nojima}
	received the B.S. and M.S. Degrees in mechanical engineering from Osaka Institute of Technology, Osaka, Japan, in 1999 and 2001, respectively, and the Ph.D. degree in system function science from Kobe University, Hyogo, Japan, in 2004.
	
	Since 2004, he has been with Osaka Prefecture University, Osaka, Japan, where he was a Professor in Department of Computer Science and Intelligent Systems from October 2020. From April 2022, he is a Professor in Department of Core Informatics, Graduate School of Informatics, Osaka Metropolitan University.
	
	His research interests include evolutionary fuzzy systems, evolutionary multiobjective optimization, and multiobjective data mining. He was a guest editor for several special issues in international journals. He was a task force chair on Evolutionary Fuzzy Systems in Fuzzy Systems Technical Committee of IEEE Computational Intelligence Society. He was an associate editor of IEEE Computational Intelligence Magazine (2014-2019).
\end{IEEEbiography}
\begin{IEEEbiography}[{\includegraphics[width=1in,height=1.25in,clip,keepaspectratio]{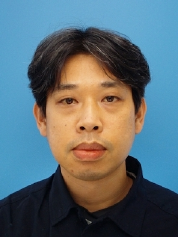}}]{Yuichiro Toda}
	received the B.E. degree and M.E. degree and Ph.D. degree from Tokyo Metropolitan University, Hino, Japan in 2011, 2013, and 2017, respectively. He was an Assistant Professor with the Faculty of Engineering, Okayama University, Okayama, Japan from 2018. He is currently an Associate Professor at Faculty of Environmental, Life, Natural Science and Technology, Okayama University, Japan. 
	
	His research interests include computational intelligence and intelligent robotics in unknown environment. He has published more than 80 refereed journal and conference papers in the interest research area.
\end{IEEEbiography}
\begin{IEEEbiography}[{\includegraphics[width=1in,height=1.25in,clip,keepaspectratio]{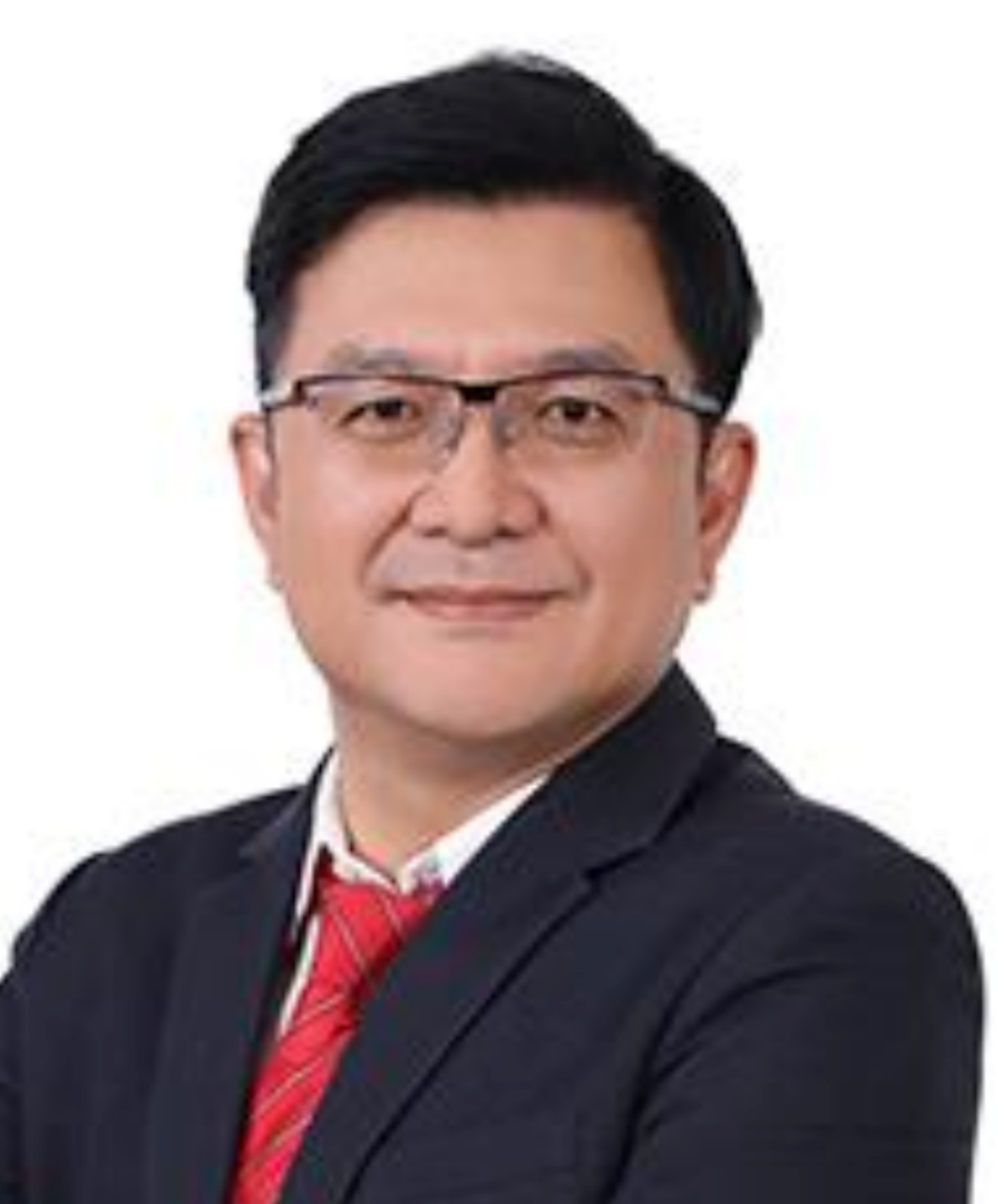}}]{Chu Kiong Loo}
	(SM'14)  holds a Ph.D. (University Sains Malaysia) and B.Eng. (First Class Hons in Mechanical Engineering from the University of Malaya).
	
	He was a Design Engineer in various industrial firms and is the founder of the Advanced Robotics Lab. at the University of Malaya. He has been involved in the application of research into Perus's Quantum Associative Model and Pribram's Holonomic Brain Model in humanoid vision projects. Currently, he is Professor of Computer Science and Information Technology at the University of Malaya, Malaysia. He has led many projects funded by the Ministry of Science in Malaysia and the High Impact Research Grant from the Ministry of Higher Education, Malaysia. Loo's research experience includes brain-inspired quantum neural networks, constructivism-inspired neural networks, synergetic neural networks and humanoid research.
\end{IEEEbiography}
\begin{IEEEbiography}[{\includegraphics[width=1in,height=1.25in,clip,keepaspectratio]{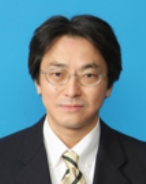}}]{Hisao Ishibuchi}
	(M'93--SM'10--F'14) received the B.S. and M.S. degrees in precision mechanics from Kyoto University, Kyoto, Japan, in 1985 and 1987, respectively, and the Ph.D. degree in computer science from Osaka Prefecture University, Sakai, Osaka, Japan, in 1992.
	
	Since 1987, he had been with Osaka Prefecture University for 30 years. He is currently a Chair Professor with the Department of Computer Science and Engineering, Southern University of Science Technology, Shenzhen, China. His current research interests include fuzzy rule-based classifiers, evolutionary multiobjective optimization, many-objective optimization, and memetic algorithms. 
	
	Dr. Ishibuchi was the IEEE Computational Intelligence Society (CIS) VicePresident for Technical Activities from 2010 to 2013. He was an IEEE CIS AdCom Member from 2014 to 2019, and from 2021 to 2023, an IEEE CIS Distinguished Lecturer from 2015 to 2017, and from 2021 to 2023, and the Editor-in-Chief of the IEEE Computational Intelligence Magazine from 2014 to 2019. He is also an Associate Editor of the ACM Computing Survey, the IEEE Transactions ON Cybernetics, and the IEEE Access.
\end{IEEEbiography}
\begin{IEEEbiography}[{\includegraphics[width=1in,height=1.25in,clip,keepaspectratio]{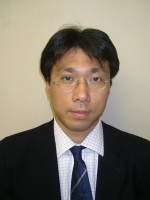}}]{Naoyuki Kubota}
	(M'01) received the M.Eng. degree from Hokkaido University, Hokkaido, Japan, in 1994, and the D.E. degree from Nagoya University, Nagoya, Japan, in 1997.

	He joined the Osaka Institute of Technology, Osaka, Japan, in 1997. In 2000, he joined the Department of Human and Artificial Intelligence Systems, Fukui University, Fukui, Japan, as an Associate Professor. He joined the Department of Mechanical Engineering, Tokyo Metropolitan University, Tokyo, Japan, in 2004, where he is a Professor with the Department of System Design.
\end{IEEEbiography}
\vfill

%
%
%




\end{document}